\documentclass[final,12pt,authoryear,times]{elsarticle}
\usepackage[T1]{fontenc}
\usepackage{graphicx}
\usepackage[table,xcdraw]{xcolor}
\usepackage{graphicx}
\usepackage{caption}
\usepackage{makecell}
\usepackage{array, booktabs, cite}
\usepackage{enumitem,kantlipsum}
\usepackage{subcaption}
\usepackage{multirow}
\usepackage{graphicx}
\usepackage{hyperref}
\usepackage{float}
\usepackage[absolute,overlay]{textpos}
\usepackage{boxedminipage}
\usepackage{amsmath}
\usepackage{url}
\usepackage[export]{adjustbox}%
\usepackage{setspace}
\usepackage{pdflscape}
\usepackage[hmargin=3.2cm]{geometry}
\usepackage{xcolor}
\definecolor{lightgray}{rgb}{0.95,0.95,0.95}
\hypersetup{hypertex=true,
    colorlinks=true,
    linkcolor=true,
    anchorcolor=true,
    citecolor=true}

\renewcommand{\arraystretch}{1.25}
\usepackage{amssymb}
\usepackage{lineno}

\author[doa]{Xiucheng Liang}
\author[sl]{Jinheng Xie}
\author[stu]{Tianhong Zhao}
\author[doa]{Rudi Stouffs}
\author[doa,dre]{Filip Biljecki\corref{cor1}}

\affiliation[doa]{organization={Department of Architecture, National University of Singapore}, country={Singapore}}
\affiliation[sl]{organization={Department of Electrical and Computer Engineering, National University of Singapore}, country={Singapore}}
\affiliation[stu]{organization={School of Artificial Intelligence, Shenzhen Technology University}, city={Shenzhen}, country={China}}
\cortext[cor1]{Corresponding author}
\affiliation[dre]{organization={Department of Real Estate, National University of Singapore}, country={Singapore}}

\journal{}

\begin{document}

\makeatletter
\renewcommand\paragraph{\@startsection{paragraph}{4}{\z@}%
  {3.25ex \@plus1ex \@minus.2ex}%
  {-1em}%
  {\normalfont\normalsize\bfseries}}
\makeatother

\begin{sloppypar}
\begin{frontmatter}

\title{OpenFACADES: An Open Framework for Architectural Caption and Attribute Data Enrichment via Street View Imagery}

\abstracttitle{Abstract}
\begin{abstract}

\begin{textblock*}{\textwidth}(3.2cm,-0.1cm) 
\begin{center}
\begin{footnotesize}
\begin{boxedminipage}{1\textwidth}
This is the Accepted Manuscript version of an article published by Elsevier in the journal \emph{ISPRS Journal of Photogrammetry and Remote Sensing} in 2025, which is available at:\\ \url{https://doi.org/10.1016/j.isprsjprs.2025.10.014}\\ Cite as:
Liang X, Xie J, Zhao T, Stouffs R, Biljecki F (2025): OpenFACADES: An open framework for architectural caption and attribute data enrichment via street view imagery. \textit{ISPRS Journal of Photogrammetry and Remote Sensing} 230: 918–942.
\end{boxedminipage}
\end{footnotesize}
\end{center}
\end{textblock*}

\begin{textblock*}{1.5\textwidth}(2.2cm,25cm)
{\tiny{\copyright{ }2025, Elsevier. Licensed under the Creative Commons Attribution-NonCommercial-NoDerivatives 4.0 International (\url{http://creativecommons.org/licenses/by-nc-nd/4.0/})}}
\end{textblock*}

Building properties, such as height, usage, and material, play a crucial role in spatial data infrastructures, supporting various urban applications.
Despite their importance, comprehensive building attribute data remain scarce in many urban areas.
Recent advances have enabled the extraction of objective building attributes using remote sensing and street-level imagery. 
However, establishing a pipeline that integrates diverse open datasets, acquires holistic building imagery, and infers comprehensive building attributes at scale remains a significant challenge.
Among the first, this study bridges the gaps by introducing OpenFACADES, an open framework that leverages multimodal crowdsourced data to enrich building profiles with both objective attributes and semantic descriptors through multimodal large language models.
First, we integrate street-level image metadata from Mapillary with OpenStreetMap geometries via isovist analysis, identifying images that provide suitable vantage points for observing target buildings. 
Second, we automate the detection of building facades in panoramic imagery and tailor a reprojection approach to convert objects into holistic perspective views that approximate real-world observation.
Third, we introduce an innovative approach that harnesses and investigates the capabilities of open-source large vision-language models (VLMs) for multi-attribute prediction and open-vocabulary captioning in building-level analytics, leveraging a globally sourced dataset of 31,180 labeled images from seven cities.
Evaluation shows that fine-tuned VLM excel in multi-attribute inference, outperforming single-attribute computer vision models and zero-shot ChatGPT-4o. Further experiments confirm its superior generalization and robustness across culturally distinct region and varying image conditions. Finally, the model is applied for large-scale building annotation, generating a dataset of 1.2 million images for half a million buildings.
This open‐source framework enhances the scope, adaptability, and granularity of building‐level assessments, enabling more fine‐grained and interpretable insights into the built environment.
Our dataset and code are available openly at: \url{https://github.com/seshing/OpenFACADES}.

\end{abstract}


\end{frontmatter}

\section{Introduction}
Buildings, as prominent artifacts within urban settings, serve as vital indicators of the management, transformation, and overall dynamism of the built environment. 
Their physical characteristics, including geometry, height, function, material, condition, and style, are the key parameters that not only support sustainable urban development but also reflect economic progress and cultural evolution over time \citep{biljecki_open_2021}.
Such rich building-level data has been instrumental in a range of applications, such as urban climate simulations for improved environmental planning~\citep{creutzig2019upscaling}, building energy modeling for resource optimization~\citep{kumar2018novel, roth2020syncity}, estimation of urban material stocks for the circular economy~\citep{raghu_towards_2023}, and disaster impact assessments to inform effective response and recovery efforts~\citep{westrope2014groundtruthing}.
Moreover, these data support more nuanced analyses of population distributions~\citep{schug2021gridded}, socio-economic conditions~\citep{feldmeyer2020using}, as well as deeper understanding of the impact on human behaviors~\citep{wang2016review} and public perception~\citep{liang2024evaluating}. 
Hence, more comprehensive and openly accessible geospatial data on building can enable the formulation of nuanced urban planning policies, fostering locally informed and globally connected approaches to efficiently support urban resilience and sustainability~\citep{elmqvist_sustainability_2019}.

Traditionally, obtaining building attributes has involved expert evaluation, government records, or crowdsourced labeling, which often require field studies.
This approach limits coverage and efficiency, leaving many buildings without detailed information.
Although platforms like OpenStreetMap (OSM) and government databases now contain diverse urban information, the incompleteness and uneven geographical distribution of global building features hinder their usability across larger regions~\citep{biljecki2023quality, milojevic2023eubucco, lei2023assessing, herfort_spatio-temporal_2023, florio2025ghs}.
With their rapid development, remote sensing-based methods have become a standard approach for extracting building information from aerial and satellite imagery, including attributes such as building height~\citep{wu2023first, frantz2021national}, and types~\citep{du_semantic_2015, zhao_exploring_2019}.
Remote sensing provides broad coverage, reduces reliance on ground surveys, and enables high-resolution tracking of urban changes over time.
In parallel, machine learning methods that leverage geometric and built environment information have been widely applied to enhance the coverage and accuracy of building data~\citep{roy2023inferring, nachtigall_predicting_2023, lei_predicting_2024, wang_multi-view_2024}.
Despite the advancements, the top-down perspective poses inherent challenges, as critical vertical details of structures remain difficult to capture from overhead imagery.

The emergence of easily accessible Street View Imagery (SVI) has transformed the way buildings are analyzed, providing a ground-level, bottom-up perspective that captures architectural details often obscured in aerial or satellite imagery~\citep{biljecki2021street,gaw2022comparing,zhang2024urban}. Leveraging this capability, numerous studies have integrated deep learning with SVI to extract and profile various building attributes, including height~\citep{yan_estimation_2022,fan_pano2geo_2024}, type and usage~\citep{kang_building_2018, zhao2021bounding, ramalingam_automatizing_2023}, architectural style~\citep{lindenthal2021machine, sun2022understanding}, and facade materials~\citep{xu_semantic_2023, raghu_towards_2023, chen_mapping_2024}. 
Beyond building profiling, these integrations also support a range of practical applications, including risk assessment~\citep{pelizari2021automated, wang2021automatic}, refinement of 3D building models~\citep{zhang2021vgi3d}, and building energy efficiency estimation~\citep{sun2022understandingenergy, mayer2023estimating}. 
These advancements have significantly contributed to SVI-based urban studies, enabling fine-grained, large-scale geospatial analyses.

Despite advancements in SVI-based methods for inferring building attributes, various challenges limit scalability and adaptability: (1) existing datasets struggle with uncertainty due to limited angular coverage in perspective views or distortions in panoramic images, hindering comprehensive observations; (2) reliance on proprietary data restricts accessibility, transparency, and adaptability, with ambiguous licensing further limiting research utility and inclusivity~\citep{helbich_use_2024}; (3) while some efforts align visual data with geolocation, annotations often focus on isolated attributes, requiring separate models. Multi-task learning has been explored~\citep{chen_deep_2022}, but class diversity remain constrained, 
limiting the ability to capture architectural complexity for more inclusive and interpretable analyses.
Consequently, SVI-based building datasets offering holistic structural perspectives, fully open data, and comprehensive architectural insights remain scarce.

Vision-language models (VLMs), uniting computer vision (CV) and natural language processing, have demonstrated the ability to interpret complex visual relationships, reason about scenes, and generate semantically rich descriptions~\citep{li_vision-language_2024}.
In the remote sensing domain, vision-language tasks have demonstrated promise for multi-scale feature understanding, multi-task learning, and applications such as visual question answering, image captioning, and semantic segmentation~\citep{zia2022transforming, hu2025rsgpt, dong_changeclip_2024, wang_earthvqanet_2024}.
More recently, multimodal large language models (MLLMs) have advanced these capabilities by integrating deep contextual and semantic representations learned from massive, multimodal datasets, thereby enabling more nuanced and precise interpretations of visual data.
This versatility highlights their potential to serve as foundational instruments in SVI-based building research, by enhancing the characterization of building properties, streamlining multi-task learning, and transcending predefined label sets in the analysis of facade features.

To advance fine-grained, bottom-up observations of buildings, we propose an open framework, OpenFACADES, that enriches a variety of building properties from a street-level perspective by leveraging multimodal crowdsourced inputs and open-source MLLMs.
First, we utilize open-source building footprints and SVI to perform visibility simulations that geospatially align building geometries with corresponding SVI shooting locations.
Second, we introduce an innovative pipeline that detects individual buildings based on their visible angles and acquires holistic building images using a custom image reprojection method.
Third, we assemble one of the largest global, multi-attribute building image datasets by combining crowdsourced building attributes with high-quality text descriptions generated by state-of-the-art MLLMs.
Leveraging this dataset, we are among the first to introduce tailored MLLMs for building profiling through multi-task learning, encompassing both single- and multi-attribute prediction tasks as well as open-vocabulary captioning.
Furthermore, we present an in-depth comparative analysis of model performance across various hyperparameter settings, cross-city generalization scenarios, and image quality variations.

The primary contributions of this work are threefold:
\begin{itemize}[topsep=5pt,itemsep=-0.9ex,partopsep=-1ex,parsep=1ex]
\item Developed a reproducible methodology that (1) geolocates, detects, and acquires holistic building images from crowdsourced SVI; (2) integrates these images with crowdsourced building data to create an open and structured building image dataset; and (3) enables future scalability by dynamically retrieving the latest available data from these sources.

\item Compiled an open global building dataset, consisting of (1) 31,180 individual building images from seven cities across three continents, annotated with attribute labels from OSM and text descriptions generated by ChatGPT-4o; and (2) large-scale automated annotations on 1.2 million images covering over half a million buildings. Each image is linked to its geospatial location and enriched with diverse attributes (e.g., building type, number of floors, age, and surface material) along with detailed textual descriptions. This forms the OpenFACADES dataset, one of the largest such resources, spanning multiple urban morphologies.

\item Introduced the first benchmark open-source MLLMs that (1) perform multi-attribute prediction on buildings, achieving robust and more accurate image labeling performance than zero-shot ChatGPT-4o; (2) generate descriptive captions on architectural features, providing comprehensive information beyond standard building attributes; and (3) demonstrate enhanced robustness and generalizability relative to prior CV models.
\end{itemize}

In summary, this work presents a comprehensive and reproducible framework that leverages multimodal crowdsourced data to develop a global street-level building dataset for training multimodal models. 
This approach enhances the scope, adaptability, and accuracy of urban analysis, enabling more detailed and interpretable assessments of the built environment.

\section{Related work}\label{sc_lr}
\subsection{Existing street-level building datasets}
With advances in geospatial artificial intelligence technologies, research in recent years has increasingly leveraged remote sensing datasets such as high-resolution satellite and aerial imagery, and LiDAR to enhance urban applications. 
These datasets enable object-based image analysis, pixel-based classification, and semantic segmentation of urban structures, providing critical insights for land use mapping, urban morphology analysis, and spatiotemporal change detection. 
As key urban components, buildings have spurred the creation of domain-specific datasets and methodologies to support applications such as urban sustainability evaluation through rooftop attributes extraction \citep{wu2021roofpedia}, infrastructure management via automated land cover classification \citep{boguszewski2021landcover}, and disaster management through assessing damage \citep{gupta2019creating, li_cross-view_2025}.

SVI, rapidly emerging as a prominent proximal remote sensing data source, has been leveraged to generate spatially enriched urban datasets that facilitate fine-grained semantic understanding of complex urban scenes~\citep{biljecki2021street}.
Among these, building-centric SVI datasets enable facade-level feature extraction, offering images that capture textural, material, and architectural features of building exteriors for environmental modeling.
For example, building age and architectural style have long been studied for their links to building thermal performance \citep{tooke2014predicting, aksoezen2015building, nouvel2017influence} and real estate pricing \citep{zietz2008determinants, lindenthal2021machine}.
Recent advances include the work of \citet{sun2022understanding}, which applies deep convolutional neural networks (CNNs) to classify buildings in Amsterdam, the Netherlands, into architectural periodization categories (e.g., revival, postwar).
Material characterization~\citep{xu_semantic_2023, chen_mapping_2024}, another aspect critical for building energy simulation \citep{nouvel2017influence}, also supports circular economy objectives by enabling lifecycle material tracking~\citep{raghu_towards_2023} and risk assessment~\citep{wang2021automatic}. 
Among these efforts, \citet{raghu_towards_2023} employ a multi-city material categories (brick, stucco, etc.) using geotagged SVI perspective views, aligning visual patterns with ground-truth material information for scalable building classification.
Combining the aspects of building age and material, \citet{ogawa2023deep} introduced a method to detect and geolocate buildings from panoramic images, automatically annotating them with objective building data in Kobe, Japan. 

Furthermore, building type or usage, a critical attribute in urban remote sensing and land use classification, is also central to street-level research \citep{kang_building_2018, zhao2021bounding, lindenthal2021machine, ramalingam_automatizing_2023, li_fine-grained_2025}.
A seminal work by \citet{kang_building_2018} introduces the BIC\_GSV dataset, a multi-city geospatial database of 19,658 SVI-derived building facades categorized into eight classes (e.g., apartment, church, garage, etc.) across North America.
These ground truth labels are generated through view-direction-aligned spatial joins with OSM building footprints, enabling parcel-scale urban pattern analysis.
Advancing this, \citet{zhao2021bounding} developed the BEAUTY dataset, which extends BIC\_GSV by incorporating both SVI-based land use classification (e.g., residential, commercial, etc.) and multi-class building detection.
Other similar research frameworks have also been applied to large-scale urban studies, integrating additional building attributes such as floor number estimation, abandoned house detection, and seismic risk assessment~\citep{iannelli2017extensive, zou_detecting_2021, rosenfelder2021predicting, pelizari2021automated, ghione2022building}. 
These workflows not only enable location-based building retrieval but also demonstrate cross-modal alignment of SVI with open geospatial building footprints.

However, several challenges still remain in street-level building research, limiting the scalability and adaptability of current approaches.
First, although many efforts have aligned visual information with building geolocation~\citep{kang_building_2018, sun2022understanding, ogawa2023deep}, they are often either reliant on perspective views with restricted angular coverage, limiting visibility of upper building elements, or on panoramic images prone to severe distortions, misaligning with actual observations.
Second, while various SVI-based building datasets have been established, their dependence on data derived from proprietary platforms introduces limitations related to accessibility, transparency, and adaptability. 
The ambiguous licensing terms of such datasets further constrain their utility for diverse research applications and compromise the integrity of work built upon them, thereby hindering inclusivity within the research community~\citep{helbich_use_2024}.
In a recent trend, crowdsourced SVI platforms have garnered attention in urban studies by producing diverse, publicly accessible imagery.
Examples include annotating points of interest~\citep{zarbakhsh_points--interest_2023}, image status~\citep{hou_global_2024}, human perception~\citep{yang2025streetscape}, and road surface type~\citep{kapp_streetsurfacevis_2025}. 
Among these, \citet{hou_global_2024} curate a manually labeled dataset to assess 10 million crowdsourced SVIs from 688 cities, enriched with metadata such as platform, weather, and lighting conditions, while \citet{kapp_streetsurfacevis_2025} utilize OSM tags and ChatGPT-4o to label and amplify underrepresented road surface classes, resulting in 9,122 labeled images. 
These initiatives illustrate the potential of crowdsourced data for broad, inclusive urban analyses.

\subsection{Vision models in urban analytics}\label{sc_lr_2}
With the rapid development of deep learning techniques over the past decade, diverse methods have been developed to extract urban cues from visual information.
In terms of building facade research, in particular, CNNs have been widely employed due to their strong feature representation capabilities. 
Among them, VGG, DenseNet, and ResNet have been extensively applied to achieve, or serve as benchmarks for, the accurate classification and evaluation of building functions~\citep{kang_building_2018}, materials~\citep{ghione2022building, raghu_towards_2023}, architectural styles~\citep{lindenthal2021machine, sun2022understanding, ogawa2023deep}, and human perceptions~\citep{liang2024evaluating}.
Additionally, Vision Transformers (ViTs) have emerged as powerful alternatives, leveraging self-attention mechanisms to capture long-range dependencies in building images.
Recent studies have demonstrated the effectiveness of ViTs in urban analytics, achieving state-of-the-art performance in material recognition, and construction period prediction~\citep{raghu_towards_2023, ogawa2023deep}.
Beyond that, hybrid models combining various model backbones have been further developed to consider multi-dimensional features as input, improving comprehensiveness and generalizability in multi-scale urban analysis~\citep{huang2023comprehensive, jia2024transformer, fujiwara2024microclimate}.

However, the annotation of building attributes remains a fundamental limitation in these approaches. 
Labels are often restricted to isolated attributes, such as building type or material, necessitating the training of separate models for different objectives. 
While multi-task learning frameworks have been explored~\citep{chen_deep_2022}, class diversity and model scalability remain constrained. Moreover, annotation schemes are typically predefined and rigid due to the availability of data, preventing adaptation to unannotated or emergent building characteristics, such as mixed-use functions or hybrid architectural materials.
This lack of multi-dimensional, context-aware labels significantly limits the ability to capture architectural complexity, hindering the development of comprehensive, inclusive, and interpretable approaches for building analysis. 

Rapid advancements in LLMs offer new avenues for extracting nuanced insights about complex urban environments. 
Notably, VLMs combine visual and linguistic modalities, leveraging deep semantic reasoning to establish rich connections between visual concepts and textual descriptions~\citep{wu_mixed_2023, li_vision-language_2024}. 
Building on these capabilities, recent work in remote sensing demonstrates how VLMs can exceed traditional CV methods by producing more context-aware and human-like interpretations~\citep{al2022open, zia2022transforming, hu2025rsgpt}, thereby providing not only precise visual recognition but also a semantic understanding of objects and their relationships within complex environments.
In terms of street-level building research, recent studies have explored the state-of-the-art models for automated building annotation. 
For example, \citet{li2024buildingview} employed ChatGPT-4o to generate structured multi-label annotations for buildings using SVIs across multiple cities. Similarly, \citet{zeng2024zero} assessed the model’s performance in zero-shot building age prediction, finding that ChatGPT-4 effectively estimates the construction period of buildings.
However, deploying proprietary LLMs such as ChatGPT-4o at scale presents limitations. 
Model inference relies on API-based access, which incurs high computational costs, making large-scale applications financially and computationally restrictive, which also constrains the efficiency for fine-tuning, limiting their adaptability for domain-specific urban studies.
To address these challenges, recent open-source initiatives have produced diverse series of LLMs, including Qwen-VL~\citep{wang2024qwen2}, Llama~\citep{dubey2024llama}, and InternVL~\citep{chen2024expanding}, enabling greater customization and efficiency in downstream tasks.
These models exhibit unified capabilities to process multi-dimensional inputs, generating context-aware descriptions informed by their pretraining on large-scale, diverse datasets. 
This capability holds significant potential for advancing street-level urban analysis, as their ability to interpret human-centric observations closely aligns with how individuals perceive and contextualize the built environment.

Hence, we propose a reproducible methodology for integrating open-source multimodal building data from global cities into a comprehensive dataset, incorporating objective attributes and detailed captions. Table~\ref{tb:dataset} provides an overview of existing SVI datasets related to building attributes, highlighting how our contribution addresses current limitations while significantly expanding the scale, scope, and dimensionality of SVI-based datasets for building-related research.
This advancement not only enhances the accessibility and adaptability of building datasets but also paves the way for broader, more inclusive, and scalable applications in urban analytics.

\begin{table}[!htb]
\caption{Characteristics of existing SVI-based datasets constructed for building-oriented CV and urban research applications, and the features of the dataset we established in this research (GSV: Google Street View).}
\centering
\renewcommand{\arraystretch}{1.5} %
\makebox[\textwidth]{ %
\resizebox{1.25\textwidth}{!}{%
\begin{tabular}{
  >{\centering\arraybackslash}m{2.8cm} %
  >{\centering\arraybackslash}m{0.1cm} %
  >{\centering\arraybackslash}m{2.0cm} %
  >{\centering\arraybackslash}m{3cm} %
  >{\centering\arraybackslash}m{0.1cm} %
  >{\centering\arraybackslash}m{1.5cm} %
  >{\centering\arraybackslash}m{2.0cm} %
  >{\centering\arraybackslash}m{0.1cm} %
  >{\centering\arraybackslash}m{1.5cm} %
  >{\centering\arraybackslash}m{1.5cm} %
  >{\centering\arraybackslash}m{2.0cm} %
  >{\centering\arraybackslash}m{0.1cm} %
  >{\raggedright\arraybackslash}m{7.5cm} %
}
\toprule[1.5pt]
\textbf{Studies} & &
  \multicolumn{2}{c}{\textbf{Purpose}} & &
  \multicolumn{2}{c}{\textbf{Lineage}} & &
  \multicolumn{3}{c}{\textbf{Coverage}} & &
  \textbf{Category} \\ 
\midrule[1.5pt]
 & & Task & Building attribute & & Image source & Image type & & No. of labeled images & No. of cities & Continent(s) & \\ 
\cline{1-1} \cline{3-4} \cline{6-7} \cline{9-11} \cline{13-13}

BIC\_GSV \citep{kang_building_2018} &
  & image classification &
  type &
  & GSV 
  & perspective &
  & 19,658 &
  More than 30
  & North America & &
  apartment, church, garage, house, industrial,
  office building, retail, roof (8 categories) \\

BEAUTY \citep{zhao2021bounding} &
  & image classification and object detection &
  type &
  & GSV 
  & perspective &
  & 19,070 &
  More than 30
  & North America & &
  \textit{Image classification}: residential, commercial, public, industrial (4 categories);\newline
  \textit{Multi-class detection}: apartment, church, garage, house, industrial, office building, retail, roof (8 categories).\\

\citet{lindenthal2021machine} &
  & image classification &
  age &
  & GSV 
  & perspective &
  & 29,177 &
  1
  & Europe & &
  Georgian, early Victorian, late Victorian/Edwardian,
  interwar, postwar, contemporary, revival (7 categories). \\

\citet{raghu_towards_2023} &
  & image classification &
  surface material &
  & GSV 
  & perspective &
  & 985 &
  3
  & Asia, North America, Europe & &
  brick, stucco, rustication, siding, wood, metal, other (7 categories) \\

SVI4BuildingFunc \citep{li_fine-grained_2025} &
  & object detection &
  type &
  & GSV 
  & panoramic &
  & 15,400 &
  4
  & North America, Europe & &
  varies by city (e.g., high residential, low residential, commercial, office, walk-up buildings, mixed-up buildings; 5 to 6 categories per city) \\
  
\hline

OpenFACADES &
  & Image labeling and captioning &
  type, age, floor, surface material, feature description &
  & Mapillary 
  & individual building images &
  & 31,180 &
  7
  & North America, Europe, Asia & & 
  \textit{Type}: apartments,
house, retail, office, hotel, industrial, religious, education, public, garage (10 categories); \newline
  \textit{Surface material}: metal, glass, brick, stone, concrete, wood, plaster (7 categories); \newline
  \textit{Age}: numeric value; \newline
  \textit{Floor}: numeric value.
 \\
\bottomrule[1.5pt]
\end{tabular}%
}}
\label{tb:dataset}
\end{table}

\section{Methodology}
\begin{figure}[ht]
    \centering
    \includegraphics[width=.9\linewidth]{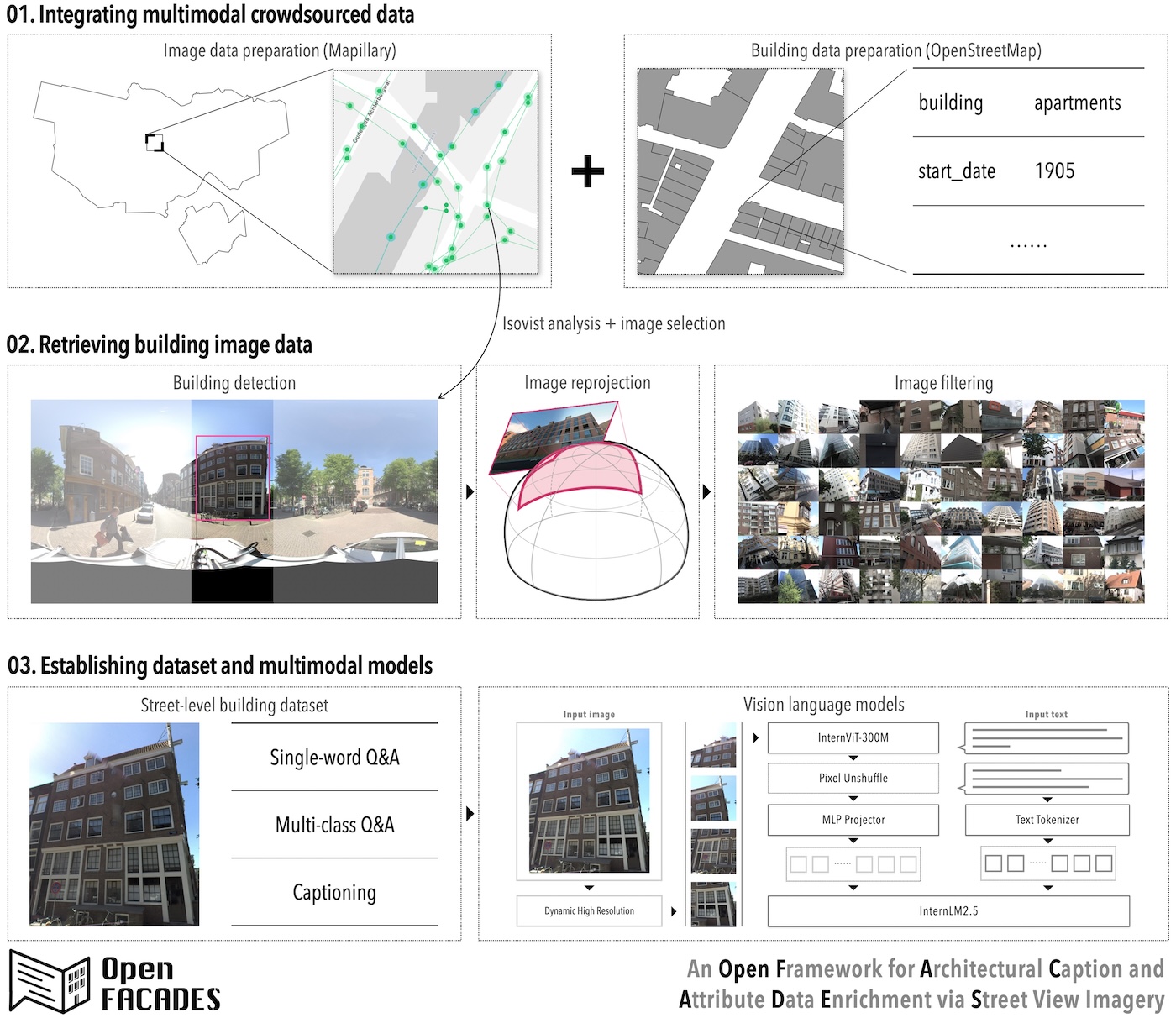}
    \caption{General framework for integrating multimodal crowdsourced data to establish a street-level building dataset and develop multimodal models. Data: (c) Mapillary and OpenStreetMap contributors.}
    \label{fig_workflow}
\end{figure}

In this study, we introduce OpenFACADES, a comprehensive framework for acquiring building images from SVI and automatically annotating them with crowdsourced data. This framework facilitates the development of large multimodal models tailored for architectural attributes question-answering and captioning. The framework is structured into three main steps, as illustrated in Figure~\ref{fig_workflow}:

(1) Integrating multimodal crowdsourced data.
Initially, crowdsourced SVI metadata and building data are collected for research areas. Then, isovist analysis is performed to simulate the theoretical angles of view (AOV) from each camera location to the target structures. 
SVIs with optimal visibility are then retrieved and filtered by image features to ensure that only high-quality candidates are retained for subsequent analysis.

(2) Retrieving building image data. 
Based on the geospatial AOVs simulated, we map the relative viewing angles and detect target buildings within the image space. This process enables us to precisely identify associate building information with their visual representations. Then, based on the coordinates of bounding boxes, building images are reprojected from panoramic to perspective view, generating holistic building images. These images further undergo a filtering process to identify high-quality and suitable building views.

(3) Establishing dataset and multimodal models. 
Building images with available crowdsourced data form a dataset with four label types: single-attribute label, single-attribute Q\&A, multi-attribute Q\&A, and captioning. Single-attribute labels are derived from building information, while single-attribute Q\&A append those attribute labels to targeted questions, generating concise question-to-label pairs.
Multi-attribute Q\&A and captioning labels are generated using ChatGPT-4o, enabling detailed textual descriptions and structured annotations for comprehensive building attribute analysis.
The last three label types are utilized to fine-tune vision-language models, enabling a versatile model for multi-attribute building labeling and captioning with enhanced contextual understanding.

\subsection{Integrating multimodal crowdsourced data}\label{method_1}
\begin{figure}[ht]
    \centering
    \makebox[\textwidth]{%
        \includegraphics[width=1.2\linewidth]{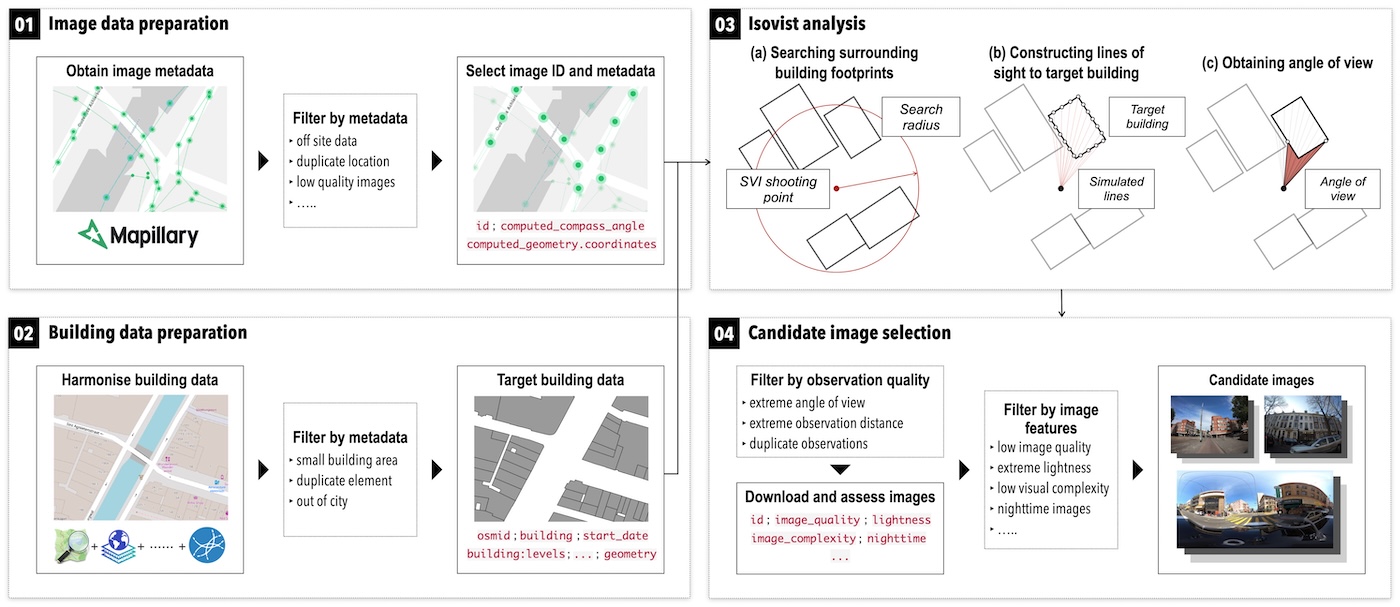}
    }
    \caption{Workflow for obtaining and integrating suitable multimodal crowdsourced data, combining street-level imagery from Mapillary and building information from OpenStreetMap, along with external sources such as Overture Maps and government data, to harmonize building dataset. Data: (c) Mapillary and OpenStreetMap contributors.}
    \label{fig_aov}
\end{figure}

The workflow of integrating multimodal crowdsourced data for building analysis is illustrated in Figure~\ref{fig_aov}. The process includes: 

\paragraph{Image data preparation.}
At the first stage, the raw metadata of street-level image data from crowdsourced platform is obtained within study areas before requesting the images. 
Here, Mapillary is chosen for its extensive global coverage, high-quality user-generated content, and open-access policies that enable reproducible and scalable urban research~\citep{hou_comprehensive_2022,kapp_streetsurfacevis_2025,danish2025citizen}.
Specifically, the metadata, comprising image type (\colorbox{lightgray}{\texttt{is\_pano}}), location coordinates (\colorbox{lightgray}{\texttt{computed\_geometry}}), compass angle \\ (\colorbox{lightgray}{\texttt{computed\_compass\_angle}}), capture time (\colorbox{lightgray}{\texttt{captured\_at}}), and quality indicator (\colorbox{lightgray}{\texttt{quality\_score}}), is utilized to structure sorting and quality assessments. 
Filtering operations select panoramic imagery (\colorbox{lightgray}{\texttt{is\_pano=True}}), remove images captured outside the defined study area, exclude multiple images from the same spatial point to prevent redundant viewpoints, and discard those with poor resolution or quality defects. 
The output of this phase is a curated set of image metadata, with their corresponding unique image IDs, coordinates, and compass angles, prepared for subsequent spatial analyses.

\paragraph{Building data preparation.}
In parallel, building geometries and associated metadata are retrieved from OpenStreetMap (OSM)\footnote{\url{https://openstreetmap.org/}}. 
Data harmonization is then conducted to supplement missing building footprints and insufficient building attributes from other data sources, such as Overture Maps\footnote{\url{https://overturemaps.org/}} and government datasets.
Attributes include unique identifiers, building type, facade material, number of floors, construction dates, and polygon geometries. 
Inconsistencies and outliers, such as footprints representing insignificant or extraneous structures (e.g., roof and underground structures), duplicates introduced by overlapping contributions, or buildings located outside the target region, are systematically removed. 
The remaining dataset delivers a precise, consistent, and high-quality representation of the built environment, ready for geometric calculations and alignment with the image data.

\paragraph{Isovist analysis.}
With both image and building datasets prepared, isovist analysis is applied to compute theoretical AOV from each camera location to the target structures, building on previous studies~\citep{lindenthal2021machine, ogawa2023deep, fan2025coverage}.
This analysis identifies each building’s perimeter segments that fall within the camera’s potential field of view and evaluates the observation efficiency of buildings from specific vantage points.
First, a search radius (e.g., 50 meter) is established to identify surrounding buildings from the SVI capture points.
Second, sampling points are generated along the polygonal geometries of buildings within the distance threshold, and lines of sight are constructed towards all sampled points of the target buildings.
Third, lines of sight intersecting with surrounding building footprints are filtered out, leaving only the largest angular span between the unobstructed lines, which represents the AOV to a building from a given image shooting point. 
Additionally, the left and right boundaries of the AOV are recorded as azimuth angles relative to the true north, providing detailed spatial orientation for subsequent tasks.
This process identifies which buildings are potentially visible from each image capture point, thereby aligning the building information with the corresponding imagery metadata.

\paragraph{Candidate image selection.}
Based on the theoretical visibility of buildings, the final stage identifies candidate images most likely to provide reliable observations. 
Criteria derived from the absolute AOV values are used to eliminate images taken from excessive distances, extreme viewing angles, or redundant perspectives.
The remained imagery IDs are then used to retrieve image data from Mapillary.
Given that crowdsourced SVI can vary in quality and may contain various errors~\citep{hou_comprehensive_2022}, these images are further assessed using quality metrics such as brightness, sharpness, and visual complexity to determine their suitability for inclusion.
Images captured under unsuitable conditions (e.g., nighttime, severe overexposure) or containing excessive visual clutter are removed based on the CV models released in NUS Global Streetscapes~\citep{hou_global_2024}.
The result is a curated set of candidate street-level images, optimized for integration with building data in subsequent object detection workflows.

\subsection{Retrieving building image data}\label{method_2}
\begin{figure}
    \centering
    \makebox[\textwidth]{%
        \includegraphics[width=1.2\linewidth]{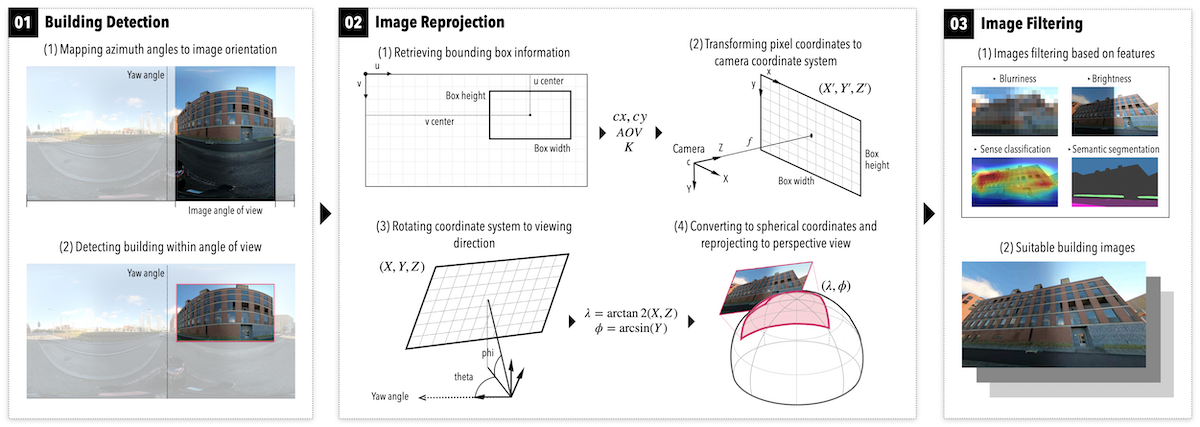}
    }
    \caption{Pipeline demonstrating the extraction and selection of building images from street-level imagery, involving object detection, pixel coordinate transformation and reprojection, and feature-based filtering. Data: (c) Mapillary contributors.}
    \label{fig_detect}
\end{figure}

Figure~\ref{fig_detect} demonstrates the pipeline for extracting and selecting building images from street-level imagery. The process consists of three main steps: Building detection, image reprojection and image filtering.

\paragraph{Building detection.}
Azimuth angles derived from isovist analysis are first used to map the relative viewing angles of a building within the image space. 
This conversion defines a focused AOV for the target building before applying object detection. 
To determine the position of buildings within panoramic imagery, their relative horizontal coordinate ratios are computed as follows:

\begin{equation}
\quad P^{n,i}_{\{l,r\}} = \frac{(A^{n,i}_{\{l,r\}} - H^i + C) \mod 360}{360}
\end{equation}

\noindent where \(P\), which ranges from 0 to 1, represents the left (\(l\)) or right (\(r\)) horizontal coordinates ratio of building \(n\) in the panoramic image \(i\). The term \(H\) denotes the yaw angle when the SVI image token, and \(C\) is an adjustable calibration constant that ranges from (0-360), depending on the part of the image the view is oriented towards. Typically, \(C\) is set to 180 in Mapillary, indicating that the center of the image is the focal direction. 

After determining the relative position of buildings in the SVI, images are cropped using the calculated horizontal coordinate ratios to isolate the AOV focused on the target buildings.
Within the focused view, object detection is performed to identify buildings.
To accomplish this, we employ GroundingDINO, a model equipped with pre-trained weights capable of detecting various objects using human inputs such as category names or referring expressions~\citep{liu2023grounding}.
Specifically, we use the ``GroundingDINO-B'' model checkpoint, which is trained on several widely-recognized object detection datasets, including COCO, O365, and OpenImage.
By assigning the category name ``building'' to this open-set detector, we generate bounding boxes around the buildings in each cropped image area. 
This process constrains the observation area to focus on each building footprint, enabling the association of visual observations with 2D building geometries. 
Additionally, it facilitates the object detection model in isolating target buildings from surrounding elements, such as adjacent structures and environmental noise.

\paragraph{Image reprojection.}
Panoramic images are formed by mapping the 3D environment onto a 2D sphere, which causes straight lines and familiar shapes to appear curved or distorted.
After retrieving the bounding box information from the object detection, the reprojection process is designed to correct these inherent distortions.
The objective of the reprojection is to take the portion of the panoramic image identified by the bounding box and present it as if it were photographed by a standard pinhole camera, providing a more intuitive and distortion-free representation of the detected object.

First, we interpret the bounding box region in terms of pixel coordinates within the panoramic imagery, obtaining the box center as ($c_u$, $c_v$), along with its $width$ and $height$, which are essential for subsequent tasks.
Second, a virtual pinhole camera model is constructed based on the specified AOV to a target building and the bounding box $width$. 
The focal length $f$ and principal point ($c_x$, $c_y$) in camera coordinate are computed as:

\begin{equation}
f = \frac{\frac{width}{2}}{\tan\left(\frac{AOV}{2} \cdot \frac{\pi}{180}\right)}
\end{equation}

\begin{equation}
c_x = \frac{width - 1}{2}, \quad c_y = \frac{height - 1}{2}
\end{equation}

These values are used to construct the intrinsic camera matrix $K$, which encapsulates the intrinsic parameters of the virtual pinhole camera. For each pixel ($x$,$y$) in the virtual panel, the transformation from the 2D pixel location to a 3D direction in the camera’s coordinate system is achieved by applying the inverse of $K$:

\begin{equation}
K = \begin{bmatrix}
f & 0 & c_x \\
0 & f & c_y \\
0 & 0 & 1
\end{bmatrix}, \begin{bmatrix} 
X{\prime} \\ Y{\prime} \\ Z{\prime} \end{bmatrix} = K^{-1}\begin{bmatrix} x \\ y \\ 1 \end{bmatrix}
\end{equation}

\noindent where resulting vector $\mathbf{v}_{cam} = (X{\prime}, Y{\prime}, Z{\prime})^T$ represents the direction of a ray emanating from the camera center through the corresponding pixel on the virtual image plane.

Third, to determine the approximate view direction of the bounding box, we use the center coordinates ($c_u$, $c_v$) of the bounding box in panoramic coordinate system and combined rotation matrix $R$ to align the camera’s direction to the rotated direction in 3D space:

\begin{equation}
\theta = \left( c_u - 0.5 \right) \cdot 360, \quad
\phi = \left( 0.5 - c_v \right) \cdot 180
\end{equation}

\begin{equation}
R = R_x(\phi) R_y(\theta)
\end{equation}

\begin{equation}
\mathbf{v}_{rot} = R \mathbf{v}_{cam}
\end{equation}

\noindent where $c_u$ and $c_v$ are normalized to a range of [0, 1], with $c_u$ as the horizontal center and $c_v$ as the vertical center of the bounding box region.
The yaw angle $\theta$ defines the horizontal rotation of the camera and spans from $-180^\circ$ to $180^\circ$. 
The pitch angle $\phi$ defines the vertical rotation of the camera and spans from $-90^\circ$ to $90^\circ$.
The combined rotation matrix $R$ is formed as the product of two individual rotation matrices based on Rodrigues' formula: $R_y$($\theta$), which rotates the coordinate system around the y-axis (yaw), and $R_x$($\phi$), which rotates the coordinate system around the x-axis (pitch). 
$\mathbf{v}_{cam}$ is the original direction vector in the camera’s coordinate system, while $\mathbf{v}_{rot}$ is the new direction vector after applying the rotations, pointing toward the desired region of the spherical panorama.

Lastly, the rotated 3D direction vector $\mathbf{v}_{rot} = (X, Y, Z)$ is normalized and converted into spherical coordinates, where longitude $\lambda$ and latitude $\varphi$ are calculated based on:

\begin{equation}
\lambda = \arctan2(X, Z), \quad \varphi = \arcsin(Y)
\end{equation}

\noindent The corresponding pixel coordinates ($X_{\text{img}}$, $Y_{\text{img}}$) in the original panoramic image (equirectangular format) are then derived as:

\begin{equation}
X_{img} = \left( \frac{\lambda}{2\pi} + 0.5 \right)(W_{\text{pano}} - 1), \quad
Y_{img} = \left( \frac{\varphi}{\pi} + 0.5 \right)(H_{\text{pano}} - 1).
\end{equation}

\begin{figure}[ht]
    \centering
    \makebox[\textwidth]{%
        \includegraphics[width=0.85\linewidth]{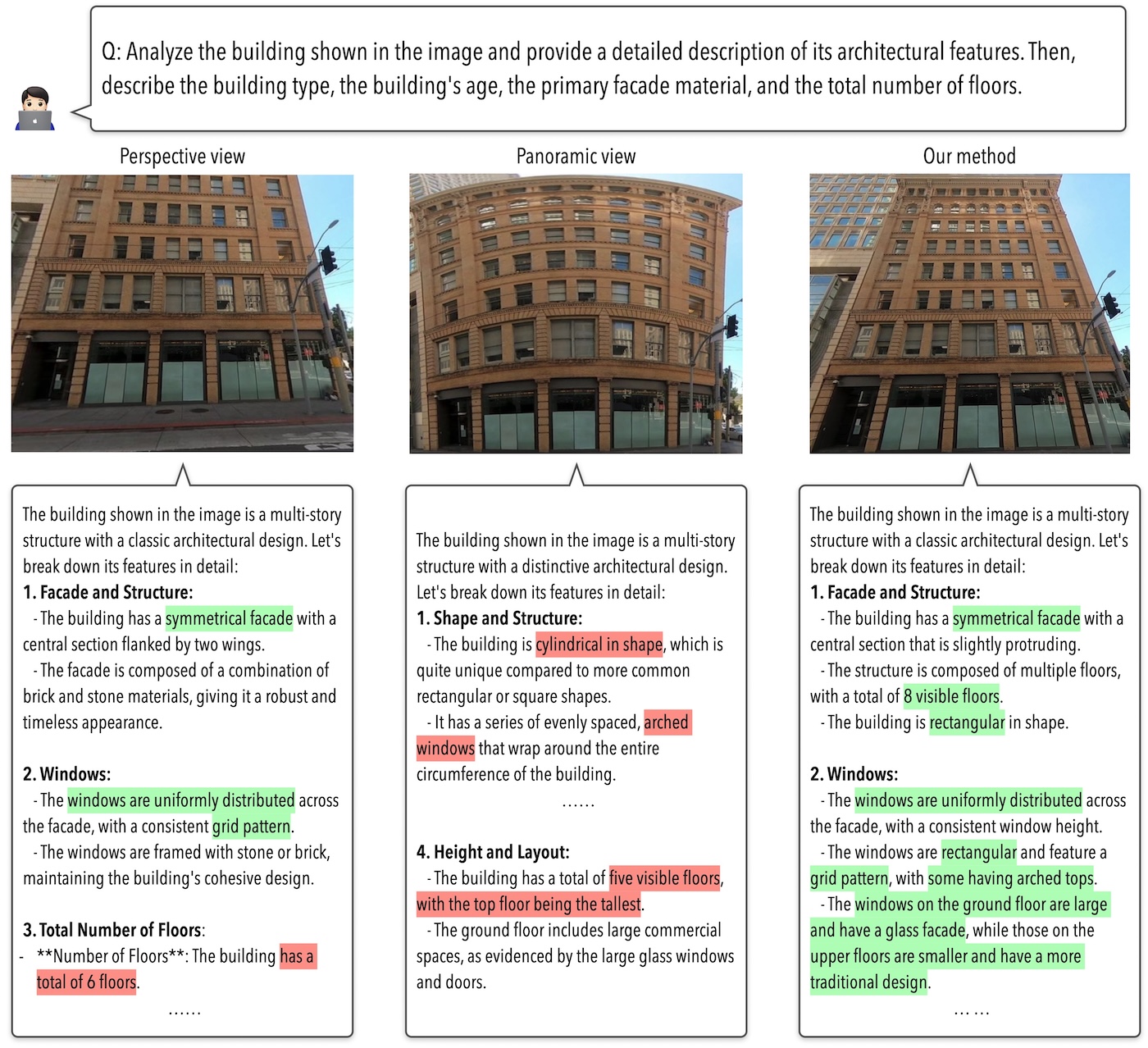}
    }
    \caption{Examples of different types of building images used as input to the vision-language model, resulting in varied responses. By generating a holistic view of individual buildings, our method facilitates a more authentic analysis and interpretation. Data: (c) Mapillary contributors.}
    \label{fig_repro_exam}
\end{figure}

\noindent At these coordinates, pixel values are sampled from the original panoramic image, and reprojected to generate the rectified perspective view using the \colorbox{lightgray}{\texttt{remap}} function from OpenCV library.
This transformation eliminates the spherical distortions inherent in panoramic imagery, producing a visually intuitive and geometrically corrected view aligned with the detected object. As examples demonstrated in Figure~\ref{fig_repro_exam}, this correction is crucial not only for preserving essential structural details for model interpretation but also for mitigating distortions that could otherwise misalign architectural features. This preprocessing step enhances the model’s ability to accurately analyze building attributes.

\paragraph{Image filtering.}
The features of the detected individual building images are further analyzed to refine and enhance the image dataset.
ZenSVI~\citep{ito2025zensvi}, an open-source library for street-level imagery analysis, is integrated into the framework to facilitate the extraction of image features.
We analyze image features across four key dimensions: blurriness, brightness, semantic segmentation, and scene classification. 
These dimensions are utilized to identify high-quality and suitable building images for inclusion in the dataset. Blurriness is evaluated using the OpenCV Laplacian operator to filter out images with motion blur or poor focus, while brightness assessment removes those with suboptimal illumination. A pre-trained Place365 model~\citep{zhou2017places} excludes indoor scenes, and semantic segmentation is applied to detect and minimize occlusions (e.g., trees, vehicles, walls), ensuring that selected images predominantly showcase building facades and maintain high visual quality.

\subsection{Establishing dataset and multimodal models}\label{method_3}

\paragraph{Street-level building dataset.}
Following the previous process, building information is assigned to the detected buildings in the imagery. In this study, we focus on building type, facade material, construction year (age), and number of floors, which have been identified as essential attributes in prior studies and are supported by relatively sufficient data for model training and evaluation.
Specifically, we utilize labels from building data corresponding to the categories: \colorbox{lightgray}{\texttt{building}}, \colorbox{lightgray}{\texttt{building:material}}, \colorbox{lightgray}{\texttt{start\_date}}, and \colorbox{lightgray}{\texttt{building:levels}} in OSM alongside supplementary datasets, as mentioned in Section~\ref{method_1}.
Here, building type and facade material are treated as categorical variables, while construction year and number of floors are represented as numerical values.

From the full set of building data, we sample buildings with available category labels to construct the dataset for subsequent model development.
The dataset is assembled and divided into training, validation, and test sets based on the following three principles: (1) ensuring sufficient labels across all classes to avoid biased predictive accuracy; (2) maintaining a balanced geospatial distribution across cities to represent the diversity of architectural designs; and (3) preventing the same building from appearing in both different sets to minimize data leakage.

\begin{figure}[ht]
    \centering
    \makebox[\textwidth]{%
        \includegraphics[width=1.2\linewidth]{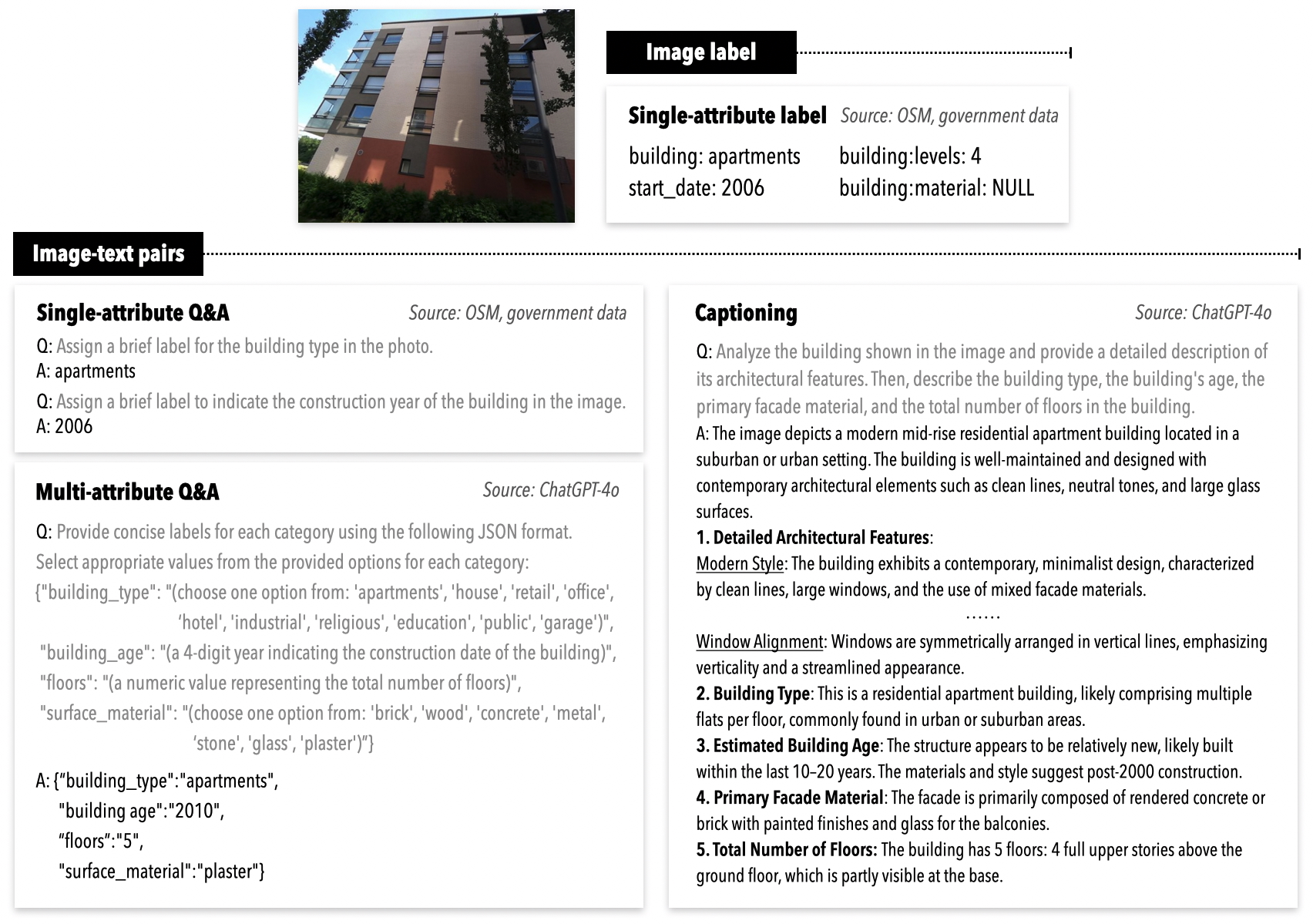}
    }
    \caption{Different label types and data collection approaches for developing a street-level building dataset. Data: (c) OSM and Mapillary contributors.}
    \label{fig_label_type}
\end{figure}

The dataset contains four types of labels: single-attribute label, single-attribute Q\&A, multi-attribute Q\&A, and captioning labels.
Single-attribute labels are used to fine-tune CV models, serving as the baseline for evaluating the performance of common practices. 
Single-attribute Q\&A labels are derived from those single-attribute labels by appending the label to specific questions about the four building attributes, thereby generating concise question-to-label pairs based on building information.
Multi-attribute Q\&A and captioning labels are generated using the state-of-the-art multimodal large language model, ChatGPT-4o , through the OpenAI API\footnote{\url{https://www.openai.com/}}. This task involves prompting the model to annotate or describe the building features visible in the images, thereby creating an image-text dataset. Figure~\ref{fig_label_type} provides detailed indication of data sources and examples of these labels.

Among these labels, the single- and multi-attribute Q\&A outputs share a consistent, structured format constrained by predefined vocabulary, whereas the captioning task provides a free-form textual description that often embeds the same attributes in more expressive language, forming a hierarchical relationship with the Q\&A formats.
For example, a building labeled as ``brick'' (material), ``1920'' (age), ``3'' (floors), and ``house'' (type) in the Q\&A tasks might be described in the captioning output as ``a three-story house with visible brickwork, built in the early 20th century'', reflecting increased richness in narrative form.
Inconsistencies can arise between OSM ground truth labels and ChatGPT-generated labels that slight discrepancies (e.g., differing building floor or age estimates) occur when ChatGPT assigns different labels to buildings with ambiguous features.

\begin{figure}[ht]
    \centering
    \makebox[\textwidth]{%
        \includegraphics[width=1.2\linewidth]{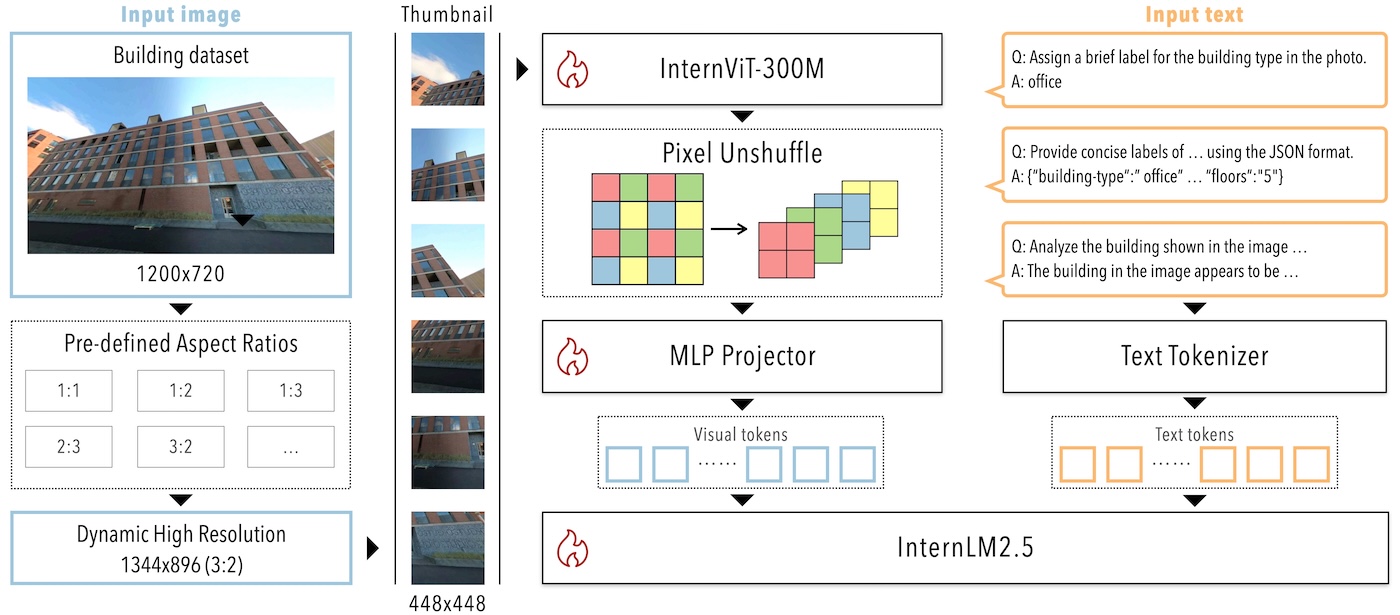}
    }
    \caption{The overall framework of the InternVL series model architecture for building-centric tasks. Data: (c) Mapillary contributors.}
    \label{fig_vlm_model}
\end{figure}

\paragraph{Vision-language models.}
To address the limitations of traditional CV models in building attribute analysis, we leverage InternVL3~\citep{zhu2025internvl3}, an open-source MLLM designed for unified visual-language reasoning. As depicted in Figure~\ref{fig_vlm_model}, InternVL3 is built on the ``ViT-MLP-LLM'' paradigm by integrating a scalable vision encoder (InternViT)~\citep{chen2024internvl}, a multi-layer perceptron (MLP) projector, and a large language model (LLM). The vision encoder is InternViT-300M-448px-V2.5, a distilled variant of the 6B-parameter model optimized via dynamic high-resolution training and next token prediction (NTP) loss \citep{chen2024expanding}. This architecture processes 448×448 pixel image tiles through a pixel unshuffle operation, reducing 1024 visual tokens to 256 for efficient cross-modal alignment.

The model is selected for its general-purpose captioning and open-vocabulary classification capabilities, critical for capturing the multifaceted attributes of buildings (e.g., material, style, type) within a unified framework. 
Unlike conventional models restricted to predefined labels, InternVL’s contrastive vision-language pretraining enables semantic reasoning over diverse facade characteristics, aligning with our goal of holistic building profiling. 
Full-model tuning is conducted through optimizing three components (Figure~\ref{fig_vlm_model}): (1) InternViT-300M Vision Encoder: Retrained on street-level building images to enhance facade feature extraction, leveraging dynamic high-resolution (448px) inputs; (2) MLP Projector: Adjusted to align building-specific visual tokens with textual embeddings in the LLM space; (3) LLM Head: Fine-tuned using the corpus of building characteristic descriptions to generate structured captions. 
Here, we recast façade profiling as a multi-attribute text prediction task: all attributes and captions are encoded in a fixed-template prompt, and the MLLM is fine-tuned to generate that structured output. A single token-level cross-entropy loss implicitly handles both attribute prediction and free-form captioning, without requiring custom loss functions.

\section{Experimental settings}

\subsection{Implementations}
To implement methods mentioned in Section~\ref{method_1},
we manually select cities that have a sufficient number of panoramic images available through the Mapillary online interface, and that also have a considerable amount of objective building attributes openly.
Ultimately, seven cities from three continents are chosen, including Amsterdam, Berlin, Helsinki, San Francisco, Washington D.C., Houston and Manila, balancing the dataset across both selection aspects.
Among them, Helsinki is selected due to its rich availability of building material data from the Buildings in Helsinki data\footnote{\url{https://hri.fi/data/en_GB/dataset/helsingin-rakennukset}}, while Amsterdam provides diverse data on building age, to add sufficient data on according aspects.

The metadata for panoramic SVIs is first downloaded within the defined city boundaries using the Mapillary Python Software Development Kit\footnote{\url{https://github.com/mapillary/mapillary-python-sdk}}, while building data is retrieved using OSMnx~\citep{boeing2017osmnx} and Overture Maps.
Subsequently, the data undergoes the process described in Section~\ref{method_1} to calculate the angle of view, evaluate observation quality, and identify candidate images. These selected images are then utilized for building detection, image reprojection, and filtering, as detailed in Section~\ref{method_2}, resulting in a collection of individual building images for each city.

To ensure the collection of high-quality building images, the filtering criteria we applied are detailed in Table~\ref{tab:image_filtering_criteria}. We acknowledge that stringent filtering may inadvertently exclude images from certain regions, e.g., areas where dense greenery obstructs building facades. To address this and support broader applicability, we have made our code flexible, allowing users to adjust filtering thresholds to accommodate different urban contexts and mitigate potential geographic bias.
Additionally, other image features, such as distance between the building and the vantage point, visibility coverage of the building perimeter, and other indicators derived from semantic segmentation and scene classification, are generated during the processing pipeline. These features can be further integrated into filtering criteria in future implementation. 

\begin{table}[!]
\centering
\caption{Image filtering criteria used for selecting building images in this study.}
\renewcommand{\arraystretch}{1.3}
\resizebox{\textwidth}{!}{
\begin{tabular}{m{5cm}m{8cm}m{7.5cm}}
\toprule[1.25pt]
\textbf{Feature} & \textbf{Description} & \textbf{Selection Rule} \\
\toprule[1.25pt]
\multicolumn{3}{l}{\textit{Mapillary metadata}} \\ \midrule
Image Type & Select only panoramic images & \texttt{is\_pano = True} \\
Location (Duplicate) & Remove images taken at the same location & Filter duplicates based on coordinates \\
Image Quality & Use Mapillary quality score $\in$ [0, 1] to retain images with valid or no score & \texttt{quality\_score $\geq$ 0.2} or \newline \texttt{quality\_score = 0} \\
\midrule
\multicolumn{3}{l}{\textit{Building metadata}} \\ \midrule
Building Area & Retain buildings with sufficient spatial footprint & \texttt{area $>$ 20 m\textsuperscript{2}} \\
Underground & Exclude buildings fully underground & \texttt{building:levels $\geq$ 0} (OSM) or \newline \texttt{is\_underground = False} (Overture) \\
\midrule
\multicolumn{3}{l}{\textit{Isovist analysis}} \\ \midrule
Angle of View (AOV) & Exclude extreme observation angles & \texttt{10° $\leq$ AOV $\leq$ 120°} \\
Observation Coverage & Ensure valid building visibility within search radius & At least one building within 30m \\
Max. Images per Building & Cap the number of images per building & $\leq$ 5 images/building \\
\midrule
\multicolumn{3}{l}{\textit{Image-based filters}} \\ \midrule
File Size & Remove corrupted or very small images & \texttt{size $\geq$ 20 KB} \\
Semantic Segmentation & Check for sufficient visible building surface & \texttt{building\_ratio $\geq$ 0.2}, \texttt{wall\_ratio $\leq$ 0.3}, and \texttt{vegetation\_ratio $\leq$ 0.75} \\
Scene Classification & Exclude indoor or irrelevant content & \texttt{environment\_type = outdoor} \\
Blurriness & Estimate clarity via Laplacian variance & \texttt{blur\_score $\leq$ 30} \\
Brightness & Filter overly dark or overexposed images & \texttt{20 $\leq$ brightness $\leq$ 200} \\
\bottomrule[1.25pt]
\end{tabular}}
\label{tab:image_filtering_criteria}
\end{table}

\subsection{Baselines} \label{sc:baseline}
\paragraph{Models.} To evaluate the effectiveness of large VLM, we compare it against a set of established CV architectures. These models span both classical CNNs and Transformer-based models to serve as baselines across various tasks. All models are implemented using the PyTorch library and initialized with pretrained weights from ImageNet-1K. 

\begin{itemize}
    \item \textbf{VGG}: A deep convolutional network characterized by its straightforward architecture of stacked convolutional layers with small receptive fields~\citep{simonyan2014very}. We employ the VGG16 variant in this study, which has previously been applied to building classification tasks~\citep{kang_building_2018}.

    \item \textbf{DenseNet}: Featuring dense connectivity between layers, DenseNet facilitates feature reuse and improves gradient flow while reducing the number of parameters compared to traditional CNNs~\citep{huang2017densely}. We adopt DenseNet201 as a benchmark model, which has demonstrated effectiveness in building material prediction~\citep{ghione2022building}.

    \item \textbf{ResNet}: We include three variants, ResNet18, ResNet50, and ResNet101, which utilize residual learning through skip connections to enable deeper architectures~\citep{he2016deep}. These models are widely used in building-related recognition tasks~\citep{gouveia2024automated, liang2024evaluating}.

    \item \textbf{ViT}: Vision Transformers (ViT) divide images into fixed-size patches and use Transformer encoders to capture global context through self-attention mechanisms. Here, we evaluate ViT16, ViT32 and Swin Transformer (Swin\_b), which have shown strong performance in previous building classification studies~\citep{raghu_towards_2023, ogawa2023deep}. 
\end{itemize}

Each CV model is fine-tuned separately for each building attribute. A systematic grid search is conducted over a range of learning rates (from 1e-7 to 1e-3), and training is performed for up to 64 epochs with early stopping based on validation performance. The checkpoint achieving the best validation metric (accuracy for classification, R-squared for regression) is selected for final evaluation on the test set and used for comparison with the VLM.

\paragraph{Evaluation Metrics.}
To comprehensively evaluate model performance across different tasks, we adopt a set of widely used metrics, tailored to the respective output types: categorical labels, numerical values, and text.

For tasks such as predicting building type and surface material, we report four standard classification metrics: Accuracy (Acc), macro Precision (Pre), macro Recall (Rec), and macro F1-score (F1). These metrics have been commonly used in prior works on building classification~\citep{he2024ub, sun2022understanding, liang2024evaluating} to evaluate model effectiveness.
For continuous attributes such as the number of floors and building age, we use $R^{2}$, Mean Absolute Error (MAE), Mean Absolute Percentage Error (MAPE), and Root Mean Squared Error (RMSE). These metrics are standard in evaluating building property predictions~\citep{lei_predicting_2024, wang_multi-view_2024}. To enable learning from numerical labels, we convert construction year into building age by subtracting the year from 2025 (i.e., \textit{age} = 2025 – \textit{year}).

Furthermore, the text generated by ChatGPT-4o on the test set serves as a baseline to evaluate performance improvements in image captioning. In this study, we adopt three commonly used metrics in natural language generation: BLEU~\citep{papineni2002bleu}, METEOR~\citep{banerjee2005meteor}, and ROUGE\_L~\citep{lin2004rouge}. BLEU evaluates n-gram precision, while METEOR considers precision, recall, synonym matching, and paraphrase alignment to capture semantic relevance more effectively. ROUGE\_L measures the longest common subsequence between generated and reference texts, highlighting fluency and textual overlap. These metrics are widely adopted in remote sensing vision-language tasks~\citep{li_vision-language_2024}, providing a comprehensive assessment of the accuracy and quality of generated descriptions.

\subsection{Generalizability and robustness} \label{sc_ex_robuts}
To align more closely with real-world building profiling practices, we further conduct comparative assessments involving both VLMs and CV models of their generalizability and robustness. These comparisons target two main objectives: (1) assessing their generalizability to unseen data, and (2) evaluating robustness to heterogeneous noises and degradation.

\paragraph{Generalization to unseen city.}
Generalizing CV models to unseen cities remains a significant challenge due to the diverse and unique architectural features across cities~\citep{sun2022understanding}. 
VLMs, pretrained on extensive, high‐quality image‐text datasets, demonstrate promising potential to overcome these limitations by leveraging their pre‐acquired semantic reasoning and contextual understanding capabilities.
To investigate this potential, we conducted an experiment on building imagery collected from Brussels, comparing the performance of established CNN and ViT architectures against our fine‐tuned VLM. 
For this evaluation, we retrieve a dataset of 3,687 labeled building images by integrating OSM attributes with buildings detected from Mapillary SVI. The dataset is composed of 3,348 images for building type, 186 for surface material, 1,234 for the number of floors, and 106 for building age. 
The supplementary material provides the detailed class distribution of this dataset. Due to the limited number of surface material labels across classes, we focus our analysis on the remaining three attributes.

\begin{figure}[ht]
    \centering
    \makebox[\textwidth]{%
        \includegraphics[width=1\linewidth]{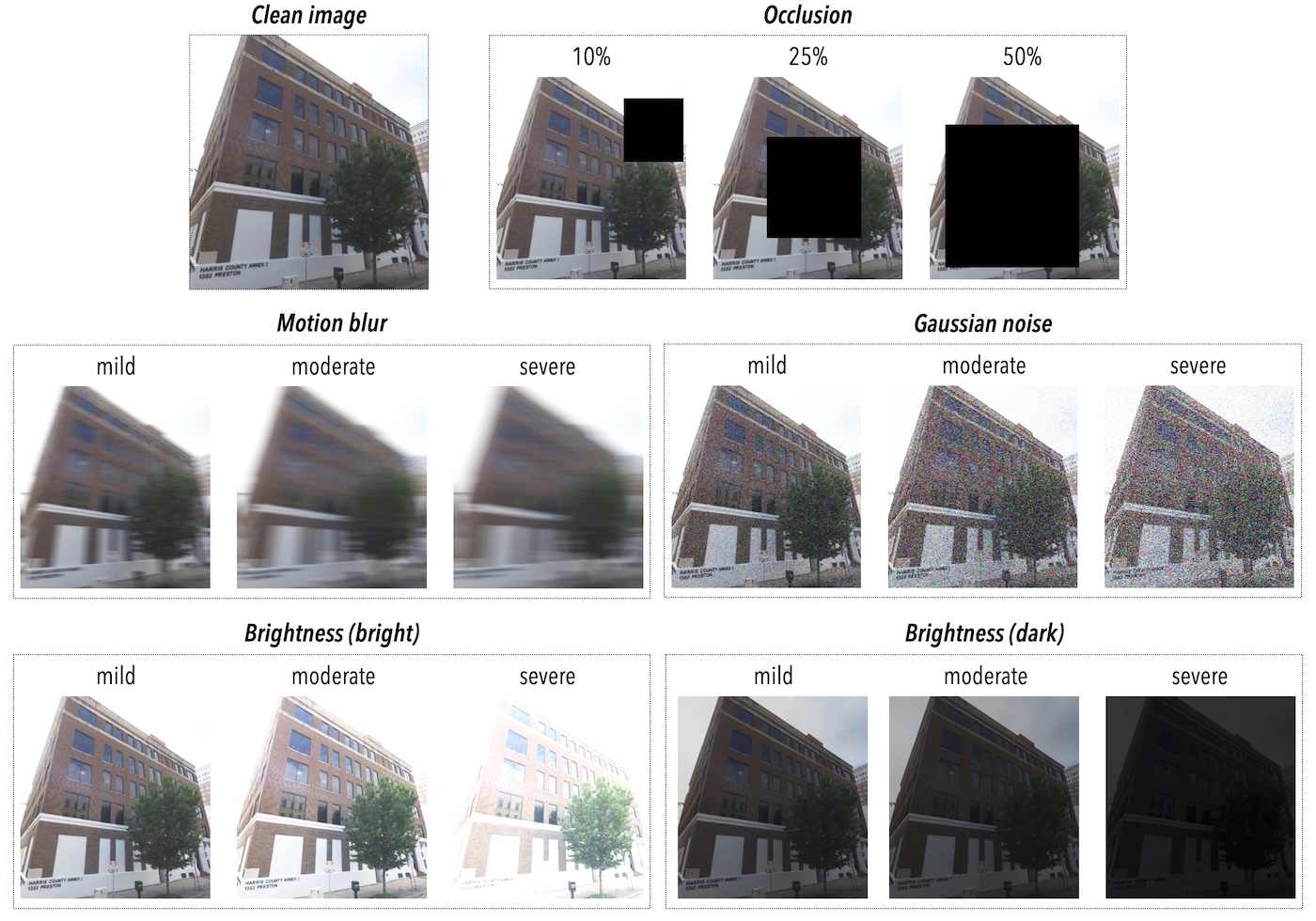}
    }
    \caption{Examples of image corruption and perturbation for robustness experiments, consisting of four categories of algorithmically generated images based on common quality issues in crowdsourced imagery. Each type of corruption has 3 levels of severity (except for brightness which has twice 3 levels of severity), resulting in a total of 15 corruption levels. Data: (c) Mapillary contributors.}
    \label{fig_corruption}
\end{figure}

\paragraph{Robustness to varying image quality.}
Crowdsourced SVI, unlike standardized remote sensing imagery, frequently exhibit diverse quality issues~\citep{hou_comprehensive_2022}. 
To systematically evaluate the robustness and stability of the model under such conditions, we adopt the methodology proposed by~\citet{hendrycks_benchmarking_2019}, which assesses model performance in the presence of common image corruptions and perturbations.
Following the image quality criteria defined by~\citet{hou_comprehensive_2022}, we algorithmically generate four types of corruption to the test set: occlusion, motion blur, Gaussian noise, and brightness alteration, as illustrated in Figure~\ref{fig_corruption}.
We acknowledge that these distortions are simulated rather than naturally occurring, but they provide a controlled and reproducible way to benchmark model behavior under common visual degradation scenarios.

We then evaluate each model’s performance under these degraded conditions using Relative Corruption Errors ($Relative$ CE)~\citep{hendrycks_benchmarking_2019}. 
First, the baseline error rate $E^m_{clean}$ is determined for model $m$ on the uncorrupted data.
Next, we compute the error rate $E^m_{c,s}$ for each corruption type $c$ at severity level $s$  (1 $\leq$ s $\leq$ 3). In classification tasks (building type and surface material), the error rate is defined as $1 - Accuracy$, whereas for numerical values (predicting number of floors and building age), it is defined as $1 - R^2$. 
Finally, to account for the varying difficulties introduced by each corruption, we normalize these error rates by dividing by the ResNet50 baseline error. $Relative$ CE is calculated as:

\begin{equation}
    Relative CE^f_c = (\sum^3_{s=1} E^f_{s,c}-E^f_{clean})/(\sum^3_{s=1} E^{ResNet50}_{s,c}-E^{ResNet50}_{clean})
\end{equation}

\noindent This normalization provides a clearer measure of how much each model’s performance declines under different corruptions. Averaging these $Relative$ CE from four types of corruptions results in the $Relative$ $m$CE, which represents the overall relative performance degradation when the models encountering corruptions.

\subsection{Ablation experiments}
As one of the very first studies to apply fine-tuned open-source VLMs to the task of building profiling, we conduct a series of ablation experiments to explore and identify effective fine-tuning strategies.

\paragraph{Fine-tuning settings.} To determine optimal training configurations, we first compared two fine-tuning strategies: full fine-tuning, where all model parameters are updated, and parameter-efficient fine-tuning using Low-Rank Adaptation (LoRA). LoRA introduces trainable rank-decomposition matrices into the model while freezing the original weights, enabling more efficient training. This comparison allowed us to evaluate the trade-offs between flexibility, performance, and training cost under different optimization schemes.

We then performed a systematic grid search over learning rates and the number of training epochs. The default learning rate of 4e-5 is used as a baseline, and we evaluat additional values (e.g., 8e-6, 4e-6, and 4e-4) within a predefined range to observe their effects on model stability and performance. We also vary the number of training epochs from 1 to 5 while keeping all other hyperparameters fixed. For each setting, the model is evaluated on a held-out validation set.

\paragraph{Model size.} We compare multiple model variants to assess how model capacity influences performance in the context of building profiling. Specifically, we evaluate InternVL3 models with 1B and 2B parameters, as well as InternVL2.5 model which has 4B parameters avaliable. These experiments are conducted to understand the scalability of different VLM sizes and to determine whether increase model complexity leads to substantial gains across all building profiling tasks.

\paragraph{Data size.} To assess the impact of training data volume, we conduct an ablation study by fine-tuning InternVL3-2B on varying proportions of the complete image-text dataset. We evaluate performance on both attribute prediction and image captioning tasks. This analysis quantifies how model accuracy scales with additional data and offered insights into the marginal benefit of increased data volume in multimodal learning.

\section{Results}
\subsection{Street-level building dataset} \label{sc_dataset}

Table~\ref{tb_num_bd} provides a detailed breakdown of the number of buildings and SVI images retrieved, individual buildings detected with associated images, and the ratio of completeness for each city. 
While completeness varies among cities due to differences in the availability and quality of Mapillary images uploaded for specific locations, around 50\% of buildings in city centers can be observed and analyzed.

\begin{table}[h]
\centering
\renewcommand{\arraystretch}{1.5}
\caption{Summary of building footprints, image retrieval, and detection completeness across cities and regions in the building dataset.}
\resizebox{\columnwidth}{!}{%
\begin{tabular}{>{\centering\arraybackslash}m{0.2\textwidth} >{\centering\arraybackslash}m{0.18\textwidth} >{\centering\arraybackslash}m{0.18\textwidth} >{\centering\arraybackslash}m{0.18\textwidth} >
{\centering\arraybackslash}m{0.18\textwidth} >{\centering\arraybackslash}m{0.18\textwidth} >{\centering\arraybackslash}m{0.18\textwidth}}
\toprule[1.5px]
\textbf{City} &
 \textbf{Total building footprints} &
  \textbf{Total images retrieved} &
  \textbf{Total individual building images} &
  \textbf{Buildings with images} &
  \textbf{Percentage detected} &
  \textbf{City center completeness (2.5km$\times$2.5km)} \\ \midrule[1.5px]
\multicolumn{6}{l}{\textit{Europe}}   \\ \hline
Amsterdam     & 195,188 & 203,570 & 330,235 & 120,154 & 61.6\% & 83.6\% \\
Helsinki      & 63,972  & 20,035 & 20,479 & 8,930  & 14.0\% & 42.5\% \\
Berlin        & 497,703 & 408,166 & 287,065 & 137,930 & 27.7\% & 46.7\% \\ \hline
\multicolumn{6}{l}{\textit{North America}}      \\ \hline
San Francisco & 160,659 & 62,521 & 91,874 & 34,510  & 21.5\% & 39.4\% \\
Houston       & 399,883 & 304,030 & 238,934 & 91,774  & 23.0\% & 53.8\% \\
Washington D.C.   & 161,190 & 269,420 & 201,955 & 86,144  & 53.4\% & 57.4\% \\ \hline
\multicolumn{6}{l}{\textit{Asia}} \\ \hline
Manila        & 105,904 & 68,706  & 48,951 & 23,911  & 22.6\% & 22.7\% \\ \midrule[1.5px]
\textbf{Total} &	1,617,019 & 1,414,288 & 1,219,493 &	503,353	 & 31.1\%	 & 49.4\%
\\ \bottomrule[1.5px]
\end{tabular}%
}
\label{tb_num_bd}
\end{table}

\begin{figure}[!]
    \centering
    \includegraphics[width=1\linewidth]{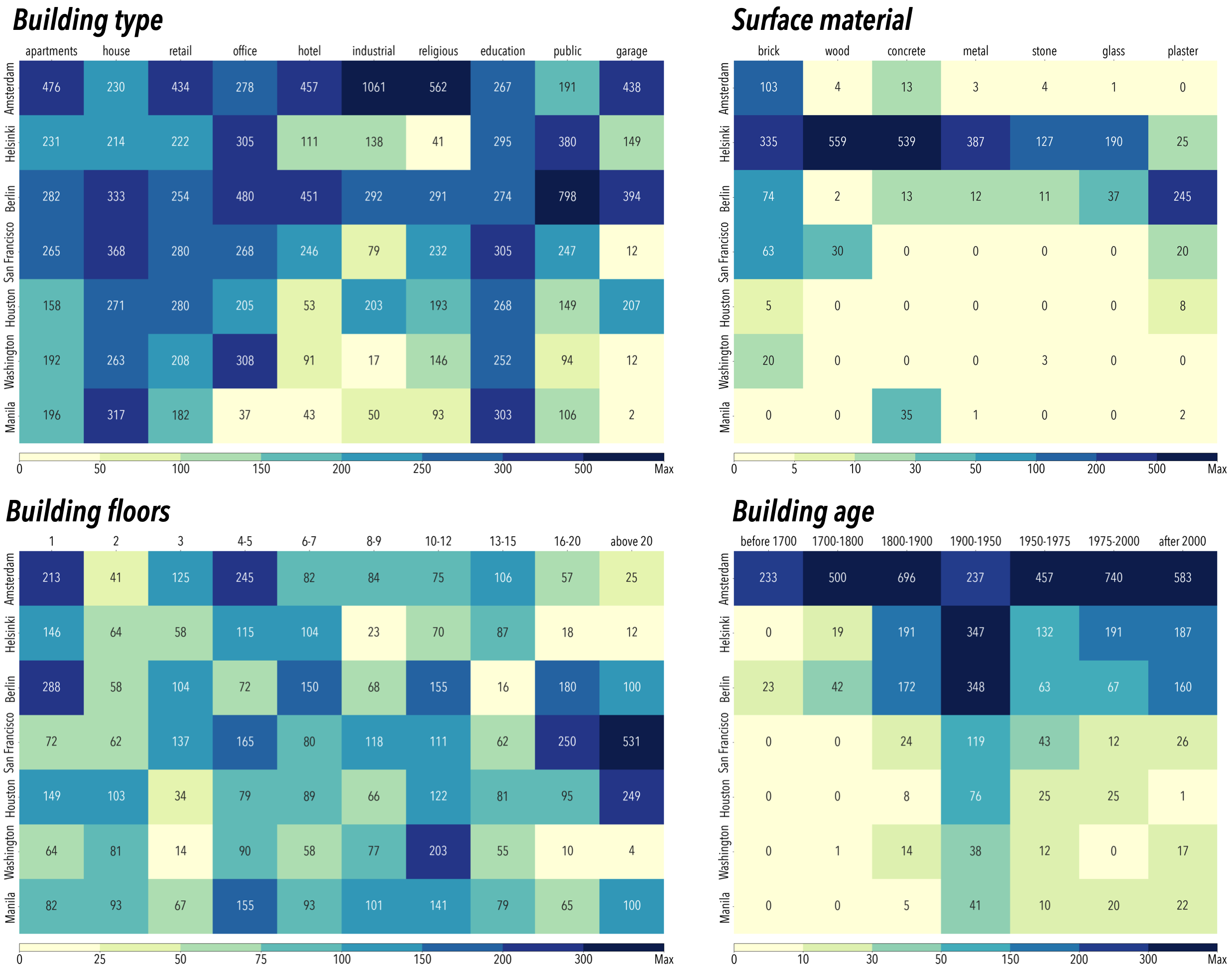}
    \caption{The distribution of the building images categorized by objective building attributes---type, age, floor, and surface material---selected for each city in dataset.}
    \label{fig_cat_num}
\end{figure}

As discussed in Section~\ref{method_3}, building images with available attributes are sampled to construct a class-sufficient dataset for model development, resulting in a total of 31,180 images. Figure~\ref{fig_cat_num} illustrates the distribution of images across categories for each attribute, comprising 17,530 images for building type, 2,871 for surface material, 7,228 for floors, and 5,927 for age.

The dataset is split into training, validation, and test sets in a 6:1:3 ratio. As detailed in Table~\ref{tab:data_summary}, the training set comprises 20,056 OSM‐sourced labeled samples from 19,443 unique images (some buildings carry multiple attribute labels). Those same 19,443 images are used to generate 19,443 multi‐attribute Q\&A pairs and 19,443 descriptive captions via ChatGPT-4o (2024-08-06 release), yielding a total of 58,942 image–text pairs for VLM training. The held-out validation set guides checkpoint selection and hyperparameter tuning for both CV and VLM models based on their performance on multi-attribute prediction tasks. The test set, consisting of 9,016 images with ChatGPT generated Q\&A and captions, is reserved for final model evaluation and comparison.
To better capture ambiguous architectural features, we also include an auxiliary classification setup by introducing prompts such as ``alternate\_building\_type'' and ``alternate\_surface\_material'', which elicit the top-two predictions from the MLLMs.

\begin{table}[!]
\centering
\caption{Summary of training, validation, and test data used in this study.}
\resizebox{0.7\textwidth}{!}{
\begin{tabular}{llccc}
\toprule
\textbf{Data Type} & \textbf{Source} & \textbf{Train} & \textbf{Validation} & \textbf{Test} \\
\toprule
\textit{Image label} & & & & \\ \midrule
Single-attribute label & OSM & 20,056 & 2,737 & 9,190 \\
\midrule
\textit{Image-text pairs} & & & & \\ \midrule
Single-attribute Q\&A & OSM & 20,056 & -- & -- \\
Multi-attribute Q\&A & ChatGPT-4o & 19,443 & -- & 9,016 \\
Captioning & ChatGPT-4o & 19,443 & -- & 9,016 \\
Total pairs & & 58,942 & -- & 18,032 \\
\bottomrule
\end{tabular}}
\label{tab:data_summary}
\end{table}

We acknowledge that the current dataset has limitations, particularly in terms of geographic diversity across continents and the availability of data for certain attributes, such as surface material and building age. 
Nevertheless, to the best of our knowledge, this dataset is both large and comprehensive compared to previous efforts highlighted in Section~\ref{sc_lr}.
Additionally, the reproducible framework established in this study enables future expansion of the dataset as more building images and their associated attributes become available through crowdsourced platforms. This iterative refinement could progressively enhance the dataset's scope and utility for broader applications.

\subsection{Model performance}
\subsubsection{General performance}
After we fine-tuned the baseline models mentioned in Section~\ref{sc:baseline}, we compare their performance with that of the zero-shot ChatGPT and fine‐tuned InternVL3‐2B on four building characteristics, as summarized in Tables~\ref{tab_cnn_class} and \ref{tab_cnn_reg}. 
To facilitate a fair comparison, we integrate GPT-generated data to supplement the missing OSM data in the training set for the CV models.

\begin{table}[!]
    \centering
    \caption{Validation performance comparison among zero-shot ChatGPT, fine-tuned InternVL3-2B and CV models.}

    \begin{subtable}[t]{0.9\textwidth}
        \centering
        \caption{Performance on classification tasks of building type and surface material}
        \resizebox{\textwidth}{!}{
        \renewcommand{\arraystretch}{1}
        \begin{tabular}{>{\centering\arraybackslash}m{1.75cm}%
                        >{\centering\arraybackslash}m{2.8cm}%
                        >{\centering\arraybackslash}m{1.2cm}%
                        >{\centering\arraybackslash}m{1cm}%
                        >{\centering\arraybackslash}m{1.5cm}%
                        >{\centering\arraybackslash}m{1.5cm}%
                        >{\centering\arraybackslash}m{1.5cm}%
                        >{\centering\arraybackslash}m{1.5cm}%
                        >{\centering\arraybackslash}m{2.2cm}}
            \toprule[1.5px]
            \textbf{Attribute} & \textbf{Model} & \textbf{LR} & \textbf{Epoch} & \textbf{Acc (\%)} & \textbf{mPre} & \textbf{mRec} & \textbf{mF1} & \textbf{Acc@2 (\%)} \\
            \midrule[1.5px]
            \multirow{9}{1.5cm}{\centering \textbf{Building type}} 
            & ChatGPT (zero-shot) &	-	& -	& \underline{57.69} & \underline{0.649} & \underline{0.566} & \underline{0.571} & \underline{75.06} \\
            & DenseNet         & 1e-5   & 14      & 53.42 & 0.540 & 0.529 & 0.529 & 69.39 \\
            & VGG              & 5e-4   & 9       & 47.66 & 0.492 & 0.470 & 0.462 & 62.03 \\ 
            & ResNet18         & 1e-4   & 19      & 50.59 & 0.509 & 0.499 & 0.493 & 66.28 \\
            & ResNet50         & 1e-4   & 24      & 52.20 & 0.526 & 0.512 & 0.504 & 65.98 \\
            & ResNet101        & 5e-5   & 13      & 53.58 & 0.543 & 0.527 & 0.524 & 68.52 \\
            & ViT16            & 5e-6   & 7       & 53.60 & 0.549 & 0.532 & 0.530 & 69.35 \\
            & ViT32            & 1e-5   & 5       & 52.24 & 0.519 & 0.515 & 0.507 & 67.93 \\
            & Swin\_b          & 5e-6   & 17      & 56.11 & 0.562 & 0.558 & 0.554 & 73.14 \\
            & InternVL3-2B     & 8e-6   & 3       & \textbf{61.27} & \textbf{0.661} & \textbf{0.602} & \textbf{0.609} & \textbf{77.31} \\ 
            \midrule
            \multirow{9}{1.5cm}{\centering \textbf{Surface material}} 
            & ChatGPT (zero-shot) &	- & -	& 65.41 & 0.581 & 0.614	& 0.553	& 79.07 \\
            & DenseNet         & 1e-4   & 5       & 65.92 & 0.575 & 0.610 & 0.566 & 81.45 \\
            & VGG              & 1e-5   & 12      & 57.02 & 0.504 & 0.529 & 0.494 & 74.06 \\ 
            & ResNet18         & 1e-5   & 23      & 63.78 & 0.560 & 0.587 & 0.554 & 79.57 \\
            & ResNet50         & 5e-5   & 30      & 65.79 & 0.601 & 0.610 & 0.578 & 81.08 \\
            & ResNet101        & 5e-5   & 25      & 66.67 & \underline{0.608} & 0.627 & 0.582 & 79.70 \\
            & ViT16            & 5e-5   & 18      & 64.79 & 0.571 & 0.587 & 0.564 & 81.20 \\
            & ViT32            & 1e-5   & 13      & 60.15 & 0.538 & 0.554 & 0.511 & 77.07 \\
            & Swin\_b          & 5e-6   & 33      & \textbf{69.17} & \textbf{0.623} & \textbf{0.655} & \textbf{0.612} & \textbf{85.84} \\
            & InternVL3-2B     & 8e-6   & 3       & \underline{68.05} & 0.598 & \underline{0.634} & \underline{0.588} & \underline{81.70} \\
            \bottomrule[1.5px]
        \end{tabular}}
        \label{tab_cnn_class}
    \end{subtable}

    \vspace{1em}

    \begin{subtable}[t]{.9\textwidth}
        \centering
        \caption{Performance on prediction tasks of number of floors and building age.}
        \resizebox{\textwidth}{!}{
        \renewcommand{\arraystretch}{1}
        \begin{tabular}{>{\centering\arraybackslash}m{1.75cm}%
                        >{\centering\arraybackslash}m{3cm}%
                        >{\centering\arraybackslash}m{1.2cm}%
                        >{\centering\arraybackslash}m{1cm}%
                        >{\centering\arraybackslash}m{2cm}%
                        >{\centering\arraybackslash}m{2cm}%
                        >{\centering\arraybackslash}m{2cm}%
                        >{\centering\arraybackslash}m{2cm}}
            \toprule[1.5px]
            \textbf{Attribute} & \textbf{Model} & \textbf{LR} & \textbf{Epoch} & \textbf{R2 ($\uparrow$)} & \textbf{MAE ($\downarrow$)} & \textbf{MAPE (\%) ($\downarrow$)} & \textbf{RMSE ($\downarrow$)} \\
            \midrule[1.5px]
            \multirow{9}{1.5cm}{\centering \textbf{Number of floors}} 
            & ChatGPT (zero-shot) &	- &	- &	0.721 &	\underline{2.36} &	\underline{38.66} &	5.01 \\
            & DenseNet         & 5e-5   & 5       & 0.774  & 2.53 & 41.91 & 4.51  \\
            & VGG              & 5e-5   & 12      & 0.689  & 3.10 & 42.43 & 5.28  \\ 
            & ResNet18         & 5e-5   & 16      & 0.741  & 2.82 & 38.81 & 4.82  \\
            & ResNet50         & 5e-5   & 23      & 0.768  & 2.48 & 41.74 & 4.56  \\
            & ResNet101        & 5e-5   & 14      & \underline{0.777}  & 2.48 & 41.97 & \underline{4.48}  \\
            & ViT16            & 5e-6   & 6       & 0.765  & 2.63 & 46.47 & 4.59  \\
            & ViT32            & 1e-6   & 22      & 0.727  & 2.83 & 45.08 & 4.95  \\
            & Swin\_b          & 1e-5   & 10      & 0.773  & 2.41 & 41.48 & 4.52  \\
            & InternVL3-2B     & 8e-6   & 3       & \textbf{0.789}  & \textbf{2.13} & \textbf{36.93} & \textbf{4.35}  \\
            \midrule
            \multirow{9}{1.5cm}{\centering \textbf{Building age}} 
            & ChatGPT (zero-shot) &	- &	- &	0.645 &	31.63 &	\underline{58.79} &	57.07 \\
            & DenseNet         & 1e-5   & 21      & 0.720 & 31.31 & 78.36 & 50.68 \\
            & VGG              & 1e-4   & 7       & 0.559 & 42.55 & 121.31 & 63.59 \\ 
            & ResNet18         & 5e-5   & 23      & 0.714 & 32.63 & 74.44 & 51.23 \\
            & ResNet50         & 1e-4   & 18      & 0.721 & 32.26 & 85.92 & 50.55 \\
            & ResNet101        & 5e-5   & 18      & \textbf{0.738} & \underline{30.61} & 78.77 & \textbf{49.01} \\
            & ViT16            & 5e-6   & 29      & 0.719 & 30.89 & 80.02 & 50.77 \\
            & ViT32            & 1e-5   & 23      & 0.675 & 33.82 & 71.89 & 54.61 \\
            & Swin\_b          & 5e-6   & 27      & \underline{0.723} & 32.72 & 93.77 & \underline{50.40} \\
            & InternVL3-2B     & 8e-6   & 3       & 0.710 & \textbf{28.06} & \textbf{58.27} & 51.50 \\
            \bottomrule[1.5px]
        \end{tabular}}
        \label{tab_cnn_reg}
    \end{subtable}

    \label{tab:cnn_combined}
\end{table}

The fine-tuned InternVL3-2B achieves the highest overall performance, particularly in predicting building type and number of floors. Although CV models slightly outperform the VLM in surface material classification and building age estimation, they require attribute-specific architectures and domain-specific tuning. In contrast, VLMs provide a unified and adaptable framework, delivering comparable or superior results across multiple prediction tasks.
Notably, zero-shot ChatGPT also demonstrates strong capability across all tasks, achieving performance close to that of some fine-tuned CV models. This suggests its potential as a practical tool for supplementing building data when labeled samples are limited.
This advantage of VLM can be attributed primarily to the semantic reasoning and contextual understanding inherent in pretrained VLMs, which significantly enhance their generalization across diverse tasks. Moreover, VLMs can simultaneously infer multiple target variables and implicitly model correlations among tokens associated with different tasks, whereas the CV baselines considered here are independently fine-tuned for each attribute.
The combination of robust predictive performance and enhanced functionality makes VLMs a compelling alternative to traditional approaches in future applications.

\subsubsection{Performance by cities}

\begin{figure}[!]
    \centering
    \makebox[\textwidth]{%
        \includegraphics[width=1.25\linewidth]{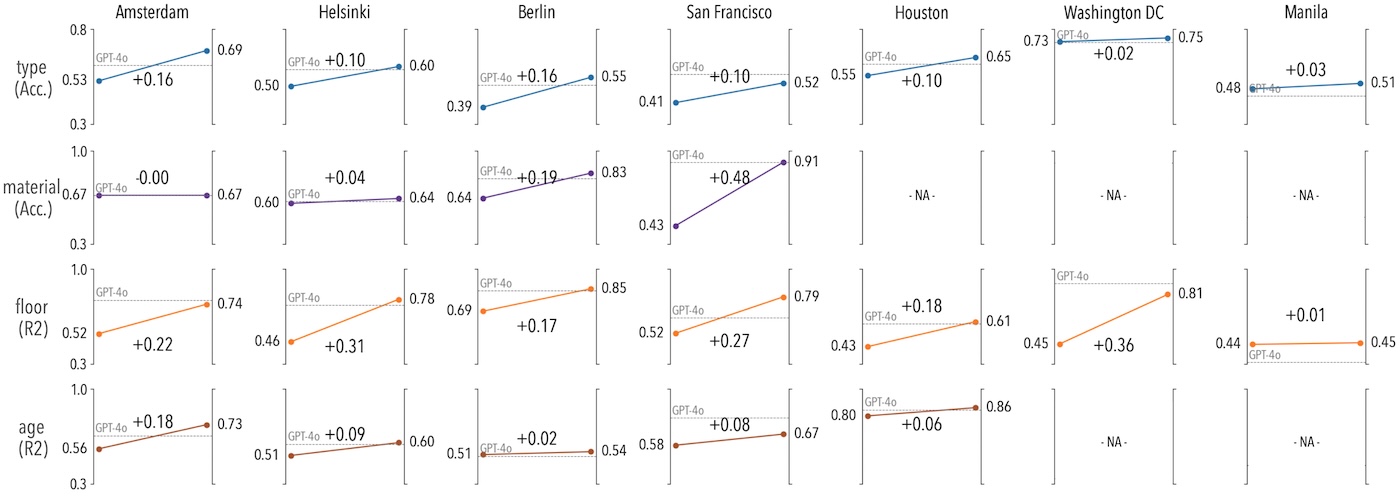}
    }
    \caption{Model performance on building attributes across different cities before and after fine-tuning the InternVL3-2B model, compared to the baseline performance of ChatGPT-4o. Building type and surface material are evaluated using classification accuracy (Acc.), while number of floors and building age are assessed using R-squared (R2). ``NA'' indicates cities with insufficient data for model evaluation (ground‐truth instances fewer than 20 in test set).}
    \label{fig_per_city}
\end{figure}

Figure~\ref{fig_per_city} presents the performance of three model variants: (1) the InternVL3‐2B model before fine‐tuning and (2) after fine‐tuning, as well as (3) a ChatGPT‐4o reference baseline, on seven cities and four building attributes.
In general, the fine‐tuned InternVL3‐2B outperforms its non‐fine‐tuned counterpart, showing consistent gains in predicting all four attributes. 
These improvements are particularly notable in Berlin and San Francisco, where building type, material, and floor performance improve substantially. Amsterdam and Helsinki also exhibit moderate but still positive gains for different tasks.

Despite the overall upward trend, improvement magnitude varies across cities and attributes, which may due to several reasons. 
First, the availability of diverse and distinct samples plays a crucial role: cities with a richer variety of building facades (e.g., Amsterdam, Berlin) yield more pronounced performance boosts. 
Conversely, location with more homogeneous or ambiguous building styles (e.g., Manila) shows relatively smaller gains. 
Second, crowdsourced labels in certain cities may be incorrect or insufficient, which can adversely affect the model’s ability to learn reliable city‐specific patterns, restraining potential performance gains.

Nevertheless, when benchmarked against the zero-shot ChatGPT‐4o baseline, the fine‐tuned InternVL3‐2B model demonstrates generally competitive or superior performance. 
These results confirm that open‐access VLMs can achieve near‐state‐of‐the‐art performance at no additional licensing cost once adequately fine‐tuned on relevant datasets. 
This highlights the effectiveness of VLMs in predicting multiple building attributes across global cities, providing a cost‐effective solution for a wide range of urban remote sensing applications.

\subsubsection{Performance by categories}
\begin{figure}[h!]
    \centering
    \includegraphics[width=1\linewidth]{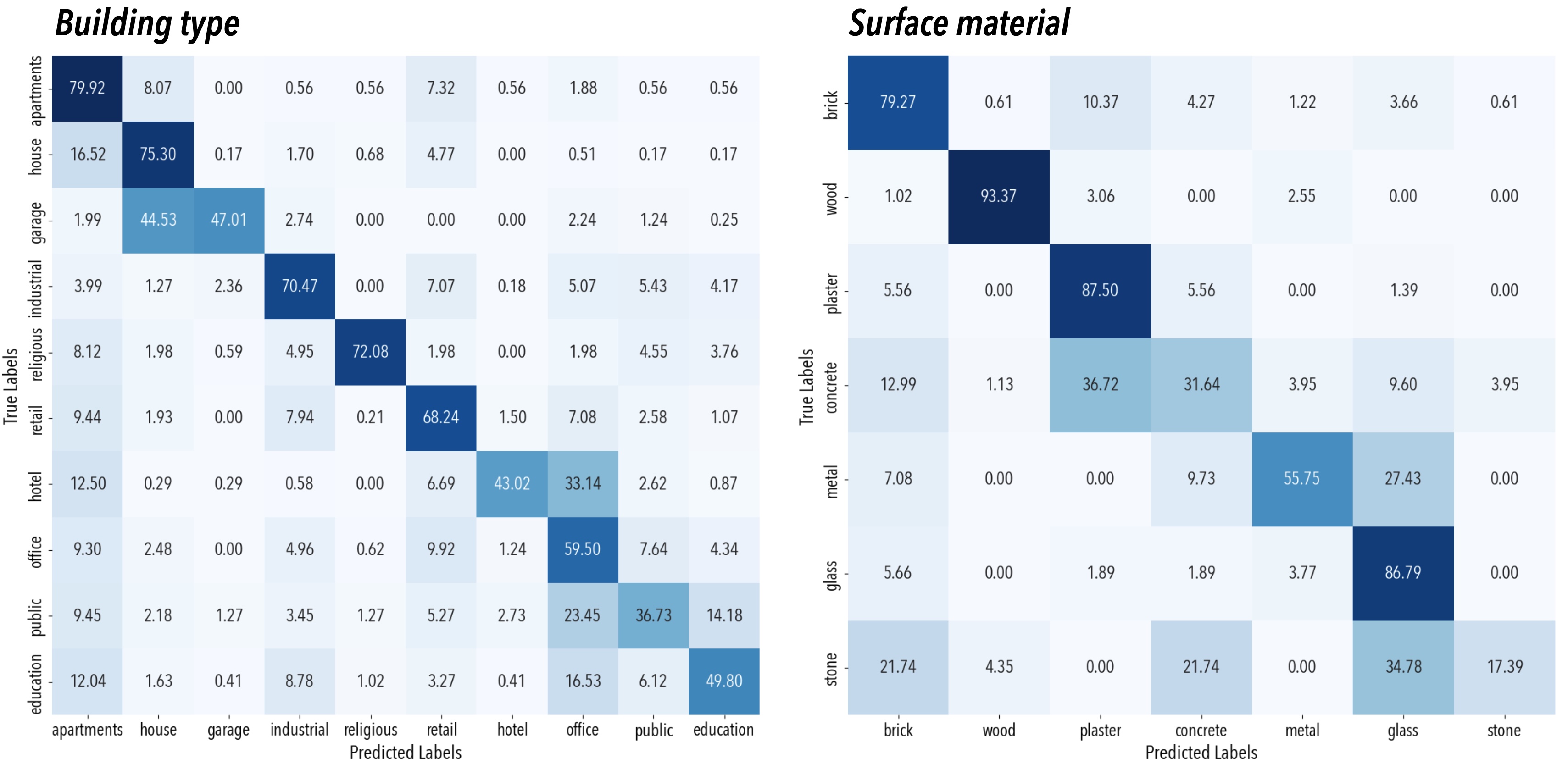}
    \caption{Confusion matrices illustrating the performance of the InternVL3-2B model on classifying different categories of building type and surface material. Darker cells indicate higher prediction accuracy (\%).}
    \label{fig_confuse_class}
\end{figure}

Figure~\ref{fig_confuse_class} presents the confusion matrices illustrating the performance of our VLM on building type and surface material on different categorical labels. 
Overall, the model demonstrates robust performance for most classes.
In terms of well‐predicted labels, visually distinctive building types such as apartments and houses show consistently high accuracies. These categories often have defining features (e.g., apartment blocks characterized by uniform facades and repetitive windows) that the model effectively captures. Similarly, for surface material, high‐frequency and visually salient classes like brick, wood, and glass yield strong performances.
Conversely, certain labels are harder to classify, yielding relatively lower accuracies. 
For building type, hotel or public categories are frequently misidentified as office, suggesting significant overlap in their institutional architectural appearance (e.g., multi-stories, ordered elements). Likewise, plaster and concrete exhibit misclassifications due to shared grayscale tones and blank textures.

\begin{figure}[!]
    \centering
    \includegraphics[width=1\linewidth]{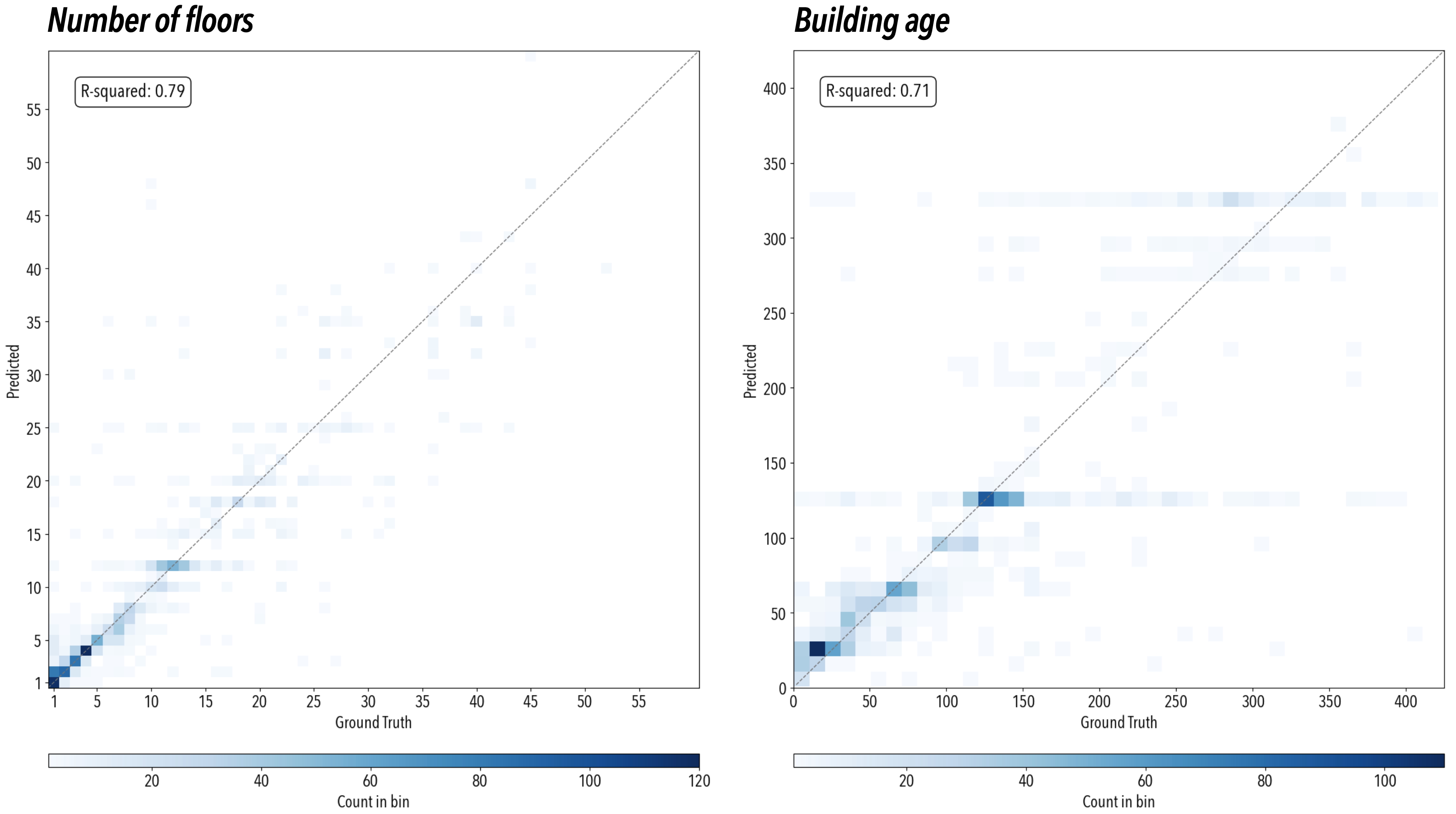}
    \caption{Heatmaps illustrating the performance of the InternVL3-2B model in predicting the number of floors (left) and building age (right). Darker colors indicate a higher frequency of predictions within each ground-truth bin.}
    \label{fig_confuse_num}
\end{figure}

Figure~\ref{fig_confuse_num} illustrates the model's ability to predict the number of floors and building age. The heatmaps provide a detailed visualization of prediction accuracy relative to ground-truth values, with darker colors indicating higher frequencies of accurate predictions within each bin.
For the number of floors (left), the heatmap demonstrates strong prediction performance for lower-rise buildings. As the number of floors increases beyond five, prediction performance start to decline. High-rise structures (stories beyond 15) exhibit more frequent errors, highlighting the challenge in distinguishing floor count for taller buildings from single viewpoint images. 

For building age (right), the heatmap reveals stronger predictive performance for more recent buildings (post-1900), reflecting their typically more distinctive and recognizable architectural characteristics. Conversely, older buildings (pre-1900) exhibit larger errors, with significant overlaps between adjacent historical periods. The subtle external distinctions, combined with modifications such as renovations and retrofits, contribute to the difficulty in accurately classifying older structures based solely on visual appearance~\citep{sun2022understanding}. Additionally, the model's performance is likely influenced by the limited representation of older building periods within the original training dataset. Nevertheless, the overall R-squared above 0.7 confirms the model's capability in capturing broad temporal patterns of floor count and building age from images.

Taken together, these results suggest that the model is capable of inferring attributes across various ranges. However, inherent visual ambiguities, particularly among structurally or stylistically similar categories, contribute to overlaps in predictions. 
Enhancing the quality, diversity, and coverage of crowdsourced data would be a valuable step toward improving the dataset and the model's performance.

\subsection{Generalizability and robustness}
\paragraph{Generalization to unseen city.}
CV models with good performance are selected for experiments in this section.
Tables~\ref{tab_gen_class} and~\ref{tab_gen_reg} show that the VLM model exhibits superior generalizability compared to commonly used computer vision models, particularly in the tasks of building type and age prediction.
This improved performance can be attributed to the VLM’s ability to leverage semantic reasoning and contextual understanding from large-scale image-text pretraining, allowing it to adapt more effectively to diverse architectural features.
While the performance on number of floors prediction is comparable across models, the lower accuracy of Swin\_b on other attributes suggests a reliance on localized visual features, which limits its generalization capability.
Despite these strengths, the VLM's low macro-F1 score for building-type classification highlights persistent challenges, notably the underrepresentation of minority classes and ambiguity in building features. To address these issues, assembling a more balanced and diverse building imagery dataset and incorporating domain-specific fine-tuning are recommended. 
Collectively, these findings underscore the potential of VLMs to enhance robustness and generalizability across varied urban contexts.

\begin{table}[h!]
  \centering
  \caption{Validation performance comparison of different models on building images in Brussels.}

  \begin{subtable}[t]{0.75\textwidth}
    \centering
    \caption{Performance on classification task of building type}
    \resizebox{\textwidth}{!}{
      \renewcommand{\arraystretch}{1.1}
      \begin{tabular}{>{\centering\arraybackslash}m{2.5cm}%
                      >{\centering\arraybackslash}m{2.5cm}%
                      >{\centering\arraybackslash}m{1.5cm}%
                      >{\centering\arraybackslash}m{1.5cm}%
                      >{\centering\arraybackslash}m{1.5cm}%
                      >{\centering\arraybackslash}m{1.5cm}}
        \toprule[1.5px]
        \textbf{Attribute} & \textbf{Model} & \textbf{Acc (\%)} & \textbf{mPre} & \textbf{mRec} & \textbf{mF1} \\
        \midrule[1.5px]
        \multirow{5}{*}{\textbf{Building type}} 
        & DenseNet      & 31.12 & 0.25 & 0.33 & 0.21 \\
        & ResNet101     & 31.20 & 0.24 & 0.34 & 0.21 \\
        & ViT16         & 32.54 & 0.26 & 0.32 & 0.21 \\
        & Swin\_b        & 24.58 & 0.26 & 0.33 & 0.20 \\
        & InternVL3-2B  & \textbf{63.80} & \textbf{0.44} & \textbf{0.50} & \textbf{0.42} \\
        \bottomrule[1.5px]
      \end{tabular}
    }
    \label{tab_gen_class}
  \end{subtable}

  \vspace{1em}

  \begin{subtable}[t]{0.85\textwidth}
    \centering
    \caption{Performance on prediction tasks of number of floors and building age}
    \resizebox{\textwidth}{!}{
      \renewcommand{\arraystretch}{1.1}
      \begin{tabular}{>{\centering\arraybackslash}m{2.5cm}%
                      >{\centering\arraybackslash}m{2.5cm}%
                      >{\centering\arraybackslash}m{2cm}%
                      >{\centering\arraybackslash}m{2cm}%
                      >{\centering\arraybackslash}m{2cm}%
                      >{\centering\arraybackslash}m{2cm}}
        \toprule[1.5px]
        \textbf{Attribute} & \textbf{Model} & \textbf{RMSE ($\downarrow$)} & \textbf{MAE ($\downarrow$)} & \textbf{MAPE (\%) ($\downarrow$)} & \textbf{R$^2$ ($\uparrow$)} \\
        \midrule[1.5px]
        \multirow{5}{*}{\textbf{Number of floors}} 
        & DenseNet      & 1.82 & 1.13 & 30.20 & 0.610 \\
        & ResNet101     & 1.90 & 1.15 & 33.46 & 0.576 \\
        & ViT16         & 1.79 & 1.15 & 33.10 & 0.624 \\
        & Swin\_b        & 1.74 & 1.08 & 29.39 & \textbf{0.645} \\
        & InternVL3-2B  & \textbf{1.73} & \textbf{0.90} & \textbf{26.20} & \textbf{0.645} \\
        \midrule
        \multirow{5}{*}{\textbf{Building age}} 
        & DenseNet      & 55.50 & 43.22 & 74.89 & 0.200 \\
        & ResNet101     & 51.55 & 39.83 & 172.65 & 0.166 \\
        & ViT16         & 51.44 & 38.85 & 168.38 & 0.157 \\
        & Swin\_b        & 54.92 & 42.27 & 195.16 & 0.170 \\
        & InternVL3-2B  & \textbf{44.67} & \textbf{30.76} & \textbf{66.66} & \textbf{0.600} \\
        \bottomrule[1.5px]
      \end{tabular}
    }
    \label{tab_gen_reg}
  \end{subtable}

  \label{tab:combined_gen}
\end{table}

\begin{figure}[!]
    \centering
    \makebox[\textwidth]{%
        \includegraphics[width=1.2\linewidth]{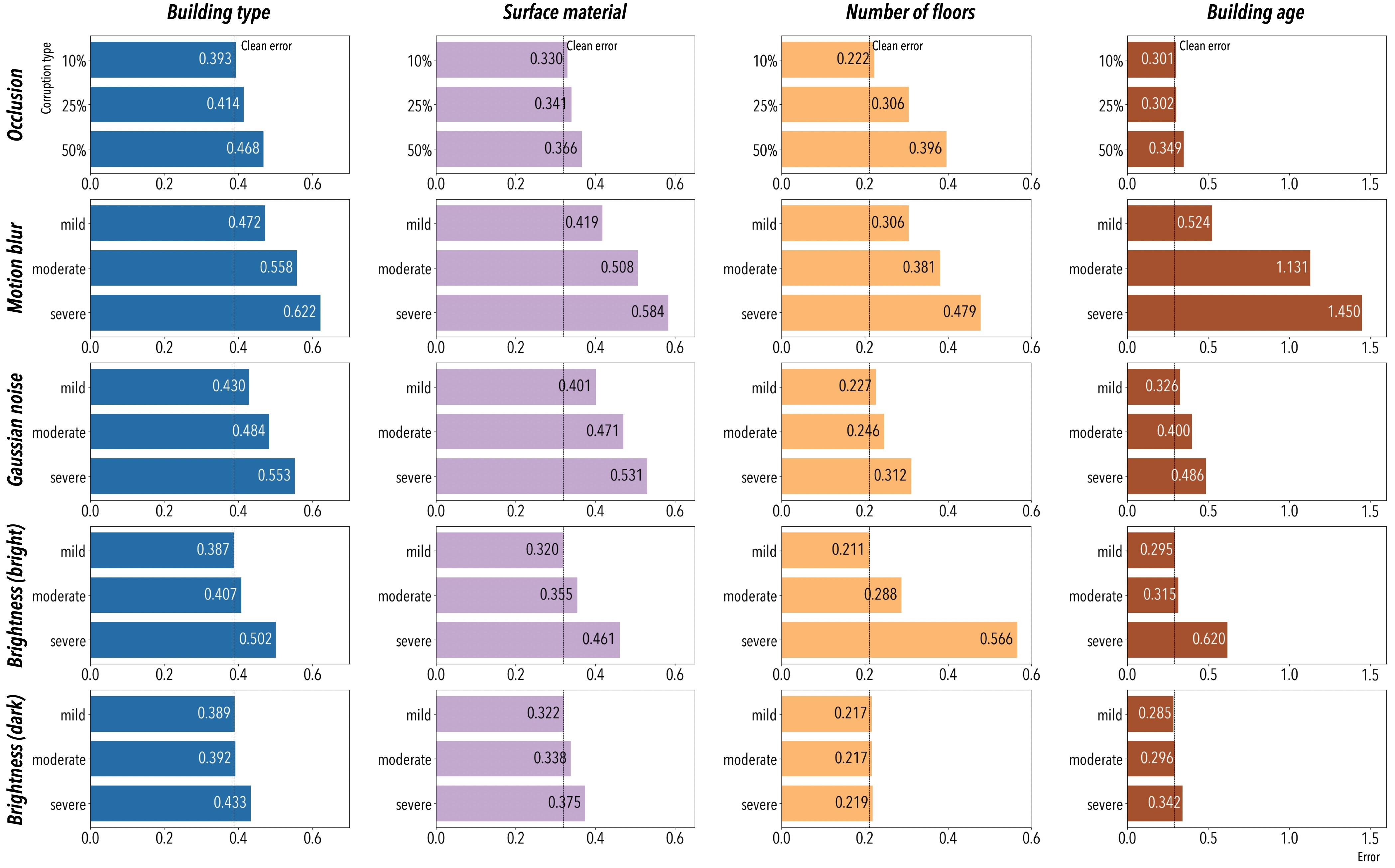}
    }
    \caption{Error rates of the VLM under different severity levels of image corruption. The dotted line represents the clean error obtained from the original test set, serving as a baseline for comparison.}
    \label{fig_corrup_ce}
\end{figure}

\paragraph{Robustness to varying image quality.}
As described in Section~\ref{sc_ex_robuts}, we evaluate the robustness of the VLM against image corruptions by testing it on a perturbed dataset derived from the test set. 
Figure~\ref{fig_corrup_ce} illustrates the model's performance under varying severity levels of occlusion, motion blur, Gaussian noise, and brightness distortions. In general, the model demonstrates resilience, with performance dropping by less than 10\% under most mild and moderate image corruptions. 
In particular, the model remains significantly stable in handling lighting variations and occlusion, both of which are common challenges in crowdsourced image datasets.
However, the model experiences a significant performance drop when confronted with moderate to severe noise and blurriness. 
In particular, the model is most affected by motion blur, where the error rate for the number of floors prediction increases from 0.21 (clean error) to 0.48, and the error rate for building age prediction rises from 0.29 (clean error) to 1.45. 
These findings emphasize the necessity of image preprocessing techniques to filter out degraded images during the image selection stage.

Furthermore, the \emph{Relative} $m$CE is computed for different models using ResNet50 as the baseline. Table~\ref{tab_relative_ce} presents the relative error rates across different building attribute prediction tasks, indicating each model’s stability compared to the baseline.
In general, different models demonstrate various capability in handing corruption data. 
Multi-attribute prediction VLM (InternVL3-2B) demonstrates superior stability compared to single-attribute CV models in most cases, having lowest mean \emph{Relative} $m$CE among all attributes.
Especially, when distinguishing building type and building age when encountering occlusion, motion and brightness variations, VLM demostrate well stability.
CNN models' stability performs comparably to more advanced models in the tasks of number of floors prediction, while ViT model performs superior in handling data with Gaussian noise. 
This outcome implies that additional domain-specific constraints or specialized training strategies might be required to enhance performance on crowdsourced image data.

\begin{table}[!]
    \centering
    \caption{Relative prediction error compared to ResNet50 under different corruptions and perturbations for different objective attributes.}
    \resizebox{.9\textwidth}{!}{%
    \renewcommand{\arraystretch}{1}
    \begin{tabular}{>{\centering\arraybackslash}m{2cm}
                    >{\centering\arraybackslash}m{3cm}
                    >{\centering\arraybackslash}m{1.8cm}
                    >{\centering\arraybackslash}m{1.8cm}
                    >{\centering\arraybackslash}m{1.8cm}
                    >{\centering\arraybackslash}m{1.8cm}
                    >{\centering\arraybackslash}m{2.8cm}}
        \toprule[1.5px]
        \textbf{Attribute} & \textbf{Model} & \textbf{Occlusion} & \textbf{Motion} & \textbf{Noise} & \textbf{Brightness} & \textbf{Relative mCE} \\
        \midrule[1.5px]
        \multirow{6}{2cm}{\centering\textbf{Building type}} 
            & ResNet50                     & 1.00 & 1.00 & 1.00 & 1.00 & 1.00 \\
            & DenseNet       & 2.10 & 0.98 & 0.86 & 0.78 & 1.18 \\
            & ResNet101    & 0.89 & 0.95 & 0.61 & 0.54 & 0.75 \\
            & ViT16         & 0.76 & 0.83 & \textbf{0.29} & 1.22 & 0.78 \\
            & Swin\_b & 0.88 & 1.12 & 0.30 & 0.47 & 0.69 \\
            & InternVL3-2B                 & \textbf{0.51} & \textbf{0.42} & 0.39 & \textbf{0.46} & \textbf{0.44} \\
        \midrule
        \multirow{6}{2cm}{\centering\textbf{Surface material}} 
            & ResNet50                     & 1.00 & 1.00 & 1.00 & 1.00 & 1.00 \\
            & DenseNet       & 1.26 & 1.09 & 1.20 & 1.16 & 1.18 \\
            & ResNet101     & 1.49 & 0.98 & 1.25 & 1.02 & 1.19 \\
            & ViT16        & \textbf{0.48} & 0.72 & \textbf{0.50} & 1.48 & 0.80 \\
            & Swin\_b & 0.99 & 1.11 & 1.03 & 0.70 & 0.96 \\
            & InternVL3-2B                 & 0.59 & \textbf{0.66} & 0.64 & \textbf{0.47} & \textbf{0.59} \\
        \midrule
        \multirow{6}{2cm}{\centering\textbf{Number of floors}} 
            & ResNet50                     & 1.00 & 1.00 & 1.00 & 1.00 & 1.00 \\
            & DenseNet        & 2.01 & 1.62 & 0.86 & 1.24 & 1.43 \\
            & ResNet101        & 0.80 & 1.19 & \textbf{0.31} & 0.69 & 0.75 \\
            & ViT16         & 1.07 & 1.41 & 0.46 & 1.61 & 1.14 \\
            & Swin\_b    & 0.66 & 1.87 & 0.49 & 0.84 & 0.97 \\
            & InternVL3-2B                 & \textbf{0.63} & \textbf{0.78} & 0.42 & \textbf{0.54} & \textbf{0.59} \\
        \midrule
        \multirow{6}{2cm}{\centering\textbf{Building age}} 
            & ResNet50                     & 1.00 & 1.00 & 1.00 & 1.00 & 1.00 \\
            & DenseNet          & 2.65 & 0.94 & 0.99 & 1.43 & 1.50 \\
            & ResNet101         & 1.04 & 0.78 & 0.48 & 1.10 & 0.85 \\
            & ViT16             & 0.50 & 0.78 & \textbf{0.27} & 1.80 & 0.84 \\
            & Swin\_b   & 0.67 & 1.11 & 2.90 & 1.43 & 1.53 \\
            & InternVL3-2B                 & \textbf{0.30} & \textbf{0.69} & 0.48 & \textbf{0.54} & \textbf{0.50} \\
        \bottomrule[1.5px]
    \end{tabular}}
    \label{tab_relative_ce}
\end{table}

In conclusion, building on insights from CV baselines, VLMs not only demonstrate robust and generalizable features for tackling diverse tasks based on crowdsourced data, but they also represent a promising framework for large-scale or cross-regional implementations that demand multi-feature prediction and flexible adaptation.

\subsection{Ablation experiments}
\paragraph{Fine-tuning settings.} As shown in Table~\ref{tb:ablation_lr}, we compare model performance across different settings, including LoRA, full fine-tuning, number of epochs, and learning rate. We summarize overall performance using two aggregate metrics: Overall Accuracy (OA), computed as the average of accuracies, and Overall R$^2$ (OR), computed as the average R$^2$. In general, full fine-tuning consistently outperforms LoRA across most tasks. Although LoRA offers parameter efficiency and retains the original checkpoint’s general capabilities, it typically yields lower performance. Therefore, we adopt full fine-tuning for further investigation.

We further investigated the effect of training epochs under the default learning rate of 4e-5. Performance generally improves from epoch 1 to 3, with diminishing or unstable gains beyond that point. For example, at epoch 3, full fine-tuning achieves strong and balanced results, including a peak OA of 64.02\% and competitive OR. Although epoch 4 yields a slightly higher OR (0.725), the improvements are marginal and come at the cost of additional computation. Thus, epoch 3 is selected as the most efficient convergence point.

We also experimented with several learning rates, including the default 4e-5 and smaller values such as 4e-6. A learning rate of 8e-6 offers improved stability and slightly better OR. In addition, our observations suggest that smaller learning rates are especially beneficial for fine-tuning on relatively small datasets, as they help retain general-purpose capabilities such as open-ended reasoning and semantic alignment.
Based on these comparisons, we adopt full fine-tuning with 3 training epochs and a learning rate of 8e-6 as our final configuration. This setting provides the best trade-off between predictive performance, training efficiency, and the preservation of the pretrained strengths of large VLMs.

\begin{table}[!]
\centering
\caption{Validation performance across epochs and learning rates for LoRA and full‐tuning schemes.}
\resizebox{\textwidth}{!}{
\begin{tabular}{cc
                cc
                cc
                c
                cc
                cc
                c}
\toprule[1.25pt]
\multirow{2}{*}{\textbf{Epoch}} 
  & \multirow{2}{*}{\textbf{LR}} 
    & \multicolumn{2}{c}{\textbf{Acc (\%)}} 
    & \multicolumn{2}{c}{\textbf{mF1}} 
    & \multirow{2}{*}{\textbf{OA}} 
    & \multicolumn{2}{c}{\textbf{R$^2$}} 
    & \multicolumn{2}{c}{\textbf{MAE}} 
    & \multirow{2}{*}{\textbf{OR}} \\
\cmidrule(lr){3-4}\cmidrule(lr){5-6}\cmidrule(lr){8-9}\cmidrule(lr){10-11}
  & 
    & Type & Material 
    & Type & Material 
    & 
    & Floor & Age 
    & Floor & Age 
    &  \\ 
\midrule[1.25pt]
\multicolumn{12}{l}{\textit{LoRA}} \\ \midrule
1 & 4e-5 & 60.29 & 60.29 & 0.583 & 0.513 & 60.29 & 0.739 & 0.524 & 2.64 & 40.44 & 0.632 \\
2 & 4e-5 & 60.66 & 62.50 & 0.593 & 0.532 & 61.58 & 0.729 & 0.560 & 2.60 & 38.01 & 0.645 \\
3 & 4e-5 & 60.52 & 63.24 & 0.592 & 0.537 & 61.88 & 0.745 & 0.595 & 2.52 & 35.79 & 0.670 \\
4 & 4e-5 & 60.11 & 64.34 & 0.583 & 0.548 & 62.22 & 0.741 & 0.593 & 2.52 & 35.73 & 0.667 \\
5 & 4e-5 & 60.34 & 64.86 & 0.586 & 0.553 & 62.60 & 0.757 & 0.571 & 2.51 & 35.97 & 0.664 \\
\midrule
\multicolumn{12}{l}{\textit{Full‐tuning}} \\ \midrule
1 & 4e-5 & 62.64 & 63.97 & 0.608 & 0.538 & 63.31 & 0.719 & 0.637 & 2.50 & 33.25 & 0.678 \\
2 & 4e-5 & 62.02 & 64.71 & 0.599 & 0.542 & 63.37 & 0.735 & 0.672 & 2.40 & 30.95 & 0.704 \\
3 & 4e-5 & 63.32 & 64.71 & 0.618 & 0.542 & \textbf{64.02} & 0.734 & 0.680 & 2.38 & 30.44 & 0.707 \\
4 & 4e-5 & 62.09 & 65.07 & 0.605 & 0.545 & \underline{63.58} & 0.748 & 0.701 & 2.31 & 29.03 & \textbf{0.725} \\
5 & 4e-5 & 62.41 & 64.34 & 0.608 & 0.541 & 63.38 & 0.741 & 0.696 & 2.39 & 29.14 & \underline{0.719} \\ 
\midrule
3 & 4e-4 & 10.31 & 16.54 & 0.030 & 0.085 & 13.43 & -0.291 & -0.060 & 7.88 & 79.28 & -0.176 \\
3 & 8e-5 & 62.36 & 63.24 & 0.609 & 0.546 & 62.80 & 0.736 & 0.646 & 2.43 & 31.73 & 0.691 \\
3 & 4e-5 & 63.32 & 64.71 & 0.618 & 0.542 & \textbf{64.02} & 0.734 & 0.680 & 2.38 & 30.44 & \underline{0.707} \\
3 & 8e-6 & 63.43 & 63.97 & 0.619 & 0.540 & \underline{63.70} & 0.762 & 0.665 & 2.19 & 31.63 & \textbf{0.714} \\
3 & 4e-6 & 63.39 & 62.50 & 0.620 & 0.532 & 62.94 & 0.748 & 0.653 & 2.33 & 31.91 & 0.701 \\
\bottomrule[1.25pt]
\end{tabular}}
\label{tb:ablation_lr}
\end{table}

\paragraph{Model size.}
We further compare the performance of VLM variants with different model sizes under both zero-shot and fine-tuned settings, as shown in Tables~\ref{tab_model_class} and~\ref{tab_model_reg}. In the zero-shot setting, larger model sizes or more recent pretrained weights generally yield better predictions. 

The results confirm that fine-tuning is essential for domain adaptation, consistently improving performance across all building profiling tasks. For example, classification accuracy improves by 7–15 percentage points over zero-shot baselines, while R$^2$ values for floor and age prediction also increase significantly.
After fine-tuning, model performance becomes more comparable across sizes, though task-specific variations remain. InternVL2.5–4B achieves the highest accuracy in classifying building type and surface material, whereas InternVL3 variants perform better on floor count and age prediction. These findings suggest that models in the 1–2B parameter range are generally sufficient for building profiling, particularly when training on relatively small datasets.

\begin{table}[!]
    \centering
    \caption{Validation performance of VLM variants with different model sizes under both zero-shot and fine-tuning conditions.}
    
    \begin{subtable}[t]{0.8\textwidth}
        \centering
        \caption{Building type and surface material classification in zero-shot and fine-tuned settings.}
        \resizebox{\textwidth}{!}{
        \renewcommand{\arraystretch}{1.1}
        \begin{tabular}{>{\centering\arraybackslash}m{2cm}
                >{\centering\arraybackslash}m{2.5cm}
                >{\centering\arraybackslash}m{1.5cm}
                >{\centering\arraybackslash}m{1.5cm}
                >{\centering\arraybackslash}m{1.5cm}
                >{\centering\arraybackslash}m{1.5cm}
                >{\centering\arraybackslash}m{1.5cm}
                >{\centering\arraybackslash}m{2.5cm}}
            \toprule[1.5px]
            \textbf{Attribute} & \textbf{Model} & \textbf{Size} & \textbf{Acc (\%)} & \textbf{mPre} & \textbf{mRec} & \textbf{mF1} & \textbf{Acc@2 (\%)} \\
            \midrule[1.5px]
            \multirow{9}{2cm}{\centering\textbf{Building type}} 
            & \multicolumn{7}{l}{\textit{Zero-shot}} \\ \cline{2-8}
            & ChatGPT-4o         & –    & 57.69   & 0.64 & 0.58 & 0.57 & 75.06 \\
            & \multirow{2}{2cm}{\centering InternVL3} 
            & 1B                 & 44.31   & 0.60 & 0.41 & 0.41 & 50.60 \\
            &                    & 2B                 & 46.99   & 0.58 & 0.46 & 0.46 & 61.38 \\
            & InternVL2.5        & 4B                 & 47.82   &  0.61 &  0.46 & 0.46 &  63.06 \\
            \cline{2-8}
            & \multicolumn{7}{l}{\textit{Fine-tuned}} \\ \cline{2-8} 
            & \multirow{2}{2cm}{\centering InternVL3} 
            & 1B                 & 60.87   &  0.65 & 0.60 &   0.60      &     77.37 \\
            &                    & 2B                 & 61.27   & 0.66 & 0.60 & 0.61 & 77.31 \\
            & InternVL2.5        & 4B                 & 62.41   &  0.67  & 0.61 & 0.62 &  76.69 \\
            \midrule[1.5px]
            \multirow{9}{2cm}{\centering\textbf{Surface material}} 
            & \multicolumn{7}{l}{\textit{Zero-shot}} \\ \cline{2-8}
            & ChatGPT-4o         & –    & 65.41& 0.58 & 0.61& 0.55  &   79.07 \\
            & \multirow{2}{2cm}{\centering InternVL3} 
            & 1B                 & 56.64   & 0.62 & 0.48 & 0.43 & 62.16 \\
            &                    & 2B                 & 61.65   & 0.55 & 0.55 & 0.52 & 69.55 \\
            & InternVL2.5        & 4B                 & 60.65    &   0.51   & 0.53   & 0.50 & 76.15 \\
            \cline{2-8}
            & \multicolumn{7}{l}{\textit{Fine-tuned}} \\ \cline{2-8}
            & \multirow{2}{2cm}{\centering InternVL3} 
            & 1B                 & 67.79   & 0.61 & 0.64  & 0.59 & 81.20 \\
            &                    & 2B                 & 68.05   & 0.60 & 0.63 & 0.58 & 81.33 \\
            & InternVL2.5        & 4B                 & 68.55 & 0.61 & 0.64 & 0.59  & 81.95 \\
            \bottomrule[1.5px]
            \end{tabular}}
        \label{tab_model_class}
    \end{subtable}
    
    \vspace{1em}
    
    \begin{subtable}[t]{0.8\textwidth}
        \centering
        \caption{Building floors and building age prediction in zero-shot and fine-tuned settings.}
        \resizebox{\textwidth}{!}{
        \renewcommand{\arraystretch}{1.1}
        \begin{tabular}{>{\centering\arraybackslash}m{2cm}
                        >{\centering\arraybackslash}m{2.5cm}
                        >{\centering\arraybackslash}m{2cm}
                        >{\centering\arraybackslash}m{2cm}
                        >{\centering\arraybackslash}m{2cm}
                        >{\centering\arraybackslash}m{2cm}
                        >{\centering\arraybackslash}m{2cm}}
            \toprule[1.5px]
            \textbf{Attribute} & \textbf{Model} & \textbf{Size} & \textbf{R2 ($\uparrow$)} & \textbf{MAE ($\downarrow$)} & \textbf{MAPE ($\downarrow$)} & \textbf{RMSE ($\downarrow$)} \\
            \midrule[1.5px]
            \multirow{9}{2cm}{\centering\textbf{Building floors}} 
            & \multicolumn{6}{l}{\textit{Zero-shot}} \\ \cline{2-7}
            & ChatGPT-4o                  & -                  & 0.721              & 2.36         & 38.66         & 5.01 \\
            & \multirow{2}{2.5cm}{\centering InternVL3} 
            & 1B                 & -3.071 & 8.82 & 105.22 & 19.04 \\
            &                             & 2B                 & 0.624 &  3.38  & 50.41 &   5.81 \\
            & InternVL2.5                   & 4B                 & 0.548             & 3.68         & 44.30         & 6.53 \\
            \cline{2-7}
            & \multicolumn{6}{l}{\textit{Fine-tuned}} \\ \cline{2-7}
            & \multirow{2}{2.5cm}{\centering InternVL3} 
            & 1B                 & 0.778 & 2.16 &  37.69 &  4.46 \\
            &                             & 2B       & 0.789 &  2.13  & 36.92 &  4.35  \\
            & InternVL2.5                   & 4B   & 0.771 & 2.32 & 35.83  &  4.53 \\
            \midrule[1.5px]
            \multirow{9}{2cm}{\centering\textbf{Building age}} 
            & \multicolumn{6}{l}{\textit{Zero-shot}} \\ \cline{2-7}
            & ChatGPT-4o                  & -                  & 0.645             & 31.63        & 58.79         & 57.07 \\
            & \multirow{2}{2.5cm}{\centering InternVL3} 
            & 1B                 & 0.091  &  57.93  &   77.44  &  91.19 \\
            &                             & 2B                 & 0.560 & 41.05  &  77.10  & 63.30 \\
            & InternVL2.5                 & 4B                 & 0.242 & 51.93 &  85.25 & 84.12 \\
            \cline{2-7}
            & \multicolumn{6}{l}{\textit{Fine-tuned}} \\ \cline{2-7}
            & \multirow{2}{2.5cm}{\centering InternVL3} 
            & 1B                 & 0.713 &  28.09  &  58.15  & 51.36 \\
            &                             & 2B   & 0.710 & 28.05 &  58.27 & 51.50 \\
            & InternVL2.5                   & 4B   & 0.707 &  29.11 &  63.73 & 51.85 \\
            \bottomrule[1.5px]
        \end{tabular}}
        \label{tab_model_reg}
    \end{subtable}
    
    \label{tab:combined}
\end{table}

Table~\ref{tab_model_text} presents captioning evaluation metrics for various models. As with the labeling task, pretrained models already exhibit reasonable performance relative to the ChatGPT-4o reference captions, while fine-tuned models incorporate domain-specific knowledge and produce more coherent, better-structured captions. 
We also observe that performance generally improves with model size, but doubling the parameter count yields diminishing marginal gains, which is likely a consequence of the dataset’s limited scale and the inherent noise in OSM derived ground truth labels. 
To balance computational cost and accuracy, we therefore select the InternVL3 model with 2 billion parameters for this study.

\begin{table}[ht]
    \centering
    \caption{Captioning performance metrics (\%) in zero-shot and fine-tuned settings, evaluated against ChatGPT-4o-generated reference captions.}
    \resizebox{.9\textwidth}{!}{
    \renewcommand{\arraystretch}{1.1}
    \begin{tabular}{>{\centering\arraybackslash}m{3cm}
                    >{\centering\arraybackslash}m{1cm}
                    >{\centering\arraybackslash}m{1.8cm}
                    >{\centering\arraybackslash}m{1.8cm}
                    >{\centering\arraybackslash}m{1.8cm}
                    >{\centering\arraybackslash}m{1.8cm}
                    >{\centering\arraybackslash}m{1.8cm}
                    >{\centering\arraybackslash}m{2cm}}
        \toprule[1.5px]
        \textbf{Model} & \textbf{Size} & \textbf{BLEU-1} & \textbf{BLEU-2} & \textbf{BLEU-3} & \textbf{BLEU-4} & \textbf{METEOR} & \textbf{ROUGE-L} \\
        \midrule[1.5px]
        \multicolumn{8}{l}{\textit{Zero-shot}} \\ \cmidrule(l){1-8}
        InternVL3     & 1B & 46.99 & 29.55 & 19.64 & 13.09 & 36.56 & 29.64 \\
        InternVL3     & 2B & 36.47 & 22.61 & 15.16 & 10.21 & 28.28 & 27.13 \\
        InternVL2.5   & 4B & 44.14 & 28.81 & 20.27 & 14.39 & 37.99 & 31.13 \\
        \midrule[1.5px]
        \multicolumn{8}{l}{\textit{Fine-tuned}} \\ \cmidrule(l){1-8}
        InternVL3     & 1B & 56.60 & 41.62 & 32.65 & 25.99 & 44.46 & 43.31 \\
        InternVL3     & 2B & 56.91 & 41.91 & 32.93 & 26.25 & 44.91 & 43.55 \\
        InternVL2.5   & 4B & 52.32 & 37.68 & 29.00 & 22.64 & 41.02 & 40.43 \\
        \bottomrule[1.5px]
    \end{tabular}}
    \label{tab_model_text}
\end{table}

\begin{figure}[!]
    \centering
    \includegraphics[width=1\linewidth]{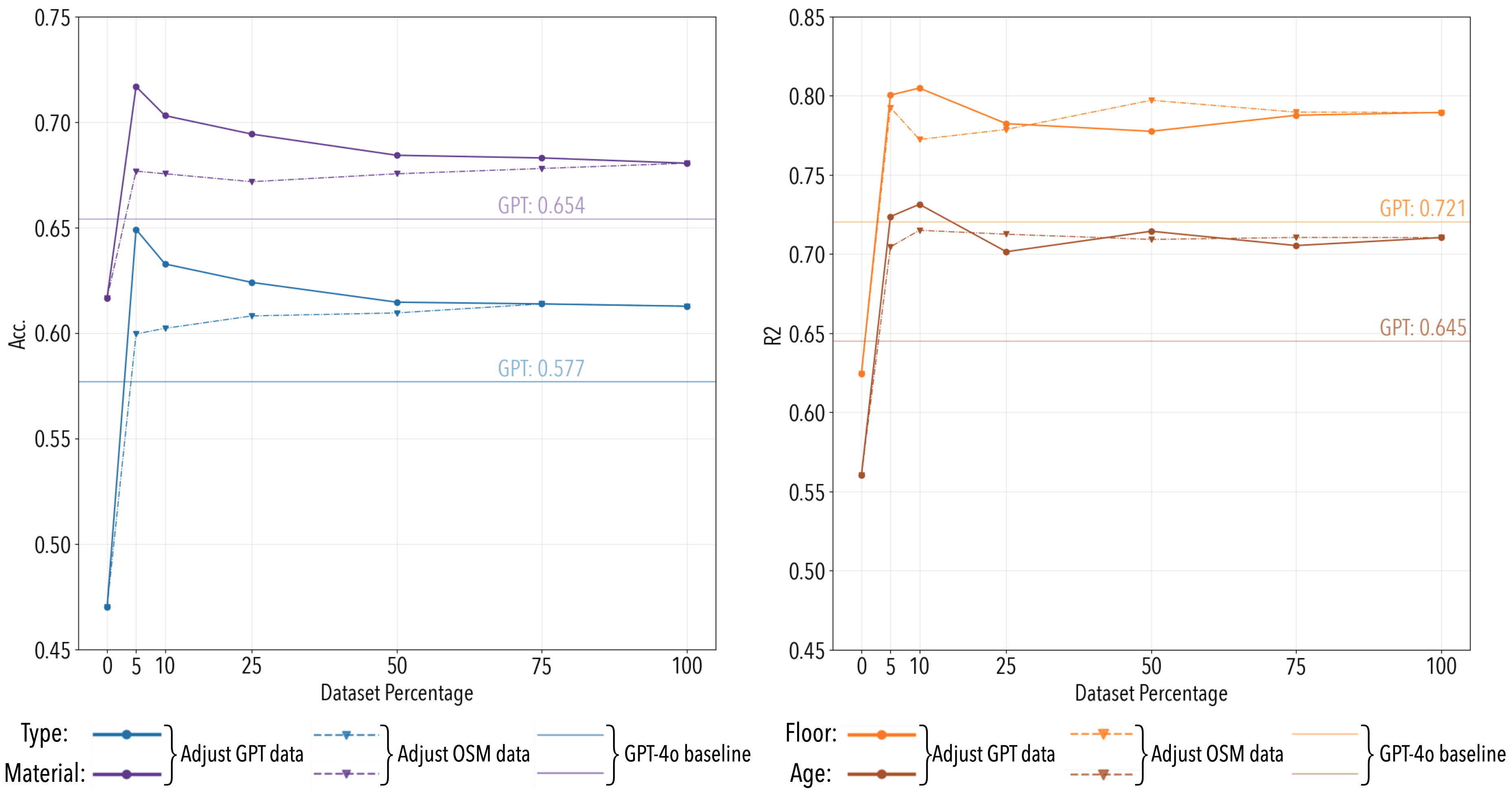}
    \caption{Model performance across varying dataset sizes by adjusting full training data (solid line) and GPT-generated data (dashed lines). The left plot shows accuracy for building type and surface material classification, while the right plot presents R-squared values for floor and age predictions, benchmarked against ChatGPT baselines.}
    \label{fig_ablation}
\end{figure}

\paragraph{Data size.}
To assess the impact of training set size on performance, we trained InternVL3-2B on varying fractions of the 58,942 image–text pairs listed in Table~\ref{tab:data_summary}. 
The left panel of Figure~\ref{fig_ablation} reports accuracy for material and type predictions, while the right panel shows R$^2$ values for floor and age predictions at each dataset proportion. 
Table~\ref{tab_ablation_text} presents BLEU, METEOR, and ROUGE-L scores relative to GPT-generated captions under each condition.

These ablation experiments reveal several important insights.
First, in multi-attribute prediction tasks (Figure~\ref{fig_ablation}), performance peaks early in both scenarios of adjusting either OSM data (single-attribute Q\&A) or GPT-generated data (multi-attribute Q\&A and image captions). Even smaller datasets (around 5–10\% of the full corpus) yield notable performance gains, highlighting the VLM’s ability to learn effectively in data-constrained scenarios. This behavior can be attributed to the pre-trained semantic relationships embedded in the VLM’s latent space from its foundational training. 
Fine-tuning on limited data stabilizes outputs by aligning task-specific features with the model’s pre-existing knowledge distribution.
Second, performance rises gradually when adding OSM ground truth data for most attribute prediction tasks, while GPT-generated data slightly diminishes performance gains. 
One possible explanation is that OSM data encodes structured, human‐validated geographic knowledge, whereas GPT‐generated samples may introduce inaccuracies or hallucinated features. 
Mitigating such noise by refining annotation procedures or excluding low‐quality samples could improve overall accuracy and robustness \citep{chen2024expanding}. 
Third, integrating single-attribute Q\&A data derived from OSM labels appears to constrain descriptive richness across tasks (Table~\ref{tab_ablation_text}). This trade-off likely reflects task interference in a multi-task learning setup, where optimizing for structured attribute prediction can suppress the model’s ability to generate diverse captions. 
To address this limitation, one could increase model capacity, curate high-quality OSM–GPT hybrid datasets, or employ techniques such as knowledge distillation to balance structured output with generative expressiveness.

\begin{table}[ht]
    \centering
    \caption{BLEU, METEOR, and ROUGE-L evaluation across OSM and GPT splits.}
    \resizebox{0.95\textwidth}{!}{
    \renewcommand{\arraystretch}{1.1}
    \begin{tabular}{c c c c c c c c}
        \toprule[1.5px]
        \multicolumn{2}{c}{\textbf{Dataset}} &  \multirow{2}{2cm}{\centering\textbf{BLEU-1}} &  \multirow{2}{2cm}{\centering\textbf{BLEU-2}} &  \multirow{2}{2cm}{\centering\textbf{BLEU-3}} & \multirow{2}{2cm}{\centering\textbf{BLEU-4}} & \multirow{2}{2cm}{\centering\textbf{METEOR}} & \multirow{2}{2cm}{\centering\textbf{ROUGE-L}} \\
        \cmidrule(lr){1-2}
        \textbf{OSM} & \textbf{GPT} & & & & & & \\
        \midrule[1.5px]
        - & - & 36.35 & 22.54 & 15.11 & 10.16 & 28.32 & 27.11 \\ \hline
        5\% & 100\% & 57.17 & 42.09 & 33.03 & 26.31 & 44.97 & 43.53 \\
        10\% & 100\% & 57.50 & 42.40 & 33.33 & 26.59 & 45.29 & 43.74 \\
        25\% & 100\% & 57.25 & 42.17 & 33.12 & 26.39 & 45.07 & 43.63 \\
        50\% & 100\% & 56.71 & 41.73 & 32.75 & 26.09 & 44.62 & 43.35 \\
        75\% & 100\% & 56.97 & 41.92 & 32.89 & 26.18 & 44.88 & 43.39 \\ \hline
        100\% & 5\% & 53.48 & 37.78 & 28.57 & 21.87 & 41.12 & 39.25 \\
        100\% & 10\% & 54.36 & 38.62 & 29.37 & 22.64 & 41.67 & 40.02 \\
        100\% & 25\% & 54.06 & 38.75 & 29.71 & 23.08 & 41.78 & 40.65 \\
        100\% & 50\% & 55.24 & 40.13 & 31.13 & 24.50 & 43.02 & 41.92 \\
        100\% & 75\% & 55.92 & 41.26 & 32.08 & 25.59 & 44.09 & 42.84 \\ \hline
        100\% & 100\% & 56.84 & 41.82 & 32.81 & 26.13 & 44.81 & 43.40 \\
        \bottomrule[1.5px]
    \end{tabular}
    }
    \label{tab_ablation_text}
\end{table}

In summary, our ablations show that (1) full fine-tuning of open-source MLLMs delivers stronger domain adaptation than adapter-based methods, with fewer epochs and smaller learning rates achieving comparable knowledge gains; (2) small-scale models strike the best efficiency–performance balance, while larger models maintain an edge in free-form captioning; and (3) a modest, well-curated dataset captures most of the benefits of large-scale pretraining, although over-emphasis on structured labels can modestly reduce generative richness. Together, these findings underscore the importance of harmonizing fine-tuning scope, model capacity, data quality, and training objectives when extending vision–language models to building-specific tasks.

\section{Discussion}
\subsection{Image labeling and captioning}
\begin{figure}[!]
    \centering
    \includegraphics[width=.85\linewidth]{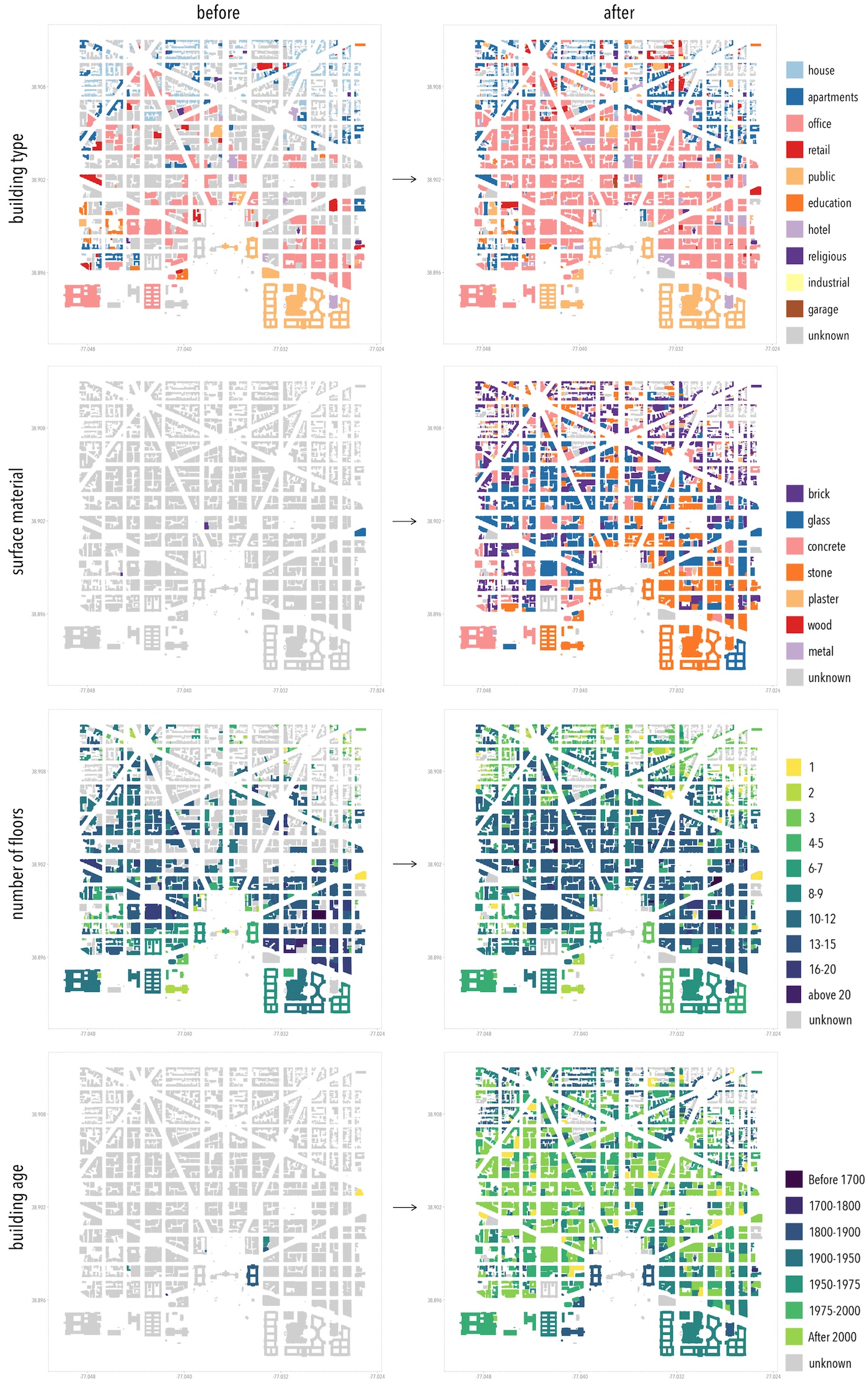}
    \caption{Comparison of OSM building data (left) and the building attributes inferred using our method (right) in Washington D.C., illustrating attributes: building type, surface material, number of floors, and age. Data: (c) OpenStreetMap contributors.}
    \label{fig_map}
\end{figure}

Detected buildings across seven global cities, introduced in Section~\ref{sc_dataset}, are subsequently processed by the fine‐tuned VLM to generate objective attributes and captions. 
Overall, data for half a million buildings are enriched using 1.2 million images, each linked to its geographical location. For buildings with multiple observations, the most frequently assigned categories are retained.
Figure~\ref{fig_map} compares the availability of building properties before and after enrichment in Washington, D.C. 
The proposed approach effectively enhances building‐level information, particularly for surface material and building age. 
The supplementary material presents the distribution of class labels for each attribute across the 1.2 million‐building dataset.

\begin{figure}[!]
    \centering
    \includegraphics[width=.85\linewidth]{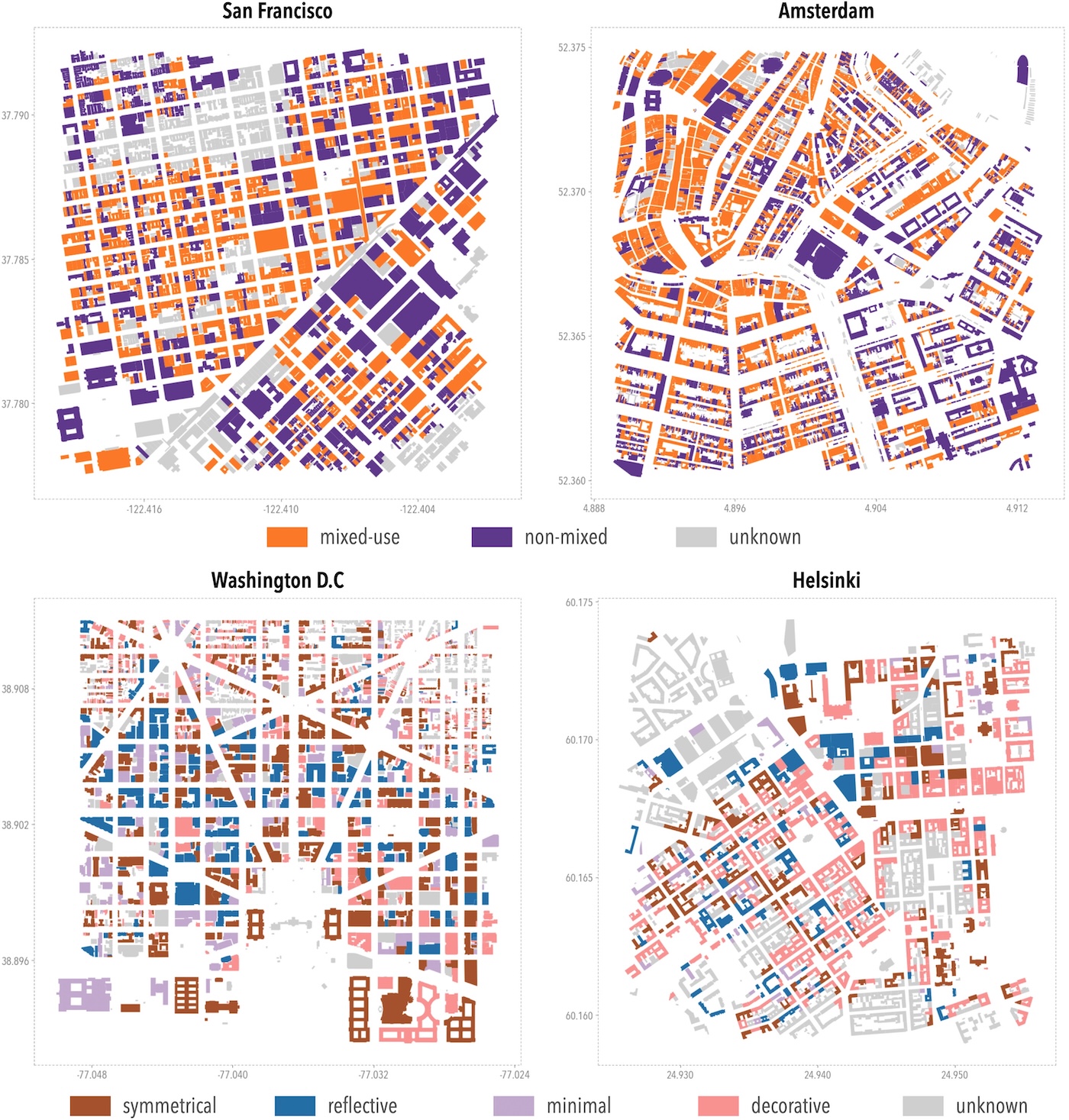}
    \caption{Spatial distribution of mixed-use buildings (top) and facade styles (bottom) in different cities. Data: (c) OpenStreetMap contributors.}
    \label{fig_map_text}
\end{figure}

Beyond the predefined labels, our dataset includes text annotations for each building image, providing a richer source of information for categorizing architectural features. These captions capture intricate details beyond standard classifications, including facade styles, structural elements, and mixed-use characteristics, offering a more nuanced understanding of urban form.
By extracting key descriptors, Figure~\ref{fig_map_text} showcases examples of mixed-use buildings and diverse facade styles identified in Washington, D.C., and San Francisco. This methodology introduces additional dimensionalities for architectural feature analysis, allowing for more detailed characterizations of urban landscapes. 
Moreover, it facilitates fine-grained comparisons across cities, helping to reveal and interpret regional architectural trends and stylistic variations. 

\subsection{Application of building image dataset}
Despite the centrality of objective building attributes in urban analytics, their scarcity still persists across cities~\citep{biljecki2023quality}.
Our open-source framework OpenFACADES addresses this gap by combining SVI, which captures pedestrian-scale visuals, with building data to train an MLLM for unified attribute extraction and semantic description.
The methodology begins by integrating crowdsourced SVI metadata with geometrical building data using isovist analysis to identify relevant images.
Buildings are then detected based on their angles of view within image space, followed by an automated process of reprojecting and filtering them into individual building images.
Lastly, a subset of this dataset is used to construct an image-text dataset designed for three tasks for VLM fine-tuning: single-attribute Q\&A, multi-attribute Q\&A, and captioning.
Our experiments indicate that the fine-tuned VLM demonstrates strong performance in multi-attribute prediction, surpassing CV models and outperforming zero-shot ChatGPT-4o baselines.
Deploying the VLM at scale, we annotate and release data of half a million buildings with both objective attributes and textual descriptions, derived from 1.2 million images across seven global cities, contributing to a scalable and automated approach for building property enrichment.

Our study directly features three main contributions to building research. 
First, our methodology detects holistic building facades and reprojects them into undistorted individual images, ensuring comprehensive visual coverage while reducing the uncertainty inherent in panoramic imagery. 
This pipeline can be integrated with existing methods to detect buildings from diverse viewing angles and associate them with geolocation, enabling nuanced and holistic observation for exterior modeling~\citep{zhang2021vgi3d}, facade material segmentation~\citep{tarkhan_mapping_2025}, and window-to-wall ratio calculation~\citep{de2024window}.
Second, this work introduces an inclusive and efficient pipeline to utilize both crowdsourced data and open-sourced LLMs for street-level research. 
This pipeline not only overcomes the challenge of relying on proprietary datasets, but also circumvents the high costs and limited adaptability associated with proprietary LLM APIs, making advanced analytical techniques more accessible and reproducible to the research community.
Future studies might adjust the pipeline to customized tasks to incorporate fine-grained visual information with tailored building data based on their objectives, such as building conditions~\citep{zou_detecting_2021, zhang2024archgpt}, human perceptual indicators~\citep{liang2024evaluating} and seismic structural types~\citep{pelizari2021automated}.  

Third, we present unified benchmark VLMs that perform multi‐task learning on building facades, generating descriptive captions while maintaining robust multi‐class predictions of objective attributes. In particular, we:

\begin{itemize}
\item Demonstrate that full fine-tuning of a open-source VLM backbone yields state-of-the-art multi-attribute extraction. It surpass zero-shot ChatGPT and matching or exceeding specialized CNN/ViT baselines on building type, material, floor count and age, simultaneously generating coherent captions within one unified framework. The model delivers consistent performance across categories and cities.

\item Reveal generalizability and robustness. In cross-city evaluations and synthetic corruption tests, our fine-tuned VLM surpasses CV models by maintaining high accuracy on unseen urban contexts and showing resilience to occlusion, blur, and lighting distortions, underscoring its suitability for heterogeneous, crowdsourced SVI.

\item Investigate efficient training paradigms. Through systematic ablations, we show that (i) full fine-tuning of open-source VLMs with a low learning rate over a few epochs, (ii) the use of a small-scale backbone, and (iii) a well-curated, balanced dataset together capture most benefits of large-scale pretraining by minimizing data, compute, and architectural complexity while preserving accurate multi-attribute prediction and high-quality captioning.
\end{itemize}

Finally, we apply the pipeline at scale by generating labels and captions for half a million buildings in seven cities, laying a foundation for future urban analyses. For instance, integrating these labeled data with geospatial information can add new dimensions to urban functional zone classification~\citep{zhang_knowledge_2023}, including potential insights into 3D functional zoning~\citep{lin2024does}. The unified model also infers multi‐dimensional building properties relevant for applications such as modeling building electricity consumption~\citep{rosenfelder2021predicting}, estimating material stocks~\citep{raghu_towards_2023}, and assessing structural risk~\citep{wang2021automatic}. Additionally, captions offer an extra layer of information about building facades, enabling the identification of mixed‐use buildings or stylistic variations. This linguistic data holds promise for exploring urban identity, supporting text‐image‐based generative design, and serving as an additional feature layer in multimodal model training.

\subsection{Limitations and future work}
Despite the advancements presented in this study, limitations remain.
First, while this study incorporates captioning data for fine-tuning VLMs, these captions are generated using commercial state-of-the-art LLMs rather than human-labeled ground truth, leaving their accuracy and reliability unverified. 
A systematic human evaluation would be valuable for future research to assess captioning quality, consistency, and semantic accuracy. 
Additionally, leveraging open-access models offers a more sustainable approach for scalable dataset expansion in future studies. 
Knowledge distillation, in which a smaller model learns from a larger teacher, offers a promising self-supervised approach to improve generalization across urban contexts.

Second, while the fine-tuned model exhibits strong generalizability across cities, the quality of crowdsourced data remains a crucial factor~\citep{biljecki2023quality, hou_comprehensive_2022}. Although multiple strategies were employed in this study to mitigate data quality issues, several challenges persist. These include incorrect or incomplete building labels, inconsistent geometric information, non-standardized image formats, and misaligned image coordinates, each contributing to various uncertainties. Additionally, as no manual image selection was performed in this study, potential biases in data collection remain unaddressed.
Future work should focus on enhancing dataset reliability through improved data filtering mechanisms. Automated repetition detection, heuristic rule-based filtering, and uncertainty-aware sampling could refine image selection and minimize inconsistencies in building attribute annotations~\citep{chen2024expanding}.

\section{Conclusion}
This comprehensive study advances spatial data infrastructures and urban data science by introducing a novel framework, OpenFACADES, which leverages volunteered geographic information to enrich building profiles on a global scale using street-level imagery and multimodal large language models. We harvest multimodal crowdsourced data and apply isovist analysis, object detection, and a tailored reprojection method to geolocate and acquire holistic building images, thereby establishing a comprehensive global building image dataset. A selection of this open dataset is then utilized for fine-tuning VLMs, enabling large-scale enrichment of building profiles through multi-attribute prediction and open-vocabulary captioning.
This framework provides a scalable solution for capturing multi-dimensional fine-grained architectural details and urban morphological characteristics. 

Our findings also demonstrate that VLMs generally outperform conventional CNN-based models and zero-shot GPT-4o baselines in predicting building attributes while generating linguistically grounded descriptions. 
This methodological advancement has enabled the creation of a large-scale dataset covering half a million buildings across seven global cities.
The enriched dataset further facilitates a more nuanced and expansive exploration of urban environments, with potential applications in energy modeling, risk assessment, and sustainable development. 

Beyond its immediate applications, we envision this framework as a foundation for comprehensive building profiling, capturing not only physical attributes but also the socio-economic and cultural narratives embedded within the built environment. 
This advancement has significant implications for urban research, including large-scale built environment analysis, building simulation, and policy-driven planning strategies.

\section*{Acknowledgments}
This research is part of the project Large-scale 3D Geospatial Data for Urban Analytics, which is supported by the National University of Singapore under the Start Up Grant.
This research is part of the project Multi-scale Digital Twins for the Urban Environment: From Heartbeats to Cities, which is supported by the Singapore Ministry of Education Academic Research Fund Tier 1.
The first author acknowledges the NUS Graduate Research Scholarship granted by the National University of Singapore (NUS).
We thank the members of the NUS Urban Analytics Lab for the discussions.
We also acknowledge the contributors of OpenStreetMap, Mapillary and other platforms for providing valuable open data resources and code that support street-level imagery research and applications.

\section*{Author contributions}
X.L.: Conceptualization, Methodology, Software, Validation, Formal analysis, Investigation, Data Curation, Writing - Original Draft, Visualization, Project administration.
J.X.: Methodology, Software, Resources, Writing - Review \& Editing.
T.Z.: Methodology, Resources, Writing - Review \& Editing.
R.S.: Writing - Review \& Editing.
F.B.: Conceptualization, Resources, Writing - Review \& Editing, Supervision, Funding acquisition.

\appendix
\section{Supplementary material}
The following is the Supplementary material related to this article. \href{https://ars.els-cdn.com/content/image/1-s2.0-S0924271625004022-mmc1.pdf}{[Link]}

\bibliographystyle{elsarticle-harv} 
\bibliography{Reference,references_zotero}

\begin{thebibliography}{99}
\expandafter\ifx\csname natexlab\endcsname\relax\def\natexlab#1{#1}\fi
\providecommand{\url}[1]{\texttt{#1}}
\providecommand{\href}[2]{#2}
\providecommand{\path}[1]{#1}
\providecommand{\DOIprefix}{doi:}
\providecommand{\ArXivprefix}{arXiv:}
\providecommand{\URLprefix}{URL: }
\providecommand{\Pubmedprefix}{pmid:}
\providecommand{\doi}[1]{\href{http://dx.doi.org/#1}{\path{#1}}}
\providecommand{\Pubmed}[1]{\href{pmid:#1}{\path{#1}}}
\providecommand{\bibinfo}[2]{#2}
\ifx\xfnm\relax \def\xfnm[#1]{\unskip,\space#1}\fi
\bibitem[{Aksoezen et~al.(2015)Aksoezen, Daniel, Hassler and
  Kohler}]{aksoezen2015building}
\bibinfo{author}{Aksoezen, M.}, \bibinfo{author}{Daniel, M.},
  \bibinfo{author}{Hassler, U.}, \bibinfo{author}{Kohler, N.},
  \bibinfo{year}{2015}.
\newblock \bibinfo{title}{Building age as an indicator for energy consumption}.
\newblock \bibinfo{journal}{Energy and Buildings} \bibinfo{volume}{87},
  \bibinfo{pages}{74--86}.
\bibitem[{Al~Rahhal et~al.(2022)Al~Rahhal, Bazi, Alsaleh, Al-Razgan, Mekhalfi,
  Al~Zuair and Alajlan}]{al2022open}
\bibinfo{author}{Al~Rahhal, M.M.}, \bibinfo{author}{Bazi, Y.},
  \bibinfo{author}{Alsaleh, S.O.}, \bibinfo{author}{Al-Razgan, M.},
  \bibinfo{author}{Mekhalfi, M.L.}, \bibinfo{author}{Al~Zuair, M.},
  \bibinfo{author}{Alajlan, N.}, \bibinfo{year}{2022}.
\newblock \bibinfo{title}{Open-ended remote sensing visual question answering
  with transformers}.
\newblock \bibinfo{journal}{International Journal of Remote Sensing}
  \bibinfo{volume}{43}, \bibinfo{pages}{6809--6823}.
\bibitem[{Banerjee and Lavie(2005)}]{banerjee2005meteor}
\bibinfo{author}{Banerjee, S.}, \bibinfo{author}{Lavie, A.},
  \bibinfo{year}{2005}.
\newblock \bibinfo{title}{Meteor: An automatic metric for mt evaluation with
  improved correlation with human judgments}, in:
  \bibinfo{booktitle}{Proceedings of the acl workshop on intrinsic and
  extrinsic evaluation measures for machine translation and/or summarization},
  pp. \bibinfo{pages}{65--72}.
\bibitem[{Biljecki et~al.(2021)Biljecki, Chew, Milojevic-Dupont and
  Creutzig}]{biljecki_open_2021}
\bibinfo{author}{Biljecki, F.}, \bibinfo{author}{Chew, L.Z.X.},
  \bibinfo{author}{Milojevic-Dupont, N.}, \bibinfo{author}{Creutzig, F.},
  \bibinfo{year}{2021}.
\newblock \bibinfo{title}{Open government geospatial data on buildings for
  planning sustainable and resilient cities}.
\newblock \URLprefix \url{http://arxiv.org/abs/2107.04023},
  \DOIprefix\doi{10.48550/arXiv.2107.04023}. \bibinfo{note}{arXiv:2107.04023}.
\bibitem[{Biljecki et~al.(2023)Biljecki, Chow and Lee}]{biljecki2023quality}
\bibinfo{author}{Biljecki, F.}, \bibinfo{author}{Chow, Y.S.},
  \bibinfo{author}{Lee, K.}, \bibinfo{year}{2023}.
\newblock \bibinfo{title}{Quality of crowdsourced geospatial building
  information: A global assessment of openstreetmap attributes}.
\newblock \bibinfo{journal}{Building and Environment} \bibinfo{volume}{237},
  \bibinfo{pages}{110295}.
\bibitem[{Biljecki and Ito(2021)}]{biljecki2021street}
\bibinfo{author}{Biljecki, F.}, \bibinfo{author}{Ito, K.},
  \bibinfo{year}{2021}.
\newblock \bibinfo{title}{Street view imagery in urban analytics and gis: A
  review}.
\newblock \bibinfo{journal}{Landscape and Urban Planning}
  \bibinfo{volume}{215}, \bibinfo{pages}{104217}.
\bibitem[{Boeing(2017)}]{boeing2017osmnx}
\bibinfo{author}{Boeing, G.}, \bibinfo{year}{2017}.
\newblock \bibinfo{title}{Osmnx: New methods for acquiring, constructing,
  analyzing, and visualizing complex street networks}.
\newblock \bibinfo{journal}{Computers, environment and urban systems}
  \bibinfo{volume}{65}, \bibinfo{pages}{126--139}.
\bibitem[{Boguszewski et~al.(2021)Boguszewski, Batorski, Ziemba-Jankowska,
  Dziedzic and Zambrzycka}]{boguszewski2021landcover}
\bibinfo{author}{Boguszewski, A.}, \bibinfo{author}{Batorski, D.},
  \bibinfo{author}{Ziemba-Jankowska, N.}, \bibinfo{author}{Dziedzic, T.},
  \bibinfo{author}{Zambrzycka, A.}, \bibinfo{year}{2021}.
\newblock \bibinfo{title}{Landcover. ai: Dataset for automatic mapping of
  buildings, woodlands, water and roads from aerial imagery}, in:
  \bibinfo{booktitle}{Proceedings of the IEEE/CVF Conference on Computer Vision
  and Pattern Recognition}, pp. \bibinfo{pages}{1102--1110}.
\bibitem[{Chen et~al.(2022)Chen, Subedi, Jahanshahi, Johnson and
  Delp}]{chen_deep_2022}
\bibinfo{author}{Chen, F.C.}, \bibinfo{author}{Subedi, A.},
  \bibinfo{author}{Jahanshahi, M.R.}, \bibinfo{author}{Johnson, D.R.},
  \bibinfo{author}{Delp, E.J.}, \bibinfo{year}{2022}.
\newblock \bibinfo{title}{Deep {Learning}–{Based} {Building} {Attribute}
  {Estimation} from {Google} {Street} {View} {Images} for {Flood} {Risk}
  {Assessment} {Using} {Feature} {Fusion} and {Task} {Relation} {Encoding}}.
\newblock \bibinfo{journal}{Journal of Computing in Civil Engineering}
  \bibinfo{volume}{36}, \bibinfo{pages}{04022031}.
\newblock \URLprefix
  \url{https://ascelibrary.org/doi/10.1061/%28ASCE%29CP.1943-5487.0001025},
  \DOIprefix\doi{10.1061/(ASCE)CP.1943-5487.0001025}. \bibinfo{note}{publisher:
  American Society of Civil Engineers}.
\bibitem[{Chen et~al.(2024a)Chen, Ding and Ye}]{chen_mapping_2024}
\bibinfo{author}{Chen, X.}, \bibinfo{author}{Ding, X.}, \bibinfo{author}{Ye,
  Y.}, \bibinfo{year}{2024}a.
\newblock \bibinfo{title}{Mapping sense of place as a measurable urban
  identity: {Using} street view images and machine learning to identify
  building façade materials}.
\newblock \bibinfo{journal}{Environment and Planning B: Urban Analytics and
  City Science} , \bibinfo{pages}{23998083241279992}\URLprefix
  \url{https://doi.org/10.1177/23998083241279992},
  \DOIprefix\doi{10.1177/23998083241279992}. \bibinfo{note}{publisher: SAGE
  Publications Ltd STM}.
\bibitem[{Chen et~al.(2024b)Chen, Wang, Cao, Liu, Gao, Cui, Zhu, Ye, Tian, Liu
  et~al.}]{chen2024expanding}
\bibinfo{author}{Chen, Z.}, \bibinfo{author}{Wang, W.}, \bibinfo{author}{Cao,
  Y.}, \bibinfo{author}{Liu, Y.}, \bibinfo{author}{Gao, Z.},
  \bibinfo{author}{Cui, E.}, \bibinfo{author}{Zhu, J.}, \bibinfo{author}{Ye,
  S.}, \bibinfo{author}{Tian, H.}, \bibinfo{author}{Liu, Z.}, et~al.,
  \bibinfo{year}{2024}b.
\newblock \bibinfo{title}{Expanding performance boundaries of open-source
  multimodal models with model, data, and test-time scaling}.
\newblock \bibinfo{journal}{arXiv preprint arXiv:2412.05271} .
\bibitem[{Chen et~al.(2024c)Chen, Wu, Wang, Su, Chen, Xing, Zhong, Zhang, Zhu,
  Lu et~al.}]{chen2024internvl}
\bibinfo{author}{Chen, Z.}, \bibinfo{author}{Wu, J.}, \bibinfo{author}{Wang,
  W.}, \bibinfo{author}{Su, W.}, \bibinfo{author}{Chen, G.},
  \bibinfo{author}{Xing, S.}, \bibinfo{author}{Zhong, M.},
  \bibinfo{author}{Zhang, Q.}, \bibinfo{author}{Zhu, X.}, \bibinfo{author}{Lu,
  L.}, et~al., \bibinfo{year}{2024}c.
\newblock \bibinfo{title}{Internvl: Scaling up vision foundation models and
  aligning for generic visual-linguistic tasks}, in:
  \bibinfo{booktitle}{Proceedings of the IEEE/CVF Conference on Computer Vision
  and Pattern Recognition}, pp. \bibinfo{pages}{24185--24198}.
\bibitem[{Creutzig et~al.(2019)Creutzig, Lohrey, Bai, Baklanov, Dawson, Dhakal,
  Lamb, McPhearson, Minx, Munoz et~al.}]{creutzig2019upscaling}
\bibinfo{author}{Creutzig, F.}, \bibinfo{author}{Lohrey, S.},
  \bibinfo{author}{Bai, X.}, \bibinfo{author}{Baklanov, A.},
  \bibinfo{author}{Dawson, R.}, \bibinfo{author}{Dhakal, S.},
  \bibinfo{author}{Lamb, W.F.}, \bibinfo{author}{McPhearson, T.},
  \bibinfo{author}{Minx, J.}, \bibinfo{author}{Munoz, E.}, et~al.,
  \bibinfo{year}{2019}.
\newblock \bibinfo{title}{Upscaling urban data science for global climate
  solutions}.
\newblock \bibinfo{journal}{Global Sustainability} \bibinfo{volume}{2},
  \bibinfo{pages}{e2}.
\bibitem[{Danish et~al.(2025)Danish, Labib, Ricker and
  Helbich}]{danish2025citizen}
\bibinfo{author}{Danish, M.}, \bibinfo{author}{Labib, S.},
  \bibinfo{author}{Ricker, B.}, \bibinfo{author}{Helbich, M.},
  \bibinfo{year}{2025}.
\newblock \bibinfo{title}{A citizen science toolkit to collect human
  perceptions of urban environments using open street view images}.
\newblock \bibinfo{journal}{Computers, Environment and Urban Systems}
  \bibinfo{volume}{116}, \bibinfo{pages}{102207}.
\bibitem[{De~Simone et~al.(2024)De~Simone, Biswas and Wu}]{de2024window}
\bibinfo{author}{De~Simone, Z.}, \bibinfo{author}{Biswas, S.},
  \bibinfo{author}{Wu, O.}, \bibinfo{year}{2024}.
\newblock \bibinfo{title}{Window to wall ratio detection using segformer}.
\newblock \bibinfo{journal}{arXiv preprint arXiv:2406.02706} .
\bibitem[{Dong et~al.(2024)Dong, Wang, Du and Meng}]{dong_changeclip_2024}
\bibinfo{author}{Dong, S.}, \bibinfo{author}{Wang, L.}, \bibinfo{author}{Du,
  B.}, \bibinfo{author}{Meng, X.}, \bibinfo{year}{2024}.
\newblock \bibinfo{title}{{ChangeCLIP}: {Remote} sensing change detection with
  multimodal vision-language representation learning}.
\newblock \bibinfo{journal}{ISPRS Journal of Photogrammetry and Remote Sensing}
  \bibinfo{volume}{208}, \bibinfo{pages}{53--69}.
\newblock \URLprefix
  \url{https://linkinghub.elsevier.com/retrieve/pii/S0924271624000042},
  \DOIprefix\doi{10.1016/j.isprsjprs.2024.01.004}.
\bibitem[{Du et~al.(2015)Du, Zhang and Zhang}]{du_semantic_2015}
\bibinfo{author}{Du, S.}, \bibinfo{author}{Zhang, F.}, \bibinfo{author}{Zhang,
  X.}, \bibinfo{year}{2015}.
\newblock \bibinfo{title}{Semantic classification of urban buildings combining
  {VHR} image and {GIS} data: {An} improved random forest approach}.
\newblock \bibinfo{journal}{ISPRS Journal of Photogrammetry and Remote Sensing}
  \bibinfo{volume}{105}, \bibinfo{pages}{107--119}.
\newblock \URLprefix
  \url{https://www.sciencedirect.com/science/article/pii/S092427161500091X},
  \DOIprefix\doi{10.1016/j.isprsjprs.2015.03.011}.
\bibitem[{Dubey et~al.(2024)Dubey, Jauhri, Pandey, Kadian, Al-Dahle, Letman,
  Mathur, Schelten, Yang, Fan et~al.}]{dubey2024llama}
\bibinfo{author}{Dubey, A.}, \bibinfo{author}{Jauhri, A.},
  \bibinfo{author}{Pandey, A.}, \bibinfo{author}{Kadian, A.},
  \bibinfo{author}{Al-Dahle, A.}, \bibinfo{author}{Letman, A.},
  \bibinfo{author}{Mathur, A.}, \bibinfo{author}{Schelten, A.},
  \bibinfo{author}{Yang, A.}, \bibinfo{author}{Fan, A.}, et~al.,
  \bibinfo{year}{2024}.
\newblock \bibinfo{title}{The llama 3 herd of models}.
\newblock \bibinfo{journal}{arXiv preprint arXiv:2407.21783} .
\bibitem[{Elmqvist et~al.(2019)Elmqvist, Andersson, Frantzeskaki, McPhearson,
  Olsson, Gaffney, Takeuchi and Folke}]{elmqvist_sustainability_2019}
\bibinfo{author}{Elmqvist, T.}, \bibinfo{author}{Andersson, E.},
  \bibinfo{author}{Frantzeskaki, N.}, \bibinfo{author}{McPhearson, T.},
  \bibinfo{author}{Olsson, P.}, \bibinfo{author}{Gaffney, O.},
  \bibinfo{author}{Takeuchi, K.}, \bibinfo{author}{Folke, C.},
  \bibinfo{year}{2019}.
\newblock \bibinfo{title}{Sustainability and resilience for transformation in
  the urban century}.
\newblock \bibinfo{journal}{Nature Sustainability} \bibinfo{volume}{2},
  \bibinfo{pages}{267--273}.
\newblock \URLprefix \url{https://www.nature.com/articles/s41893-019-0250-1},
  \DOIprefix\doi{10.1038/s41893-019-0250-1}. \bibinfo{note}{publisher: Nature
  Publishing Group}.
\bibitem[{Fan et~al.(2024)Fan, Lin, Wu and Xu}]{fan_pano2geo_2024}
\bibinfo{author}{Fan, K.}, \bibinfo{author}{Lin, A.}, \bibinfo{author}{Wu, H.},
  \bibinfo{author}{Xu, Z.}, \bibinfo{year}{2024}.
\newblock \bibinfo{title}{{Pano2Geo}: {An} efficient and robust building height
  estimation model using street-view panoramas}.
\newblock \bibinfo{journal}{ISPRS Journal of Photogrammetry and Remote Sensing}
  \bibinfo{volume}{215}, \bibinfo{pages}{177--191}.
\newblock \URLprefix
  \url{https://linkinghub.elsevier.com/retrieve/pii/S0924271624002727},
  \DOIprefix\doi{10.1016/j.isprsjprs.2024.07.005}.
\bibitem[{Fan et~al.(2025)Fan, Feng and Biljecki}]{fan2025coverage}
\bibinfo{author}{Fan, Z.}, \bibinfo{author}{Feng, C.C.},
  \bibinfo{author}{Biljecki, F.}, \bibinfo{year}{2025}.
\newblock \bibinfo{title}{Coverage and bias of street view imagery in mapping
  the urban environment}.
\newblock \bibinfo{journal}{Computers, Environment and Urban Systems}
  \bibinfo{volume}{117}, \bibinfo{pages}{102253}.
\bibitem[{Feldmeyer et~al.(2020)Feldmeyer, Meisch, Sauter and
  Birkmann}]{feldmeyer2020using}
\bibinfo{author}{Feldmeyer, D.}, \bibinfo{author}{Meisch, C.},
  \bibinfo{author}{Sauter, H.}, \bibinfo{author}{Birkmann, J.},
  \bibinfo{year}{2020}.
\newblock \bibinfo{title}{Using openstreetmap data and machine learning to
  generate socio-economic indicators}.
\newblock \bibinfo{journal}{ISPRS International Journal of Geo-Information}
  \bibinfo{volume}{9}, \bibinfo{pages}{498}.
\bibitem[{Florio et~al.(2025)Florio, Politis, Krasnod{\k{e}}bska, Uhl,
  Melchiorri, Martinez, Kakoulaki, Pesaresi and Kemper}]{florio2025ghs}
\bibinfo{author}{Florio, P.}, \bibinfo{author}{Politis, P.},
  \bibinfo{author}{Krasnod{\k{e}}bska, K.}, \bibinfo{author}{Uhl, J.H.},
  \bibinfo{author}{Melchiorri, M.}, \bibinfo{author}{Martinez, A.M.},
  \bibinfo{author}{Kakoulaki, G.}, \bibinfo{author}{Pesaresi, M.},
  \bibinfo{author}{Kemper, T.}, \bibinfo{year}{2025}.
\newblock \bibinfo{title}{Ghs-obat: Global, open building attribute data
  reporting age, function, height and compactness at footprint level}.
\newblock \bibinfo{journal}{Data in Brief} , \bibinfo{pages}{111751}.
\bibitem[{Frantz et~al.(2021)Frantz, Schug, Okujeni, Navacchi, Wagner, van~der
  Linden and Hostert}]{frantz2021national}
\bibinfo{author}{Frantz, D.}, \bibinfo{author}{Schug, F.},
  \bibinfo{author}{Okujeni, A.}, \bibinfo{author}{Navacchi, C.},
  \bibinfo{author}{Wagner, W.}, \bibinfo{author}{van~der Linden, S.},
  \bibinfo{author}{Hostert, P.}, \bibinfo{year}{2021}.
\newblock \bibinfo{title}{National-scale mapping of building height using
  sentinel-1 and sentinel-2 time series}.
\newblock \bibinfo{journal}{Remote Sensing of Environment}
  \bibinfo{volume}{252}, \bibinfo{pages}{112128}.
\bibitem[{Fujiwara et~al.(2024)Fujiwara, Khomiakov, Yap, Ignatius and
  Biljecki}]{fujiwara2024microclimate}
\bibinfo{author}{Fujiwara, K.}, \bibinfo{author}{Khomiakov, M.},
  \bibinfo{author}{Yap, W.}, \bibinfo{author}{Ignatius, M.},
  \bibinfo{author}{Biljecki, F.}, \bibinfo{year}{2024}.
\newblock \bibinfo{title}{Microclimate vision: Multimodal prediction of
  climatic parameters using street-level and satellite imagery}.
\newblock \bibinfo{journal}{Sustainable Cities and Society}
  \bibinfo{volume}{114}, \bibinfo{pages}{105733}.
\bibitem[{Gaw et~al.(2022)Gaw, Chen, Chow, Lee and Biljecki}]{gaw2022comparing}
\bibinfo{author}{Gaw, L.}, \bibinfo{author}{Chen, S.}, \bibinfo{author}{Chow,
  Y.}, \bibinfo{author}{Lee, K.}, \bibinfo{author}{Biljecki, F.},
  \bibinfo{year}{2022}.
\newblock \bibinfo{title}{Comparing street view imagery and aerial perspectives
  in the built environment}.
\newblock \bibinfo{journal}{ISPRS Annals of the Photogrammetry, Remote Sensing
  and Spatial Information Sciences} \bibinfo{volume}{10},
  \bibinfo{pages}{49--56}.
\bibitem[{Ghione et~al.(2022)Ghione, M{\ae}land, Meslem and
  Oye}]{ghione2022building}
\bibinfo{author}{Ghione, F.}, \bibinfo{author}{M{\ae}land, S.},
  \bibinfo{author}{Meslem, A.}, \bibinfo{author}{Oye, V.},
  \bibinfo{year}{2022}.
\newblock \bibinfo{title}{Building stock classification using machine learning:
  A case study for oslo, norway}.
\newblock \bibinfo{journal}{Frontiers in Earth Science} \bibinfo{volume}{10},
  \bibinfo{pages}{886145}.
\bibitem[{Gouveia et~al.(2024)Gouveia, Silva, Lopes, Moreira, Torres and
  Simas~Guerreiro}]{gouveia2024automated}
\bibinfo{author}{Gouveia, F.}, \bibinfo{author}{Silva, V.},
  \bibinfo{author}{Lopes, J.}, \bibinfo{author}{Moreira, R.S.},
  \bibinfo{author}{Torres, J.M.}, \bibinfo{author}{Simas~Guerreiro, M.},
  \bibinfo{year}{2024}.
\newblock \bibinfo{title}{Automated identification of building features with
  deep learning for risk analysis}.
\newblock \bibinfo{journal}{Discover Applied Sciences} \bibinfo{volume}{6},
  \bibinfo{pages}{466}.
\bibitem[{Gupta et~al.(2019)Gupta, Goodman, Patel, Hosfelt, Sajeev, Heim,
  Doshi, Lucas, Choset and Gaston}]{gupta2019creating}
\bibinfo{author}{Gupta, R.}, \bibinfo{author}{Goodman, B.},
  \bibinfo{author}{Patel, N.}, \bibinfo{author}{Hosfelt, R.},
  \bibinfo{author}{Sajeev, S.}, \bibinfo{author}{Heim, E.},
  \bibinfo{author}{Doshi, J.}, \bibinfo{author}{Lucas, K.},
  \bibinfo{author}{Choset, H.}, \bibinfo{author}{Gaston, M.},
  \bibinfo{year}{2019}.
\newblock \bibinfo{title}{Creating xbd: A dataset for assessing building damage
  from satellite imagery}, in: \bibinfo{booktitle}{Proceedings of the IEEE/CVF
  conference on computer vision and pattern recognition workshops}, pp.
  \bibinfo{pages}{10--17}.
\bibitem[{He et~al.(2016)He, Zhang, Ren and Sun}]{he2016deep}
\bibinfo{author}{He, K.}, \bibinfo{author}{Zhang, X.}, \bibinfo{author}{Ren,
  S.}, \bibinfo{author}{Sun, J.}, \bibinfo{year}{2016}.
\newblock \bibinfo{title}{Deep residual learning for image recognition}, in:
  \bibinfo{booktitle}{Proceedings of the IEEE conference on computer vision and
  pattern recognition}, pp. \bibinfo{pages}{770--778}.
\bibitem[{He et~al.(2024)He, Yao, Shao and Wang}]{he2024ub}
\bibinfo{author}{He, Z.}, \bibinfo{author}{Yao, W.}, \bibinfo{author}{Shao,
  J.}, \bibinfo{author}{Wang, P.}, \bibinfo{year}{2024}.
\newblock \bibinfo{title}{Ub-finenet: Urban building fine-grained
  classification network for open-access satellite images}.
\newblock \bibinfo{journal}{ISPRS Journal of Photogrammetry and Remote Sensing}
  \bibinfo{volume}{217}, \bibinfo{pages}{76--90}.
\bibitem[{Helbich et~al.(2024)Helbich, Danish, Labib and
  Ricker}]{helbich_use_2024}
\bibinfo{author}{Helbich, M.}, \bibinfo{author}{Danish, M.},
  \bibinfo{author}{Labib, S.M.}, \bibinfo{author}{Ricker, B.},
  \bibinfo{year}{2024}.
\newblock \bibinfo{title}{To use or not to use proprietary street view images
  in (health and place) research? {That} is the question}.
\newblock \bibinfo{journal}{Health \& Place} \bibinfo{volume}{87},
  \bibinfo{pages}{103244}.
\newblock \URLprefix
  \url{https://www.sciencedirect.com/science/article/pii/S1353829224000728},
  \DOIprefix\doi{10.1016/j.healthplace.2024.103244}.
\bibitem[{Hendrycks and Dietterich(2019)}]{hendrycks_benchmarking_2019}
\bibinfo{author}{Hendrycks, D.}, \bibinfo{author}{Dietterich, T.},
  \bibinfo{year}{2019}.
\newblock \bibinfo{title}{Benchmarking {Neural} {Network} {Robustness} to
  {Common} {Corruptions} and {Perturbations}} .
\bibitem[{Herfort et~al.(2023)Herfort, Lautenbach, Porto De~Albuquerque,
  Anderson and Zipf}]{herfort_spatio-temporal_2023}
\bibinfo{author}{Herfort, B.}, \bibinfo{author}{Lautenbach, S.},
  \bibinfo{author}{Porto De~Albuquerque, J.}, \bibinfo{author}{Anderson, J.},
  \bibinfo{author}{Zipf, A.}, \bibinfo{year}{2023}.
\newblock \bibinfo{title}{A spatio-temporal analysis investigating completeness
  and inequalities of global urban building data in {OpenStreetMap}}.
\newblock \bibinfo{journal}{Nature Communications} \bibinfo{volume}{14},
  \bibinfo{pages}{3985}.
\newblock \URLprefix \url{https://www.nature.com/articles/s41467-023-39698-6},
  \DOIprefix\doi{10.1038/s41467-023-39698-6}.
\bibitem[{Hou and Biljecki(2022)}]{hou_comprehensive_2022}
\bibinfo{author}{Hou, Y.}, \bibinfo{author}{Biljecki, F.},
  \bibinfo{year}{2022}.
\newblock \bibinfo{title}{A comprehensive framework for evaluating the quality
  of street view imagery}.
\newblock \bibinfo{journal}{International Journal of Applied Earth Observation
  and Geoinformation} \bibinfo{volume}{115}, \bibinfo{pages}{103094}.
\newblock \URLprefix
  \url{https://linkinghub.elsevier.com/retrieve/pii/S1569843222002825},
  \DOIprefix\doi{10.1016/j.jag.2022.103094}.
\bibitem[{Hou et~al.(2024)Hou, Quintana, Khomiakov, Yap, Ouyang, Ito, Wang,
  Zhao and Biljecki}]{hou_global_2024}
\bibinfo{author}{Hou, Y.}, \bibinfo{author}{Quintana, M.},
  \bibinfo{author}{Khomiakov, M.}, \bibinfo{author}{Yap, W.},
  \bibinfo{author}{Ouyang, J.}, \bibinfo{author}{Ito, K.},
  \bibinfo{author}{Wang, Z.}, \bibinfo{author}{Zhao, T.},
  \bibinfo{author}{Biljecki, F.}, \bibinfo{year}{2024}.
\newblock \bibinfo{title}{Global {Streetscapes} — {A} comprehensive dataset
  of 10 million street-level images across 688 cities for urban science and
  analytics}.
\newblock \bibinfo{journal}{ISPRS Journal of Photogrammetry and Remote Sensing}
  \bibinfo{volume}{215}, \bibinfo{pages}{216--238}.
\newblock \URLprefix
  \url{https://linkinghub.elsevier.com/retrieve/pii/S0924271624002612},
  \DOIprefix\doi{10.1016/j.isprsjprs.2024.06.023}.
\bibitem[{Hu et~al.(2025)Hu, Yuan, Wen, Lu, Liu and Li}]{hu2025rsgpt}
\bibinfo{author}{Hu, Y.}, \bibinfo{author}{Yuan, J.}, \bibinfo{author}{Wen,
  C.}, \bibinfo{author}{Lu, X.}, \bibinfo{author}{Liu, Y.},
  \bibinfo{author}{Li, X.}, \bibinfo{year}{2025}.
\newblock \bibinfo{title}{Rsgpt: A remote sensing vision language model and
  benchmark}.
\newblock \bibinfo{journal}{ISPRS Journal of Photogrammetry and Remote Sensing}
  \bibinfo{volume}{224}, \bibinfo{pages}{272--286}.
\bibitem[{Huang et~al.(2017)Huang, Liu, Van Der~Maaten and
  Weinberger}]{huang2017densely}
\bibinfo{author}{Huang, G.}, \bibinfo{author}{Liu, Z.}, \bibinfo{author}{Van
  Der~Maaten, L.}, \bibinfo{author}{Weinberger, K.Q.}, \bibinfo{year}{2017}.
\newblock \bibinfo{title}{Densely connected convolutional networks}, in:
  \bibinfo{booktitle}{Proceedings of the IEEE conference on computer vision and
  pattern recognition}, pp. \bibinfo{pages}{4700--4708}.
\bibitem[{Huang et~al.(2023)Huang, Zhang, Gao, Tu, Duarte, Ratti, Guo and
  Liu}]{huang2023comprehensive}
\bibinfo{author}{Huang, Y.}, \bibinfo{author}{Zhang, F.}, \bibinfo{author}{Gao,
  Y.}, \bibinfo{author}{Tu, W.}, \bibinfo{author}{Duarte, F.},
  \bibinfo{author}{Ratti, C.}, \bibinfo{author}{Guo, D.}, \bibinfo{author}{Liu,
  Y.}, \bibinfo{year}{2023}.
\newblock \bibinfo{title}{Comprehensive urban space representation with varying
  numbers of street-level images}.
\newblock \bibinfo{journal}{Computers, Environment and Urban Systems}
  \bibinfo{volume}{106}, \bibinfo{pages}{102043}.
\bibitem[{Iannelli and Dell’Acqua(2017)}]{iannelli2017extensive}
\bibinfo{author}{Iannelli, G.C.}, \bibinfo{author}{Dell’Acqua, F.},
  \bibinfo{year}{2017}.
\newblock \bibinfo{title}{Extensive exposure mapping in urban areas through
  deep analysis of street-level pictures for floor count determination}.
\newblock \bibinfo{journal}{Urban Science} \bibinfo{volume}{1},
  \bibinfo{pages}{16}.
\bibitem[{Ito et~al.(2025)Ito, Zhu, Abdelrahman, Liang, Fan, Hou, Zhao, Ma,
  Fujiwara, Ouyang et~al.}]{ito2025zensvi}
\bibinfo{author}{Ito, K.}, \bibinfo{author}{Zhu, Y.},
  \bibinfo{author}{Abdelrahman, M.}, \bibinfo{author}{Liang, X.},
  \bibinfo{author}{Fan, Z.}, \bibinfo{author}{Hou, Y.}, \bibinfo{author}{Zhao,
  T.}, \bibinfo{author}{Ma, R.}, \bibinfo{author}{Fujiwara, K.},
  \bibinfo{author}{Ouyang, J.}, et~al., \bibinfo{year}{2025}.
\newblock \bibinfo{title}{Zensvi: An open-source software for the integrated
  acquisition, processing and analysis of street view imagery towards scalable
  urban science}.
\newblock \bibinfo{journal}{Computers, Environment and Urban Systems}
  \bibinfo{volume}{119}, \bibinfo{pages}{102283}.
\bibitem[{Jia et~al.(2024)Jia, Dong, Huang, Chen, Wang, Peng, Guo, Ma, Zhang
  and Liu}]{jia2024transformer}
\bibinfo{author}{Jia, F.}, \bibinfo{author}{Dong, Q.}, \bibinfo{author}{Huang,
  Z.}, \bibinfo{author}{Chen, X.J.}, \bibinfo{author}{Wang, Y.},
  \bibinfo{author}{Peng, X.}, \bibinfo{author}{Guo, Y.}, \bibinfo{author}{Ma,
  R.}, \bibinfo{author}{Zhang, F.}, \bibinfo{author}{Liu, Y.},
  \bibinfo{year}{2024}.
\newblock \bibinfo{title}{A transformer-based multi-modal model for urban-rural
  fringe identification}.
\newblock \bibinfo{journal}{IEEE Journal of Selected Topics in Applied Earth
  Observations and Remote Sensing} .
\bibitem[{Kang et~al.(2018)Kang, Körner, Wang, Taubenböck and
  Zhu}]{kang_building_2018}
\bibinfo{author}{Kang, J.}, \bibinfo{author}{Körner, M.},
  \bibinfo{author}{Wang, Y.}, \bibinfo{author}{Taubenböck, H.},
  \bibinfo{author}{Zhu, X.X.}, \bibinfo{year}{2018}.
\newblock \bibinfo{title}{Building instance classification using street view
  images}.
\newblock \bibinfo{journal}{ISPRS Journal of Photogrammetry and Remote Sensing}
  \bibinfo{volume}{145}, \bibinfo{pages}{44--59}.
\newblock \URLprefix
  \url{https://www.sciencedirect.com/science/article/pii/S0924271618300352},
  \DOIprefix\doi{10.1016/j.isprsjprs.2018.02.006}.
\bibitem[{Kapp et~al.(2025)Kapp, Hoffmann, Weigmann and
  Mihaljević}]{kapp_streetsurfacevis_2025}
\bibinfo{author}{Kapp, A.}, \bibinfo{author}{Hoffmann, E.},
  \bibinfo{author}{Weigmann, E.}, \bibinfo{author}{Mihaljević, H.},
  \bibinfo{year}{2025}.
\newblock \bibinfo{title}{{StreetSurfaceVis}: a dataset of crowdsourced
  street-level imagery annotated by road surface type and quality}.
\newblock \bibinfo{journal}{Scientific Data} \bibinfo{volume}{12},
  \bibinfo{pages}{92}.
\newblock \URLprefix \url{https://www.nature.com/articles/s41597-024-04295-9},
  \DOIprefix\doi{10.1038/s41597-024-04295-9}. \bibinfo{note}{publisher: Nature
  Publishing Group}.
\bibitem[{Kumar et~al.(2018)Kumar, Pal and Singh}]{kumar2018novel}
\bibinfo{author}{Kumar, S.}, \bibinfo{author}{Pal, S.K.},
  \bibinfo{author}{Singh, R.P.}, \bibinfo{year}{2018}.
\newblock \bibinfo{title}{A novel method based on extreme learning machine to
  predict heating and cooling load through design and structural attributes}.
\newblock \bibinfo{journal}{Energy and Buildings} \bibinfo{volume}{176},
  \bibinfo{pages}{275--286}.
\bibitem[{Lei et~al.(2024)Lei, Liu, Milojevic-Dupont and
  Biljecki}]{lei_predicting_2024}
\bibinfo{author}{Lei, B.}, \bibinfo{author}{Liu, P.},
  \bibinfo{author}{Milojevic-Dupont, N.}, \bibinfo{author}{Biljecki, F.},
  \bibinfo{year}{2024}.
\newblock \bibinfo{title}{Predicting building characteristics at urban scale
  using graph neural networks and street-level context}.
\newblock \bibinfo{journal}{Computers, Environment and Urban Systems}
  \bibinfo{volume}{111}, \bibinfo{pages}{102129}.
\newblock \URLprefix
  \url{https://linkinghub.elsevier.com/retrieve/pii/S0198971524000589},
  \DOIprefix\doi{10.1016/j.compenvurbsys.2024.102129}.
\bibitem[{Lei et~al.(2023)Lei, Stouffs and Biljecki}]{lei2023assessing}
\bibinfo{author}{Lei, B.}, \bibinfo{author}{Stouffs, R.},
  \bibinfo{author}{Biljecki, F.}, \bibinfo{year}{2023}.
\newblock \bibinfo{title}{Assessing and benchmarking 3d city models}.
\newblock \bibinfo{journal}{International Journal of Geographical Information
  Science} \bibinfo{volume}{37}, \bibinfo{pages}{788--809}.
\bibitem[{Li et~al.(2025a)Li, Deuser, Yin, Luo, Walther, Mai, Huang and
  Werner}]{li_cross-view_2025}
\bibinfo{author}{Li, H.}, \bibinfo{author}{Deuser, F.}, \bibinfo{author}{Yin,
  W.}, \bibinfo{author}{Luo, X.}, \bibinfo{author}{Walther, P.},
  \bibinfo{author}{Mai, G.}, \bibinfo{author}{Huang, W.},
  \bibinfo{author}{Werner, M.}, \bibinfo{year}{2025}a.
\newblock \bibinfo{title}{Cross-view geolocalization and disaster mapping with
  street-view and {VHR} satellite imagery: {A} case study of {Hurricane}
  {IAN}}.
\newblock \bibinfo{journal}{ISPRS Journal of Photogrammetry and Remote Sensing}
  \bibinfo{volume}{220}, \bibinfo{pages}{841--854}.
\newblock \URLprefix
  \url{https://linkinghub.elsevier.com/retrieve/pii/S0924271625000036},
  \DOIprefix\doi{10.1016/j.isprsjprs.2025.01.003}.
\bibitem[{Li et~al.(2025b)Li, Yu, Chen, Lin, Dong, Zhang, He and
  Fu}]{li_fine-grained_2025}
\bibinfo{author}{Li, W.}, \bibinfo{author}{Yu, J.}, \bibinfo{author}{Chen, D.},
  \bibinfo{author}{Lin, Y.}, \bibinfo{author}{Dong, R.},
  \bibinfo{author}{Zhang, X.}, \bibinfo{author}{He, C.}, \bibinfo{author}{Fu,
  H.}, \bibinfo{year}{2025}b.
\newblock \bibinfo{title}{Fine-grained building function recognition with
  street-view images and {GIS} map data via geometry-aware semi-supervised
  learning}.
\newblock \bibinfo{journal}{International Journal of Applied Earth Observation
  and Geoinformation} \bibinfo{volume}{137}, \bibinfo{pages}{104386}.
\newblock \URLprefix
  \url{https://linkinghub.elsevier.com/retrieve/pii/S1569843225000330},
  \DOIprefix\doi{10.1016/j.jag.2025.104386}.
\bibitem[{Li et~al.(2024a)Li, Wen, Hu, Yuan and Zhu}]{li_vision-language_2024}
\bibinfo{author}{Li, X.}, \bibinfo{author}{Wen, C.}, \bibinfo{author}{Hu, Y.},
  \bibinfo{author}{Yuan, Z.}, \bibinfo{author}{Zhu, X.X.},
  \bibinfo{year}{2024}a.
\newblock \bibinfo{title}{Vision-{Language} {Models} in {Remote} {Sensing}:
  {Current} progress and future trends}.
\newblock \bibinfo{journal}{IEEE Geoscience and Remote Sensing Magazine}
  \bibinfo{volume}{12}, \bibinfo{pages}{32--66}.
\newblock \URLprefix
  \url{https://ieeexplore.ieee.org/document/10506064/?arnumber=10506064},
  \DOIprefix\doi{10.1109/MGRS.2024.3383473}. \bibinfo{note}{conference Name:
  IEEE Geoscience and Remote Sensing Magazine}.
\bibitem[{Li et~al.(2024b)Li, Su, Zhu and Zhao}]{li2024buildingview}
\bibinfo{author}{Li, Z.}, \bibinfo{author}{Su, Y.}, \bibinfo{author}{Zhu, C.},
  \bibinfo{author}{Zhao, W.}, \bibinfo{year}{2024}b.
\newblock \bibinfo{title}{Buildingview: Constructing urban building exteriors
  databases with street view imagery and multimodal large language mode}.
\newblock \bibinfo{journal}{arXiv preprint arXiv:2409.19527} .
\bibitem[{Liang et~al.(2024)Liang, Chang, Gao, Zhao and
  Biljecki}]{liang2024evaluating}
\bibinfo{author}{Liang, X.}, \bibinfo{author}{Chang, J.H.},
  \bibinfo{author}{Gao, S.}, \bibinfo{author}{Zhao, T.},
  \bibinfo{author}{Biljecki, F.}, \bibinfo{year}{2024}.
\newblock \bibinfo{title}{Evaluating human perception of building exteriors
  using street view imagery}.
\newblock \bibinfo{journal}{Building and Environment} \bibinfo{volume}{263},
  \bibinfo{pages}{111875}.
\bibitem[{Lin et~al.(2024)Lin, Wu, Luo, Fan and Liu}]{lin2024does}
\bibinfo{author}{Lin, A.}, \bibinfo{author}{Wu, H.}, \bibinfo{author}{Luo, W.},
  \bibinfo{author}{Fan, K.}, \bibinfo{author}{Liu, H.}, \bibinfo{year}{2024}.
\newblock \bibinfo{title}{How does urban heat island differ across urban
  functional zones? insights from 2d/3d urban morphology using geospatial big
  data}.
\newblock \bibinfo{journal}{Urban Climate} \bibinfo{volume}{53},
  \bibinfo{pages}{101787}.
\bibitem[{Lin(2004)}]{lin2004rouge}
\bibinfo{author}{Lin, C.Y.}, \bibinfo{year}{2004}.
\newblock \bibinfo{title}{Rouge: A package for automatic evaluation of
  summaries}, in: \bibinfo{booktitle}{Text summarization branches out}, pp.
  \bibinfo{pages}{74--81}.
\bibitem[{Lindenthal and Johnson(2021)}]{lindenthal2021machine}
\bibinfo{author}{Lindenthal, T.}, \bibinfo{author}{Johnson, E.B.},
  \bibinfo{year}{2021}.
\newblock \bibinfo{title}{Machine learning, architectural styles and property
  values}.
\newblock \bibinfo{journal}{The journal of real estate finance and economics} ,
  \bibinfo{pages}{1--32}.
\bibitem[{Liu et~al.(2023)Liu, Zeng, Ren, Li, Zhang, Yang, Li, Yang, Su, Zhu
  et~al.}]{liu2023grounding}
\bibinfo{author}{Liu, S.}, \bibinfo{author}{Zeng, Z.}, \bibinfo{author}{Ren,
  T.}, \bibinfo{author}{Li, F.}, \bibinfo{author}{Zhang, H.},
  \bibinfo{author}{Yang, J.}, \bibinfo{author}{Li, C.}, \bibinfo{author}{Yang,
  J.}, \bibinfo{author}{Su, H.}, \bibinfo{author}{Zhu, J.}, et~al.,
  \bibinfo{year}{2023}.
\newblock \bibinfo{title}{Grounding dino: Marrying dino with grounded
  pre-training for open-set object detection}.
\newblock \bibinfo{journal}{arXiv e-prints} , \bibinfo{pages}{arXiv--2303}.
\bibitem[{Mayer et~al.(2023)Mayer, Haas, Huang, Bernab{\'e}-Moreno, Rajagopal
  and Fischer}]{mayer2023estimating}
\bibinfo{author}{Mayer, K.}, \bibinfo{author}{Haas, L.},
  \bibinfo{author}{Huang, T.}, \bibinfo{author}{Bernab{\'e}-Moreno, J.},
  \bibinfo{author}{Rajagopal, R.}, \bibinfo{author}{Fischer, M.},
  \bibinfo{year}{2023}.
\newblock \bibinfo{title}{Estimating building energy efficiency from street
  view imagery, aerial imagery, and land surface temperature data}.
\newblock \bibinfo{journal}{Applied Energy} \bibinfo{volume}{333},
  \bibinfo{pages}{120542}.
\bibitem[{Milojevic-Dupont et~al.(2023)Milojevic-Dupont, Wagner, Nachtigall,
  Hu, Br{\"u}ser, Zumwald, Biljecki, Heeren, Kaack, Pichler
  et~al.}]{milojevic2023eubucco}
\bibinfo{author}{Milojevic-Dupont, N.}, \bibinfo{author}{Wagner, F.},
  \bibinfo{author}{Nachtigall, F.}, \bibinfo{author}{Hu, J.},
  \bibinfo{author}{Br{\"u}ser, G.B.}, \bibinfo{author}{Zumwald, M.},
  \bibinfo{author}{Biljecki, F.}, \bibinfo{author}{Heeren, N.},
  \bibinfo{author}{Kaack, L.H.}, \bibinfo{author}{Pichler, P.P.}, et~al.,
  \bibinfo{year}{2023}.
\newblock \bibinfo{title}{Eubucco v0. 1: European building stock
  characteristics in a common and open database for 200+ million individual
  buildings}.
\newblock \bibinfo{journal}{Scientific Data} \bibinfo{volume}{10},
  \bibinfo{pages}{147}.
\bibitem[{Nachtigall et~al.(2023)Nachtigall, Milojevic-Dupont, Wagner and
  Creutzig}]{nachtigall_predicting_2023}
\bibinfo{author}{Nachtigall, F.}, \bibinfo{author}{Milojevic-Dupont, N.},
  \bibinfo{author}{Wagner, F.}, \bibinfo{author}{Creutzig, F.},
  \bibinfo{year}{2023}.
\newblock \bibinfo{title}{Predicting building age from urban form at large
  scale}.
\newblock \bibinfo{journal}{Computers, Environment and Urban Systems}
  \bibinfo{volume}{105}, \bibinfo{pages}{102010}.
\newblock \URLprefix
  \url{https://linkinghub.elsevier.com/retrieve/pii/S019897152300073X},
  \DOIprefix\doi{10.1016/j.compenvurbsys.2023.102010}.
\bibitem[{Nouvel et~al.(2017)Nouvel, Zirak, Coors and
  Eicker}]{nouvel2017influence}
\bibinfo{author}{Nouvel, R.}, \bibinfo{author}{Zirak, M.},
  \bibinfo{author}{Coors, V.}, \bibinfo{author}{Eicker, U.},
  \bibinfo{year}{2017}.
\newblock \bibinfo{title}{The influence of data quality on urban heating demand
  modeling using 3d city models}.
\newblock \bibinfo{journal}{Computers, Environment and Urban Systems}
  \bibinfo{volume}{64}, \bibinfo{pages}{68--80}.
\bibitem[{Ogawa et~al.(2023)Ogawa, Zhao, Oki, Chen and
  Sekimoto}]{ogawa2023deep}
\bibinfo{author}{Ogawa, Y.}, \bibinfo{author}{Zhao, C.}, \bibinfo{author}{Oki,
  T.}, \bibinfo{author}{Chen, S.}, \bibinfo{author}{Sekimoto, Y.},
  \bibinfo{year}{2023}.
\newblock \bibinfo{title}{Deep learning approach for classifying the built year
  and structure of individual buildings by automatically linking street view
  images and gis building data}.
\newblock \bibinfo{journal}{IEEE Journal of Selected Topics in Applied Earth
  Observations and Remote Sensing} \bibinfo{volume}{16},
  \bibinfo{pages}{1740--1755}.
\bibitem[{Papineni et~al.(2002)Papineni, Roukos, Ward and
  Zhu}]{papineni2002bleu}
\bibinfo{author}{Papineni, K.}, \bibinfo{author}{Roukos, S.},
  \bibinfo{author}{Ward, T.}, \bibinfo{author}{Zhu, W.J.},
  \bibinfo{year}{2002}.
\newblock \bibinfo{title}{Bleu: a method for automatic evaluation of machine
  translation}, in: \bibinfo{booktitle}{Proceedings of the 40th annual meeting
  of the Association for Computational Linguistics}, pp.
  \bibinfo{pages}{311--318}.
\bibitem[{Pelizari et~al.(2021)Pelizari, Gei{\ss}, Aguirre, Santa~Mar{\'\i}a,
  Pe{\~n}a and Taubenb{\"o}ck}]{pelizari2021automated}
\bibinfo{author}{Pelizari, P.A.}, \bibinfo{author}{Gei{\ss}, C.},
  \bibinfo{author}{Aguirre, P.}, \bibinfo{author}{Santa~Mar{\'\i}a, H.},
  \bibinfo{author}{Pe{\~n}a, Y.M.}, \bibinfo{author}{Taubenb{\"o}ck, H.},
  \bibinfo{year}{2021}.
\newblock \bibinfo{title}{Automated building characterization for seismic risk
  assessment using street-level imagery and deep learning}.
\newblock \bibinfo{journal}{ISPRS Journal of Photogrammetry and Remote Sensing}
  \bibinfo{volume}{180}, \bibinfo{pages}{370--386}.
\bibitem[{Raghu et~al.(2023)Raghu, Bucher and De~Wolf}]{raghu_towards_2023}
\bibinfo{author}{Raghu, D.}, \bibinfo{author}{Bucher, M.J.J.},
  \bibinfo{author}{De~Wolf, C.}, \bibinfo{year}{2023}.
\newblock \bibinfo{title}{Towards a ‘resource cadastre’ for a circular
  economy – {Urban}-scale building material detection using street view
  imagery and computer vision}.
\newblock \bibinfo{journal}{Resources, Conservation and Recycling}
  \bibinfo{volume}{198}, \bibinfo{pages}{107140}.
\newblock \URLprefix
  \url{https://www.sciencedirect.com/science/article/pii/S0921344923002768},
  \DOIprefix\doi{10.1016/j.resconrec.2023.107140}.
\bibitem[{Ramalingam and Kumar(2023)}]{ramalingam_automatizing_2023}
\bibinfo{author}{Ramalingam, S.P.}, \bibinfo{author}{Kumar, V.},
  \bibinfo{year}{2023}.
\newblock \bibinfo{title}{Automatizing the generation of building usage maps
  from geotagged street view images using deep learning}.
\newblock \bibinfo{journal}{Building and Environment} \bibinfo{volume}{235},
  \bibinfo{pages}{110215}.
\newblock \URLprefix
  \url{https://www.sciencedirect.com/science/article/pii/S0360132323002421},
  \DOIprefix\doi{10.1016/j.buildenv.2023.110215}.
\bibitem[{Rosenfelder et~al.(2021)Rosenfelder, Wussow, Gust, Cremades and
  Neumann}]{rosenfelder2021predicting}
\bibinfo{author}{Rosenfelder, M.}, \bibinfo{author}{Wussow, M.},
  \bibinfo{author}{Gust, G.}, \bibinfo{author}{Cremades, R.},
  \bibinfo{author}{Neumann, D.}, \bibinfo{year}{2021}.
\newblock \bibinfo{title}{Predicting residential electricity consumption using
  aerial and street view images}.
\newblock \bibinfo{journal}{Applied Energy} \bibinfo{volume}{301},
  \bibinfo{pages}{117407}.
\bibitem[{Roth et~al.(2020)Roth, Martin, Miller and Jain}]{roth2020syncity}
\bibinfo{author}{Roth, J.}, \bibinfo{author}{Martin, A.},
  \bibinfo{author}{Miller, C.}, \bibinfo{author}{Jain, R.K.},
  \bibinfo{year}{2020}.
\newblock \bibinfo{title}{Syncity: Using open data to create a synthetic city
  of hourly building energy estimates by integrating data-driven and
  physics-based methods}.
\newblock \bibinfo{journal}{Applied Energy} \bibinfo{volume}{280},
  \bibinfo{pages}{115981}.
\bibitem[{Roy et~al.(2023)Roy, Pronk, Agugiaro and Ledoux}]{roy2023inferring}
\bibinfo{author}{Roy, E.}, \bibinfo{author}{Pronk, M.},
  \bibinfo{author}{Agugiaro, G.}, \bibinfo{author}{Ledoux, H.},
  \bibinfo{year}{2023}.
\newblock \bibinfo{title}{Inferring the number of floors for residential
  buildings}.
\newblock \bibinfo{journal}{International Journal of Geographical Information
  Science} \bibinfo{volume}{37}, \bibinfo{pages}{938--962}.
\bibitem[{Schug et~al.(2021)Schug, Frantz, van~der Linden and
  Hostert}]{schug2021gridded}
\bibinfo{author}{Schug, F.}, \bibinfo{author}{Frantz, D.},
  \bibinfo{author}{van~der Linden, S.}, \bibinfo{author}{Hostert, P.},
  \bibinfo{year}{2021}.
\newblock \bibinfo{title}{Gridded population mapping for germany based on
  building density, height and type from earth observation data using census
  disaggregation and bottom-up estimates}.
\newblock \bibinfo{journal}{Plos one} \bibinfo{volume}{16},
  \bibinfo{pages}{e0249044}.
\bibitem[{Simonyan and Zisserman(2014)}]{simonyan2014very}
\bibinfo{author}{Simonyan, K.}, \bibinfo{author}{Zisserman, A.},
  \bibinfo{year}{2014}.
\newblock \bibinfo{title}{Very deep convolutional networks for large-scale
  image recognition}.
\newblock \bibinfo{journal}{arXiv preprint arXiv:1409.1556} .
\bibitem[{Sun et~al.(2022a)Sun, Han, Nie, Xu, Zhang and
  Zhao}]{sun2022understandingenergy}
\bibinfo{author}{Sun, M.}, \bibinfo{author}{Han, C.}, \bibinfo{author}{Nie,
  Q.}, \bibinfo{author}{Xu, J.}, \bibinfo{author}{Zhang, F.},
  \bibinfo{author}{Zhao, Q.}, \bibinfo{year}{2022}a.
\newblock \bibinfo{title}{Understanding building energy efficiency with
  administrative and emerging urban big data by deep learning in glasgow}.
\newblock \bibinfo{journal}{Energy and buildings} \bibinfo{volume}{273},
  \bibinfo{pages}{112331}.
\bibitem[{Sun et~al.(2022b)Sun, Zhang, Duarte and Ratti}]{sun2022understanding}
\bibinfo{author}{Sun, M.}, \bibinfo{author}{Zhang, F.},
  \bibinfo{author}{Duarte, F.}, \bibinfo{author}{Ratti, C.},
  \bibinfo{year}{2022}b.
\newblock \bibinfo{title}{Understanding architecture age and style through deep
  learning}.
\newblock \bibinfo{journal}{Cities} \bibinfo{volume}{128},
  \bibinfo{pages}{103787}.
\bibitem[{Tarkhan et~al.(2025)Tarkhan, Klimenka, Fang, Duarte, Ratti and
  Reinhart}]{tarkhan_mapping_2025}
\bibinfo{author}{Tarkhan, N.}, \bibinfo{author}{Klimenka, M.},
  \bibinfo{author}{Fang, K.}, \bibinfo{author}{Duarte, F.},
  \bibinfo{author}{Ratti, C.}, \bibinfo{author}{Reinhart, C.},
  \bibinfo{year}{2025}.
\newblock \bibinfo{title}{Mapping facade materials utilizing zero-shot
  segmentation for applications in urban microclimate research}.
\newblock \bibinfo{journal}{Scientific Reports} \bibinfo{volume}{15},
  \bibinfo{pages}{5492}.
\newblock \URLprefix \url{https://www.nature.com/articles/s41598-025-86307-1},
  \DOIprefix\doi{10.1038/s41598-025-86307-1}. \bibinfo{note}{publisher: Nature
  Publishing Group}.
\bibitem[{Tooke et~al.(2014)Tooke, Coops and Webster}]{tooke2014predicting}
\bibinfo{author}{Tooke, T.R.}, \bibinfo{author}{Coops, N.C.},
  \bibinfo{author}{Webster, J.}, \bibinfo{year}{2014}.
\newblock \bibinfo{title}{Predicting building ages from lidar data with random
  forests for building energy modeling}.
\newblock \bibinfo{journal}{Energy and Buildings} \bibinfo{volume}{68},
  \bibinfo{pages}{603--610}.
\bibitem[{Wang et~al.(2021)Wang, Antos and Triveno}]{wang2021automatic}
\bibinfo{author}{Wang, C.}, \bibinfo{author}{Antos, S.E.},
  \bibinfo{author}{Triveno, L.M.}, \bibinfo{year}{2021}.
\newblock \bibinfo{title}{Automatic detection of unreinforced masonry buildings
  from street view images using deep learning-based image segmentation}.
\newblock \bibinfo{journal}{Automation in Construction} \bibinfo{volume}{132},
  \bibinfo{pages}{103968}.
\bibitem[{Wang et~al.(2024a)Wang, Ma, Chen, Zheng, Wan, Zhang and
  Zhong}]{wang_earthvqanet_2024}
\bibinfo{author}{Wang, J.}, \bibinfo{author}{Ma, A.}, \bibinfo{author}{Chen,
  Z.}, \bibinfo{author}{Zheng, Z.}, \bibinfo{author}{Wan, Y.},
  \bibinfo{author}{Zhang, L.}, \bibinfo{author}{Zhong, Y.},
  \bibinfo{year}{2024}a.
\newblock \bibinfo{title}{{EarthVQANet}: {Multi}-task visual question answering
  for remote sensing image understanding}.
\newblock \bibinfo{journal}{ISPRS Journal of Photogrammetry and Remote Sensing}
  \bibinfo{volume}{212}, \bibinfo{pages}{422--439}.
\newblock \URLprefix
  \url{https://linkinghub.elsevier.com/retrieve/pii/S0924271624001990},
  \DOIprefix\doi{10.1016/j.isprsjprs.2024.05.001}.
\bibitem[{Wang et~al.(2024b)Wang, Bai, Tan, Wang, Fan, Bai, Chen, Liu, Wang, Ge
  et~al.}]{wang2024qwen2}
\bibinfo{author}{Wang, P.}, \bibinfo{author}{Bai, S.}, \bibinfo{author}{Tan,
  S.}, \bibinfo{author}{Wang, S.}, \bibinfo{author}{Fan, Z.},
  \bibinfo{author}{Bai, J.}, \bibinfo{author}{Chen, K.}, \bibinfo{author}{Liu,
  X.}, \bibinfo{author}{Wang, J.}, \bibinfo{author}{Ge, W.}, et~al.,
  \bibinfo{year}{2024}b.
\newblock \bibinfo{title}{Qwen2-vl: Enhancing vision-language model's
  perception of the world at any resolution}.
\newblock \bibinfo{journal}{arXiv preprint arXiv:2409.12191} .
\bibitem[{Wang et~al.(2016)Wang, Chau, Ng and Leung}]{wang2016review}
\bibinfo{author}{Wang, Y.}, \bibinfo{author}{Chau, C.K.}, \bibinfo{author}{Ng,
  W.}, \bibinfo{author}{Leung, T.}, \bibinfo{year}{2016}.
\newblock \bibinfo{title}{A review on the effects of physical built environment
  attributes on enhancing walking and cycling activity levels within
  residential neighborhoods}.
\newblock \bibinfo{journal}{Cities} \bibinfo{volume}{50},
  \bibinfo{pages}{1--15}.
\bibitem[{Wang et~al.(2024c)Wang, Zhang, Dong, Guo, Tao and
  Zhang}]{wang_multi-view_2024}
\bibinfo{author}{Wang, Y.}, \bibinfo{author}{Zhang, Y.}, \bibinfo{author}{Dong,
  Q.}, \bibinfo{author}{Guo, H.}, \bibinfo{author}{Tao, Y.},
  \bibinfo{author}{Zhang, F.}, \bibinfo{year}{2024}c.
\newblock \bibinfo{title}{A multi-view graph neural network for building age
  prediction}.
\newblock \bibinfo{journal}{ISPRS Journal of Photogrammetry and Remote Sensing}
  \bibinfo{volume}{218}, \bibinfo{pages}{294--311}.
\newblock \URLprefix
  \url{https://linkinghub.elsevier.com/retrieve/pii/S0924271624003885},
  \DOIprefix\doi{10.1016/j.isprsjprs.2024.10.011}.
\bibitem[{Westrope et~al.(2014)Westrope, Banick and
  Levine}]{westrope2014groundtruthing}
\bibinfo{author}{Westrope, C.}, \bibinfo{author}{Banick, R.},
  \bibinfo{author}{Levine, M.}, \bibinfo{year}{2014}.
\newblock \bibinfo{title}{Groundtruthing openstreetmap building damage
  assessment}.
\newblock \bibinfo{journal}{Procedia engineering} \bibinfo{volume}{78},
  \bibinfo{pages}{29--39}.
\bibitem[{Wu and Biljecki(2021)}]{wu2021roofpedia}
\bibinfo{author}{Wu, A.N.}, \bibinfo{author}{Biljecki, F.},
  \bibinfo{year}{2021}.
\newblock \bibinfo{title}{Roofpedia: Automatic mapping of green and solar roofs
  for an open roofscape registry and evaluation of urban sustainability}.
\newblock \bibinfo{journal}{Landscape and Urban Planning}
  \bibinfo{volume}{214}, \bibinfo{pages}{104167}.
\bibitem[{Wu et~al.(2023a)Wu, Huang, Gao and Zhang}]{wu_mixed_2023}
\bibinfo{author}{Wu, M.}, \bibinfo{author}{Huang, Q.}, \bibinfo{author}{Gao,
  S.}, \bibinfo{author}{Zhang, Z.}, \bibinfo{year}{2023}a.
\newblock \bibinfo{title}{Mixed land use measurement and mapping with street
  view images and spatial context-aware prompts via zero-shot multimodal
  learning}.
\newblock \bibinfo{journal}{International Journal of Applied Earth Observation
  and Geoinformation} \bibinfo{volume}{125}, \bibinfo{pages}{103591}.
\newblock \URLprefix
  \url{https://linkinghub.elsevier.com/retrieve/pii/S1569843223004156},
  \DOIprefix\doi{10.1016/j.jag.2023.103591}.
\bibitem[{Wu et~al.(2023b)Wu, Ma, Banzhaf, Meadows, Yu, Guo, Sengupta, Cai and
  Zhao}]{wu2023first}
\bibinfo{author}{Wu, W.B.}, \bibinfo{author}{Ma, J.}, \bibinfo{author}{Banzhaf,
  E.}, \bibinfo{author}{Meadows, M.E.}, \bibinfo{author}{Yu, Z.W.},
  \bibinfo{author}{Guo, F.X.}, \bibinfo{author}{Sengupta, D.},
  \bibinfo{author}{Cai, X.X.}, \bibinfo{author}{Zhao, B.},
  \bibinfo{year}{2023}b.
\newblock \bibinfo{title}{A first chinese building height estimate at 10 m
  resolution (cnbh-10 m) using multi-source earth observations and machine
  learning}.
\newblock \bibinfo{journal}{Remote Sensing of Environment}
  \bibinfo{volume}{291}, \bibinfo{pages}{113578}.
\bibitem[{Xu et~al.(2023)Xu, Wong, Zhu, Heo and Shi}]{xu_semantic_2023}
\bibinfo{author}{Xu, F.}, \bibinfo{author}{Wong, M.S.}, \bibinfo{author}{Zhu,
  R.}, \bibinfo{author}{Heo, J.}, \bibinfo{author}{Shi, G.},
  \bibinfo{year}{2023}.
\newblock \bibinfo{title}{Semantic segmentation of urban building surface
  materials using multi-scale contextual attention network}.
\newblock \bibinfo{journal}{ISPRS Journal of Photogrammetry and Remote Sensing}
  \bibinfo{volume}{202}, \bibinfo{pages}{158--168}.
\newblock \URLprefix
  \url{https://www.sciencedirect.com/science/article/pii/S0924271623001600},
  \DOIprefix\doi{10.1016/j.isprsjprs.2023.06.001}.
\bibitem[{Yan and Huang(2022)}]{yan_estimation_2022}
\bibinfo{author}{Yan, Y.}, \bibinfo{author}{Huang, B.}, \bibinfo{year}{2022}.
\newblock \bibinfo{title}{Estimation of building height using a single street
  view image via deep neural networks}.
\newblock \bibinfo{journal}{ISPRS Journal of Photogrammetry and Remote Sensing}
  \bibinfo{volume}{192}, \bibinfo{pages}{83--98}.
\newblock \URLprefix
  \url{https://www.sciencedirect.com/science/article/pii/S0924271622002106},
  \DOIprefix\doi{10.1016/j.isprsjprs.2022.08.006}.
\bibitem[{Yang et~al.(2025)Yang, Lindquist and
  Van~Berkel}]{yang2025streetscape}
\bibinfo{author}{Yang, X.}, \bibinfo{author}{Lindquist, M.},
  \bibinfo{author}{Van~Berkel, D.}, \bibinfo{year}{2025}.
\newblock \bibinfo{title}{“streetscape” package in r: A reproducible method
  for analyzing open-source street view datasets and facilitating research for
  urban analytics}.
\newblock \bibinfo{journal}{SoftwareX} \bibinfo{volume}{29},
  \bibinfo{pages}{101981}.
\bibitem[{Zarbakhsh and McArdle(2023)}]{zarbakhsh_points--interest_2023}
\bibinfo{author}{Zarbakhsh, N.}, \bibinfo{author}{McArdle, G.},
  \bibinfo{year}{2023}.
\newblock \bibinfo{title}{Points-of-{Interest} from {Mapillary} {Street}-level
  {Imagery}: {A} {Dataset} {For} {Neighborhood} {Analytics}}, in:
  \bibinfo{booktitle}{2023 {IEEE} 39th {International} {Conference} on {Data}
  {Engineering} {Workshops} ({ICDEW})}, pp. \bibinfo{pages}{154--161}.
\newblock \URLprefix
  \url{https://ieeexplore.ieee.org/document/10148212/authors#authors},
  \DOIprefix\doi{10.1109/ICDEW58674.2023.00030}. \bibinfo{note}{iSSN:
  2473-3490}.
\bibitem[{Zeng et~al.(2024)Zeng, Goo, Wang, Chi, Wang and Boehm}]{zeng2024zero}
\bibinfo{author}{Zeng, Z.}, \bibinfo{author}{Goo, J.M.}, \bibinfo{author}{Wang,
  X.}, \bibinfo{author}{Chi, B.}, \bibinfo{author}{Wang, M.},
  \bibinfo{author}{Boehm, J.}, \bibinfo{year}{2024}.
\newblock \bibinfo{title}{Zero-shot building age classification from facade
  image using gpt-4}.
\newblock \bibinfo{journal}{The International Archives of the Photogrammetry,
  Remote Sensing and Spatial Information Sciences} \bibinfo{volume}{48},
  \bibinfo{pages}{457--464}.
\bibitem[{Zhang et~al.(2021)Zhang, Fan and Kong}]{zhang2021vgi3d}
\bibinfo{author}{Zhang, C.}, \bibinfo{author}{Fan, H.}, \bibinfo{author}{Kong,
  G.}, \bibinfo{year}{2021}.
\newblock \bibinfo{title}{Vgi3d: an interactive and low-cost solution for 3d
  building modelling from street-level vgi images}.
\newblock \bibinfo{journal}{Journal of Geovisualization and Spatial Analysis}
  \bibinfo{volume}{5}, \bibinfo{pages}{18}.
\bibitem[{Zhang et~al.(2024a)Zhang, Salazar-Miranda, Duarte, Vale, Hack, Chen,
  Liu, Batty and Ratti}]{zhang2024urban}
\bibinfo{author}{Zhang, F.}, \bibinfo{author}{Salazar-Miranda, A.},
  \bibinfo{author}{Duarte, F.}, \bibinfo{author}{Vale, L.},
  \bibinfo{author}{Hack, G.}, \bibinfo{author}{Chen, M.}, \bibinfo{author}{Liu,
  Y.}, \bibinfo{author}{Batty, M.}, \bibinfo{author}{Ratti, C.},
  \bibinfo{year}{2024}a.
\newblock \bibinfo{title}{Urban visual intelligence: Studying cities with
  artificial intelligence and street-level imagery}.
\newblock \bibinfo{journal}{Annals of the American Association of Geographers}
  \bibinfo{volume}{114}, \bibinfo{pages}{876--897}.
\bibitem[{Zhang et~al.(2024b)Zhang, Xiang, Kuang, Wang and
  Li}]{zhang2024archgpt}
\bibinfo{author}{Zhang, J.}, \bibinfo{author}{Xiang, R.},
  \bibinfo{author}{Kuang, Z.}, \bibinfo{author}{Wang, B.}, \bibinfo{author}{Li,
  Y.}, \bibinfo{year}{2024}b.
\newblock \bibinfo{title}{Archgpt: harnessing large language models for
  supporting renovation and conservation of traditional architectural
  heritage}.
\newblock \bibinfo{journal}{Heritage Science} \bibinfo{volume}{12},
  \bibinfo{pages}{220}.
\bibitem[{Zhang et~al.(2023)Zhang, Liu and Biljecki}]{zhang_knowledge_2023}
\bibinfo{author}{Zhang, Y.}, \bibinfo{author}{Liu, P.},
  \bibinfo{author}{Biljecki, F.}, \bibinfo{year}{2023}.
\newblock \bibinfo{title}{Knowledge and topology: {A} two layer spatially
  dependent graph neural networks to identify urban functions with time-series
  street view image}.
\newblock \bibinfo{journal}{ISPRS Journal of Photogrammetry and Remote Sensing}
  \bibinfo{volume}{198}, \bibinfo{pages}{153--168}.
\newblock \URLprefix
  \url{https://linkinghub.elsevier.com/retrieve/pii/S0924271623000680},
  \DOIprefix\doi{10.1016/j.isprsjprs.2023.03.008}.
\bibitem[{Zhao et~al.(2021)Zhao, Liu, Hao, Lu, Liu and Zhou}]{zhao2021bounding}
\bibinfo{author}{Zhao, K.}, \bibinfo{author}{Liu, Y.}, \bibinfo{author}{Hao,
  S.}, \bibinfo{author}{Lu, S.}, \bibinfo{author}{Liu, H.},
  \bibinfo{author}{Zhou, L.}, \bibinfo{year}{2021}.
\newblock \bibinfo{title}{Bounding boxes are all we need: street view image
  classification via context encoding of detected buildings}.
\newblock \bibinfo{journal}{IEEE Transactions on Geoscience and Remote Sensing}
  \bibinfo{volume}{60}, \bibinfo{pages}{1--17}.
\bibitem[{Zhao et~al.(2019)Zhao, Bo, Chen, Tiede, Blaschke and
  Emery}]{zhao_exploring_2019}
\bibinfo{author}{Zhao, W.}, \bibinfo{author}{Bo, Y.}, \bibinfo{author}{Chen,
  J.}, \bibinfo{author}{Tiede, D.}, \bibinfo{author}{Blaschke, T.},
  \bibinfo{author}{Emery, W.J.}, \bibinfo{year}{2019}.
\newblock \bibinfo{title}{Exploring semantic elements for urban scene
  recognition: {Deep} integration of high-resolution imagery and
  {OpenStreetMap} ({OSM})}.
\newblock \bibinfo{journal}{ISPRS Journal of Photogrammetry and Remote Sensing}
  \bibinfo{volume}{151}, \bibinfo{pages}{237--250}.
\newblock \URLprefix
  \url{https://linkinghub.elsevier.com/retrieve/pii/S0924271619300887},
  \DOIprefix\doi{10.1016/j.isprsjprs.2019.03.019}.
\bibitem[{Zhou et~al.(2017)Zhou, Lapedriza, Khosla, Oliva and
  Torralba}]{zhou2017places}
\bibinfo{author}{Zhou, B.}, \bibinfo{author}{Lapedriza, A.},
  \bibinfo{author}{Khosla, A.}, \bibinfo{author}{Oliva, A.},
  \bibinfo{author}{Torralba, A.}, \bibinfo{year}{2017}.
\newblock \bibinfo{title}{Places: A 10 million image database for scene
  recognition}.
\newblock \bibinfo{journal}{IEEE Transactions on Pattern Analysis and Machine
  Intelligence} .
\bibitem[{Zhu et~al.(2025)Zhu, Wang, Chen, Liu, Ye, Gu, Tian, Duan, Su, Shao,
  Gao, Cui, Wang, Cao, Liu, Wei, Zhang, Wang, Xu, Li, Wang, Deng, Li, He,
  Jiang, Luo, Wang, He, Shi, Zhang, Shao, He, Xiong, Qu, Sun, Jiao, Lv, Wu,
  Zhang, Deng, Ge, Chen, Wang, Dou, Lu, Zhu, Lu, Lin, Qiao, Dai and
  Wang}]{zhu2025internvl3}
\bibinfo{author}{Zhu, J.}, \bibinfo{author}{Wang, W.}, \bibinfo{author}{Chen,
  Z.}, \bibinfo{author}{Liu, Z.}, \bibinfo{author}{Ye, S.},
  \bibinfo{author}{Gu, L.}, \bibinfo{author}{Tian, H.}, \bibinfo{author}{Duan,
  Y.}, \bibinfo{author}{Su, W.}, \bibinfo{author}{Shao, J.},
  \bibinfo{author}{Gao, Z.}, \bibinfo{author}{Cui, E.}, \bibinfo{author}{Wang,
  X.}, \bibinfo{author}{Cao, Y.}, \bibinfo{author}{Liu, Y.},
  \bibinfo{author}{Wei, X.}, \bibinfo{author}{Zhang, H.},
  \bibinfo{author}{Wang, H.}, \bibinfo{author}{Xu, W.}, \bibinfo{author}{Li,
  H.}, \bibinfo{author}{Wang, J.}, \bibinfo{author}{Deng, N.},
  \bibinfo{author}{Li, S.}, \bibinfo{author}{He, Y.}, \bibinfo{author}{Jiang,
  T.}, \bibinfo{author}{Luo, J.}, \bibinfo{author}{Wang, Y.},
  \bibinfo{author}{He, C.}, \bibinfo{author}{Shi, B.}, \bibinfo{author}{Zhang,
  X.}, \bibinfo{author}{Shao, W.}, \bibinfo{author}{He, J.},
  \bibinfo{author}{Xiong, Y.}, \bibinfo{author}{Qu, W.}, \bibinfo{author}{Sun,
  P.}, \bibinfo{author}{Jiao, P.}, \bibinfo{author}{Lv, H.},
  \bibinfo{author}{Wu, L.}, \bibinfo{author}{Zhang, K.}, \bibinfo{author}{Deng,
  H.}, \bibinfo{author}{Ge, J.}, \bibinfo{author}{Chen, K.},
  \bibinfo{author}{Wang, L.}, \bibinfo{author}{Dou, M.}, \bibinfo{author}{Lu,
  L.}, \bibinfo{author}{Zhu, X.}, \bibinfo{author}{Lu, T.},
  \bibinfo{author}{Lin, D.}, \bibinfo{author}{Qiao, Y.}, \bibinfo{author}{Dai,
  J.}, \bibinfo{author}{Wang, W.}, \bibinfo{year}{2025}.
\newblock \bibinfo{title}{Internvl3: Exploring advanced training and test-time
  recipes for open-source multimodal models}.
\newblock \URLprefix \url{https://arxiv.org/abs/2504.10479},
  \href{http://arxiv.org/abs/2504.10479}{{\tt arXiv:2504.10479}}.
\bibitem[{Zia et~al.(2022)Zia, Riaz and Ghafoor}]{zia2022transforming}
\bibinfo{author}{Zia, U.}, \bibinfo{author}{Riaz, M.M.},
  \bibinfo{author}{Ghafoor, A.}, \bibinfo{year}{2022}.
\newblock \bibinfo{title}{Transforming remote sensing images to textual
  descriptions}.
\newblock \bibinfo{journal}{International Journal of Applied Earth Observation
  and Geoinformation} \bibinfo{volume}{108}, \bibinfo{pages}{102741}.
\bibitem[{Zietz et~al.(2008)Zietz, Zietz and Sirmans}]{zietz2008determinants}
\bibinfo{author}{Zietz, J.}, \bibinfo{author}{Zietz, E.N.},
  \bibinfo{author}{Sirmans, G.S.}, \bibinfo{year}{2008}.
\newblock \bibinfo{title}{Determinants of house prices: a quantile regression
  approach}.
\newblock \bibinfo{journal}{The Journal of Real Estate Finance and Economics}
  \bibinfo{volume}{37}, \bibinfo{pages}{317--333}.
\bibitem[{Zou and Wang(2021)}]{zou_detecting_2021}
\bibinfo{author}{Zou, S.}, \bibinfo{author}{Wang, L.}, \bibinfo{year}{2021}.
\newblock \bibinfo{title}{Detecting individual abandoned houses from google
  street view: {A} hierarchical deep learning approach}.
\newblock \bibinfo{journal}{ISPRS Journal of Photogrammetry and Remote Sensing}
  \bibinfo{volume}{175}, \bibinfo{pages}{298--310}.
\newblock \URLprefix
  \url{https://linkinghub.elsevier.com/retrieve/pii/S0924271621000915},
  \DOIprefix\doi{10.1016/j.isprsjprs.2021.03.020}.

\end{thebibliography}
\clearpage

\end{sloppypar}
\end{document}